\RequirePackage{fix-cm}
\documentclass{wechat}
\DeclareFontShape{T1}{wchatsans}{b}{n}{<-> s * [1] wechat/wchatsans}{}
\DeclareFontShape{T1}{wchatsans}{bx}{n}{<-> s * [1] wechat/wchatsans}{}
\usepackage[T1]{fontenc}
\usepackage{amsmath,amssymb}
\usepackage{graphicx}
\usepackage{booktabs,tabularx,array}
\usepackage{longtable}
\usepackage{xcolor}
\usepackage{colortbl} % Shaded model-group rows in the supplementary result tables.
\usepackage{tikz}
\usetikzlibrary{arrows.meta,positioning}
\usepackage{hyperref}
\hypersetup{
  pdftitle={From Evaluation to Evolution: Benchmarking Image Editing and Evolving Image Models},
  pdfauthor={Jun He, Canmiao Fu, Ying Zhang, Zhipeng Huang, Chenyu Mu, Yazhen Xie, Chen Li, Jing LYU, Weijia Li, Xiang Zhang},
  pdfsubject={Preprint},
  colorlinks=true,
  linkcolor=wchatgreen,
  citecolor=wchatgreen,
  urlcolor=wchatgreen
}
\usepackage{url}

\usepackage{xurl}
\usepackage{needspace}
\AtBeginDocument{%
}

\let\reportsection\section
\renewcommand{\section}{\Needspace*{5\baselineskip}\reportsection}

\DeclareRobustCommand{\benchname}[1]{{\fontfamily{wchatsansmedium}\fontseries{m}\selectfont #1}}
\newcommand{\bench}{\benchname{WeGenEditBench}}
\newcommand{\studio}{\benchname{WeGenEvoStudio}}
\newcommand{\single}{\benchname{WeGenEditBench-SingleImageEdit}}
\newcommand{\multi}{\benchname{WeGenEditBench-MultiImageEdit}}
\newcommand{\textedit}{\benchname{WeGenEditBench-TextEdit}}

\newcolumntype{Y}{>{\raggedright\arraybackslash}X}
\graphicspath{{figures/}}

\definecolor{WeGenGreen}{HTML}{07C160}
\newcommand{\hlbest}[1]{\textbf{#1}}
\newcommand{\hlsecond}[1]{\underline{#1}}

\definecolor{EvoGain}{HTML}{16834A}
\definecolor{EvoLoss}{HTML}{C43B3B}
\newcommand{\evogain}[1]{\textcolor{EvoGain}{#1}}
\newcommand{\evoloss}[1]{\textcolor{EvoLoss}{#1}}
\DeclareMathSizes{7.5}{10}{7}{5}
\DeclareMathSizes{8.5}{10}{7}{5}

\newcommand{\modelgroup}[2]{\multicolumn{#1}{@{}l@{}}{\makebox[0pt][l]{\begingroup\setlength{\fboxsep}{1.5pt}\colorbox{TableTint}{\makebox[\dimexpr\linewidth-2\fboxsep\relax]{\strut\textit{#2}}}\endgroup}}\\}

\title{\textcolor{wchatgreen}{From Evaluation to Evolution:} Benchmarking Image Editing and Evolving Image Models}
\author[1,2,*]{Jun He}
\author[2]{Canmiao Fu}
\author[2]{Ying Zhang}
\author[2]{Zhipeng Huang}
\author[3]{Chenyu Mu}
\author[4]{Yazhen Xie}
\author[2]{Chen Li}
\author[2]{Jing LYU}
\author[5]{Weijia Li}
\author[1,\dagger]{Xiang Zhang}
\affiliation[1]{Sun Yat-sen University}
\affiliation[2]{WeChat Vision, Tencent Inc.}
\affiliation[3]{Xidian University}
\affiliation[4]{Fudan University}
\affiliation[5]{Tsinghua Shenzhen International Graduate School}

\newcommand{\manuscriptabstract}{
Real-world image editing requires coordinating multiple visual transformations with content preservation and reference constraints, yet aggregate scores can readily obscure unmet local requirements. Effective evaluation must therefore identify local failures and translate them into priorities for subsequent model improvement. We introduce \bench{}, comprising 2,740 evaluation instances and 148 task types across single-image, text-intensive, and multi-reference editing. Pre-generated checklists shared across models decompose each request into atomic requirements; item-level deductions and visual evidence are established first and then summarized into scores for edit execution, preservation, visual quality, and task-specific correctness. To support the experience-acquisition component of recursive self-improvement (RSI), we develop \studio{}, a unified multi-agent harness for RSI-oriented learning that integrates model-conditioned failure diagnosis, retrieval-grounded data construction, supervised adaptation, and re-evaluation. Among 15 evaluated configurations, GPT-Image-2 leads single-image editing with an Overall score of 4.076, while Seedream-5.0-Pro leads text-intensive editing at 4.291. Evolution experiments across four generation and editing benchmarks demonstrate substantial targeted gains: on their report-selected Top-5 single-image targets, FLUX.2 [dev] improves Overall from 1.083 to 3.517, while Qwen-Image-Edit-2511 gains 13.2\%. Together, our benchmark and harness provide a foundation for developing more reliable image models through interpretable evaluation and feedback-driven learning.
}

\newcommand{\manuscriptteaser}{%
\begingroup
\setlength{\intextsep}{6pt}
\begin{figure}[!ht]
\centering
\setlength{\abovecaptionskip}{4pt}
\includegraphics[width=\linewidth]{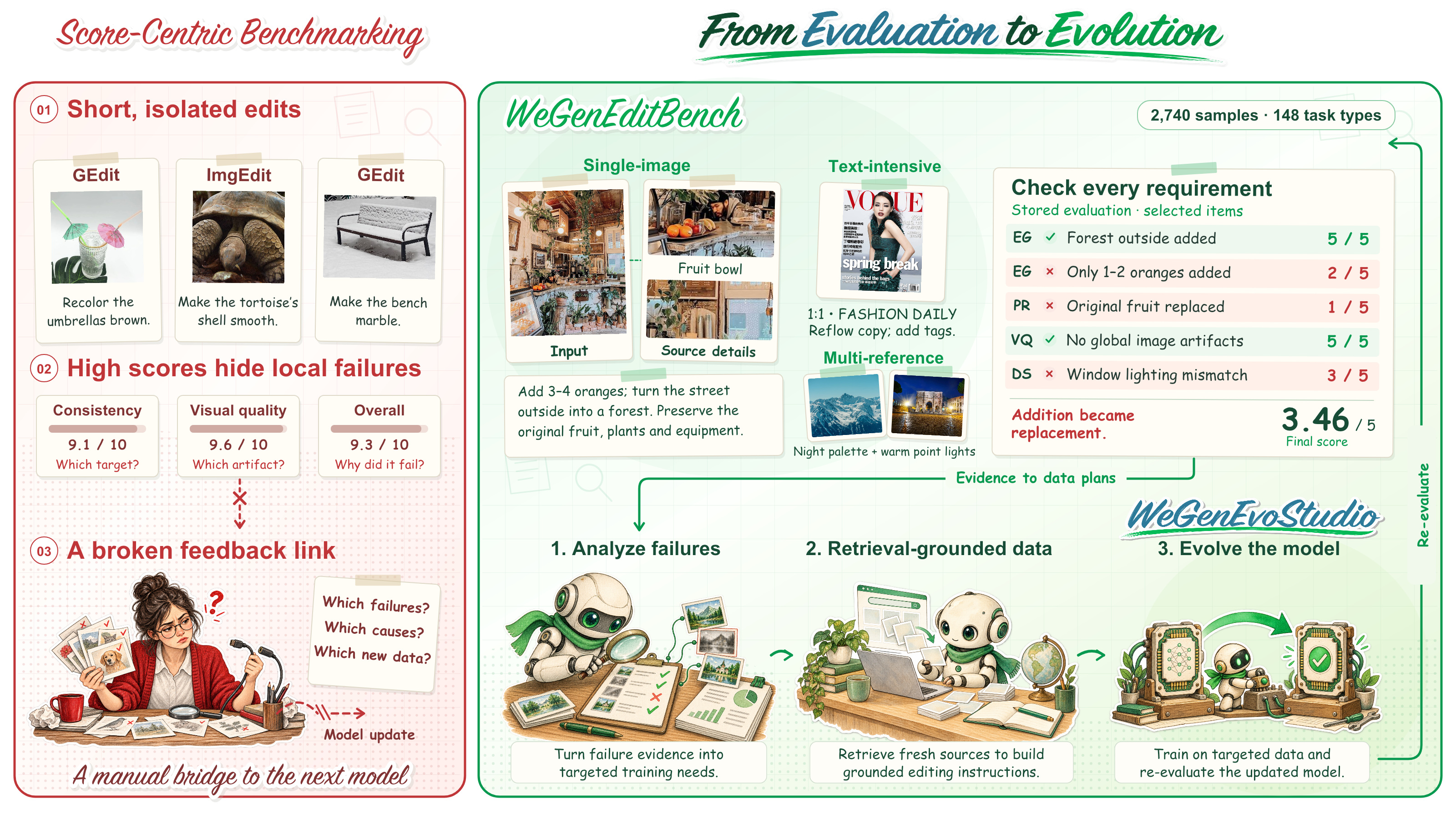}
\caption{From evaluation to evolution. Aggregate scores can obscure unmet editing requirements and leave training priorities unspecified (left). \bench{} exposes these requirements through sample-specific checklists, scores, and rationales; \studio{} turns the resulting evidence into targeted data construction, model updates, and re-evaluation (right). Only selected checklist items are shown.}
\label{fig:teaser}
\end{figure}
\endgroup
}

\abstract{\manuscriptabstract}

\newcommand{\InternshipInstitution}{WeChat Vision, Tencent Inc.}

\begin{document}
\maketitle
\begingroup
\renewcommand{\thefootnote}{\fnsymbol{footnote}}
\footnotetext[1]{Work done during Jun He's internship at \InternshipInstitution}
\footnotetext[2]{Corresponding author(s). E-mail(s): \texttt{zhangx795@mail.sysu.edu.cn}}
\endgroup
\manuscriptteaser
\section{Introduction}
\label{sec:introduction}

Rapid advances in text-to-image and unified multimodal models are expanding the scope of natural-language image editing. Proprietary systems such as GPT-Image-2 and Nano Banana Pro, alongside open-weight models such as Qwen-Image-Edit and FLUX.2 [dev], demonstrate strong instruction understanding and precise editing capabilities~\citep{gptimage2,nanobananapro,qwenedit,flux2}. Real-world requests, however, increasingly couple multiple visual transformations with preservation and reference constraints, such as replacing a product background while retaining its packaging text and appearance. Current models still exhibit a high-score paradox in these settings: visual plausibility or success on a single salient edit does not establish that all coupled requirements have been jointly satisfied. As illustrated in Figure~\ref{fig:teaser} (left), aggregate scores can conceal local omissions and leave the feedback connection between evaluation and model improvement incomplete.

Existing benchmarks have advanced editing evaluation along complementary directions. GEdit and ImgEdit cover common operations, with ImgEdit also assessing multi-turn interaction~\citep{step1x,imgedit}; Complex-Edit, GEditBench v2, UnicBench, and ConceptEdit extend evaluation to compositional operations, open-ended requests, reasoning, and fine-grained concepts, respectively~\citep{complexedit,geditv2,unicedit,conceptedit}. Evaluation is also becoming more granular: ComplexBench-Edit assesses instruction dependencies~\citep{complexbenchedit}, Qwen-Image-Bench introduces hierarchical criteria for text-to-image generation~\citep{qwenimagebench}, and Mind-Brush incorporates checklist-based evaluation of knowledge- and reasoning-driven generation~\citep{mindbrush}. Nevertheless, general image editing requires sample-specific criteria grounded in source images to jointly verify requested changes, preservation conditions, and reference assignments across these three settings. Top-down dimensional scoring can be influenced by salient visual changes, leaving local omissions and constraint violations difficult to trace.

To this end, we introduce \bench{}, comprising 2,740 instances and 148 task types across single-image, text-intensive, and multi-reference editing. Real-world requests guide automated construction and human calibration. Pre-generated checklists shared across models decompose each instance into atomic requirements under Edit Goal (EG), Preservation (PR), Visual Quality (VQ), and Domain-specific Capability (DS). A fixed vision-language model (VLM) judge must record item-level rationales and deductions before assigning dimensional scores, making omitted edits, unintended changes, and incorrect reference binding traceable to specific requirements and visual evidence.

Evaluation should support not only model comparison but also decisions about what a model should learn next. Recent research on recursive self-improvement (RSI) studies how systems can retain capability gains and use them in subsequent improvement cycles~\citep{stop,dgm,rsi_survey}. For image models, a key step is to translate observed failures into training experience tailored to their current weaknesses. Data-centric development, exemplified by DataPerf, and feedback-based reinforcement learning, exemplified by DDPO, provide methodological foundations~\citep{dataperf,ddpo}. OmniEdit, AnyEdit, and ConceptEdit enrich supervision through specialist models and concept expansion~\citep{omniedit,anyedit,conceptedit}; DataEvolver explores validation-driven data iteration~\citep{dataevolver_text,dataevolver_visual}, while IterComp and JarvisEvo jointly refine generators or editors and their evaluators~\citep{itercomp,jarvisevo}. General image editing nevertheless faces a central planning challenge: translating unmet requirements into subsequent data construction plans and refreshing their priorities as model capabilities evolve. Coarse category averages can locate broad weaknesses, but do not specify recurring constraint violations or the targeted supervision needed to address them.

To connect failure diagnosis with targeted learning, we introduce \studio{}, a unified multi-agent harness for RSI-oriented model improvement (Figure~\ref{fig:teaser}, right). The Diagnosis Agent combines category statistics, item-level rationales, and cross-case visual review to formulate task-level data prescriptions; the Construction Agent uses these prescriptions to retrieve new concepts and source images and synthesize targeted editing instructions and paired training examples. After supervised adaptation produces an updated model, the re-evaluation interface can refresh subsequent learning priorities, supporting a feedback loop for targeted experience acquisition in RSI-oriented systems~\citep{absolutezero,rsi_survey}. Experiments across four editing and text-to-image benchmarks reveal fine-grained capability gaps, targeted adaptation gains, and cross-task trade-offs. Our main contributions are:

\begin{itemize}
\item \textbf{A fine-grained image editing benchmark.} \bench{} comprises 2,740 instances and 148 task types across single-image, text-intensive, and multi-reference editing. Sample-specific atomic checklists establish item-level deductions before dimensional aggregation, connecting scores to underlying visual evidence.
\item \textbf{A multi-agent harness for model improvement.} \studio{} translates benchmark failure diagnosis into task-level data prescriptions, retrieval-grounded experience synthesis, and supervised parameter updates. Its re-evaluation interface refreshes learning priorities to support RSI-oriented experience acquisition.
\item \textbf{Comprehensive capability diagnosis and empirical model evolution.} We systematically evaluate 15 widely used configurations to identify bottlenecks in execution, preservation, and reference binding, and conduct model-evolution experiments with Qwen-Image and FLUX.2 [dev], demonstrating substantial targeted gains and cross-task capability trade-offs.
\end{itemize}

\vspace{-3mm}
\section{Related Work}
\label{sec:related}

\paragraph{Image editing benchmarks.}
Evaluation of image generation and editing typically examines instruction following, preservation, visual quality, and knowledge-grounded synthesis~\citep{step1x,imgedit,unicedit,gptimgeval}, and has recently expanded to fine-grained visual reasoning and the explanation of synthetic-content anomalies~\citep{blinktwice,loki}. EditBench evaluates inpainting~\citep{editbench}, and MagicBrush provides single- and multi-turn annotations~\citep{magicbrush}. GEdit covers common operations~\citep{step1x}, while ImgEdit introduces task-specific criteria and multi-turn evaluation~\citep{imgedit}; however, concise requests do not fully reflect intertwined transformations and preservation constraints. More recent work examines reasoning in UnicBench~\citep{unicedit}, open-ended requests in GEditBench v2~\citep{geditv2}, controlled operation counts in Complex-Edit~\citep{complexedit}, and instructions and their dependencies in ComplexBench-Edit~\citep{complexbenchedit}. Aggregate scores can nevertheless obscure failed local targets or constraints. Joint verification of fine-grained requirements across single-image, text-intensive, and multi-reference editing remains underexplored.

\paragraph{Agentic multimodal data construction.}
Closing capability gaps requires diverse and reliable supervision~\citep{ultraedit,conceptedit,echo4o}. InstructPix2Pix synthesizes editing pairs~\citep{instructpix2pix}, OmniEdit combines specialist supervision with sampling~\citep{omniedit}, and AnyEdit expands task coverage through adaptive synthesis~\citep{anyedit}; these approaches largely follow predefined task inventories. Agentic methods introduce greater adaptability: Gen-n-Val couples prompt generation with visual validation~\citep{gennval}, ScaleEditor coordinates expansion and verification through hierarchical agents~\citep{scaleedit}, and DataEvolver uses rejection feedback to revise retrieval and synthesis~\citep{dataevolver_text}. Other systems use drafts or executable canvases for controlled refinement~\citep{draco,genclaw}. Improving synthesis quality alone, however, does not determine which supervision a target model needs to address its specific failure modes.

\paragraph{Feedback-driven learning and recursive self-improvement.}
Feedback can support output refinement in Self-Refine~\citep{selfrefine}, retention of verbal experience in Reflexion~\citep{reflexion}, or policy optimization in DDPO~\citep{ddpo}. RSI research further investigates how retained improvements participate in subsequent cycles and how the improvement mechanism itself can evolve~\citep{rsi_survey}. Examples include code or agent-logic self-modification in STOP~\citep{stop}, G\"odel Agent~\citep{godelagent}, and the Darwin G\"odel Machine~\citep{dgm}, as well as task proposal conditioned on learner progress in Absolute Zero~\citep{absolutezero}. In multimodal learning, IterComp updates reward models and generators~\citep{itercomp}, JarvisEvo jointly optimizes editing and evaluation policies~\citep{jarvisevo}, and Visual DataEvolver performs validation-driven data expansion~\citep{dataevolver_visual}. Our focus is complementary: translating a target image model's unmet requirements into task-level data prescriptions and new training experience, with traceable connections between diagnostic evidence and supervision throughout the feedback loop.

\vspace{-3mm}
\section{WeGenEditBench}
\label{sec:benchmark}

\subsection{Real-World Scenarios and Task Coverage}
\label{sec:benchmark-overview}
\label{sec:benchmark-statistics}
\label{sec:tasks}

\begin{figure}[t!]
\centering
\includegraphics[width=\linewidth]{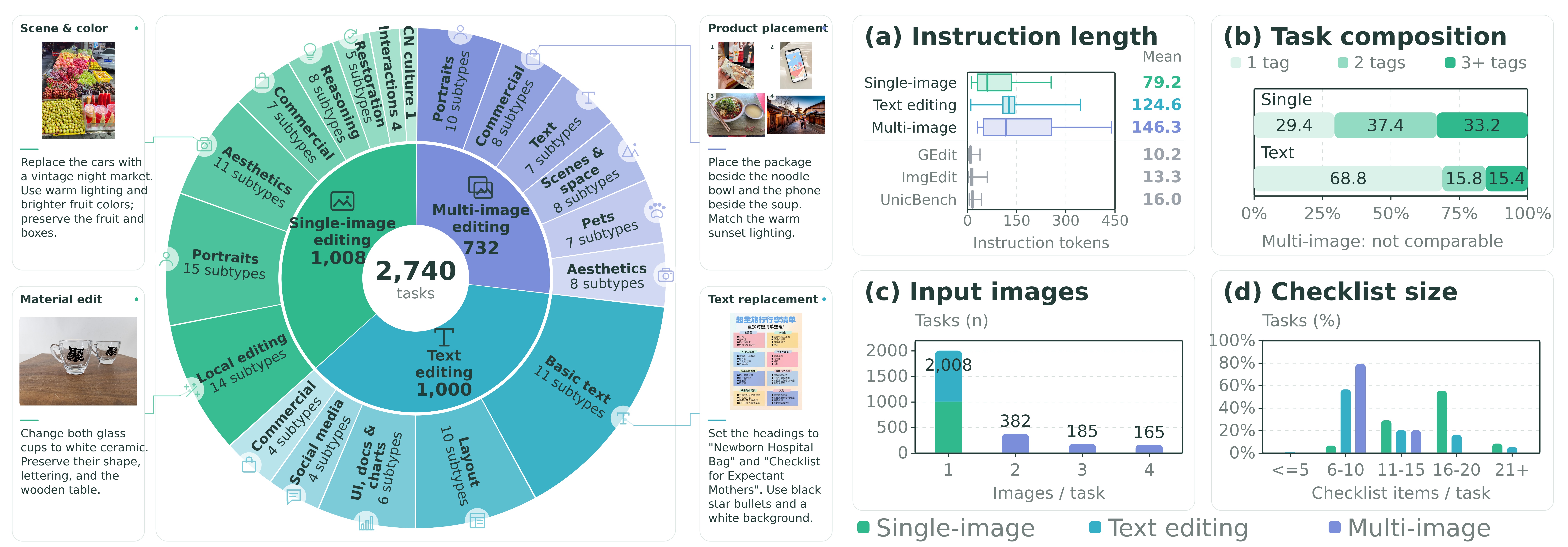}
\caption{Taxonomy and characteristics of \bench{}. Left: settings, task families, examples, instance counts (inner), and subtypes (outer). Right: distributions of instruction length (a), tag multiplicity (b), input images (c), and checklist items (d). Colors indicate settings; lengths preserve source languages.}
\label{fig:benchmark-statistics}
\end{figure}

To assess the joint fulfillment of real-world composite editing requirements, \bench{} covers three complementary settings: \textbf{single-image editing}, \textbf{text-intensive editing}, and \textbf{multi-reference editing} (Figure~\ref{fig:benchmark-statistics}, left). These settings reflect common user needs in everyday retouching, layout design, and attribute customization from multiple sources. The taxonomy is derived from over 16,000 real-world requests and examples collected from public web sources, model documentation, and open-source projects. Single-image tasks cover local, portrait, aesthetic, commercial, interaction, restoration, and reasoning-oriented edits; text-intensive tasks cover textual content, typography, and layout; multi-reference tasks test the binding of identity, product, style, and scene attributes to designated sources. Automated construction and human calibration yield 2,740 instances spanning 19 task families and 148 task types, connecting application scenarios to concrete preservation, typography, and reference constraints.

\subsection{Source-Grounded Synthesis and Instruction Complexity}
\label{sec:construction}

To ensure that composite instructions are visually actionable rather than arbitrary combinations, we compose compatible operations around visible subjects in the source image, explicitly specify preservation, layout, and reference constraints, and use human review to remove unsupported or contradictory combinations (Figure~\ref{fig:benchmark-pipeline} in the appendix). As shown in Figure~\ref{fig:benchmark-statistics}(a--c), mean instruction lengths reach 79.2, 124.6, and 146.3 tokens for the single-image, text-intensive, and multi-reference subsets, respectively, compared with 10.2--16.0 tokens for GEdit, ImgEdit, and UnicBench. The additional text primarily specifies multi-step transformations and preservation boundaries: 70.6\% of single-image instances and 31.2\% of text-intensive instances carry multiple task labels, while multi-reference tasks coordinate two to four input images with distinct roles. These dense requirements probe model bottlenecks in simultaneously satisfying interacting transformations, fine-grained preservation, and reference binding.

\subsection{Evidence-Grounded Evaluation and Failure Diagnosis}
\label{sec:atomic-checklists}
\label{sec:metrics}

To expose local failures hidden by high aggregate scores, we evaluate four dimensions where applicable: \textbf{Edit Goal (EG)}, \textbf{Preservation (PR)}, \textbf{Visual Quality (VQ)}, and \textbf{Domain-specific Capability (DS)}. From each instruction and its source images, we derive independently verifiable items specifying a target, attribute or location, and expected state. Text editing separates text-side and image-side checks, while multi-reference editing binds attributes to designated source images. Checklists are generated before evaluation, fixed within each evaluation version, and shared across compared models. As shown in Figure~\ref{fig:benchmark-statistics}(d), single-image checklists most often contain 16--20 items, whereas text-intensive and multi-reference checklists most often contain 6--10.

\begin{figure}[t]
\centering
\includegraphics[width=\linewidth]{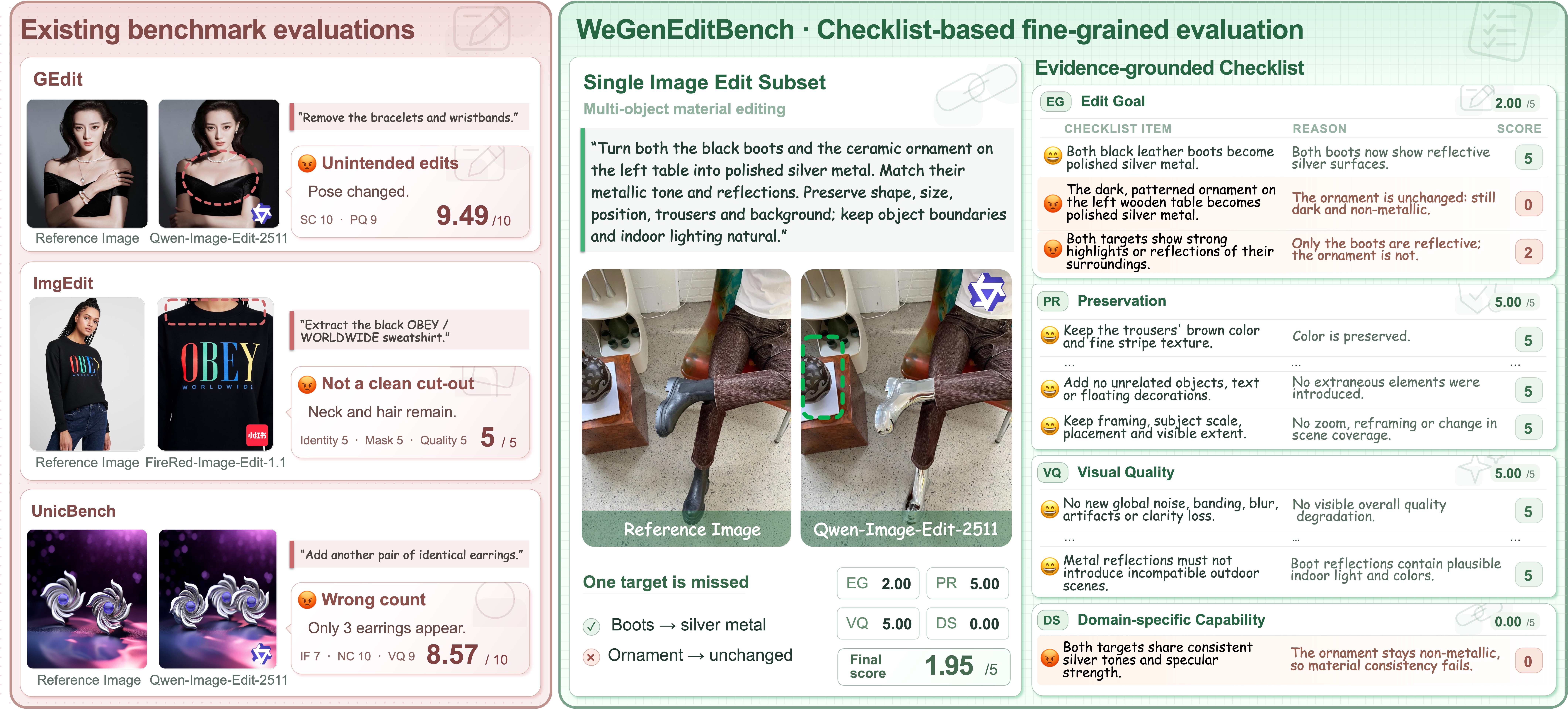}
\vspace{-2mm}
\caption{From scores to requirement-level diagnosis. Left: selected cases receive high scores despite visible errors. Right: \bench{} identifies an omitted edit through checklist scores and rationales.}
\label{fig:atomic-evaluation-example}
\end{figure}

Holistic dimensional judgments can overemphasize salient visual changes while overlooking missed targets or unintended alterations. Figure~\ref{fig:atomic-evaluation-example} illustrates this motivation through high-scoring failures on existing benchmarks (left) and the verification mechanism of \bench{} (right). Our fixed VLM judge must first record visual rationales and item-level deductions $\delta_{i,j}\in[0,5]$, and then assign dimensional scores on a 0--5 scale according to each item's importance under the scoring rubric. This structured attribution helps reduce salience bias. In the right-hand example, silver boots coexist with an unchanged ceramic ornament: although $\mathrm{PR}=\mathrm{VQ}=5$, item-level verification exposes the missed target and inconsistent material luster across targets, yielding $\mathrm{EG}=2$, $\mathrm{DS}=0$, and an Overall score of 1.95. Three trained experts randomly audit 10\% of instances in each subset. Following checklist corrections, all compared models are re-evaluated on affected instances using the same revised checklist; Appendix~\ref{app:human-audit} details the audit scope.

\begin{figure}[t!]
\centering
\includegraphics[width=\linewidth]{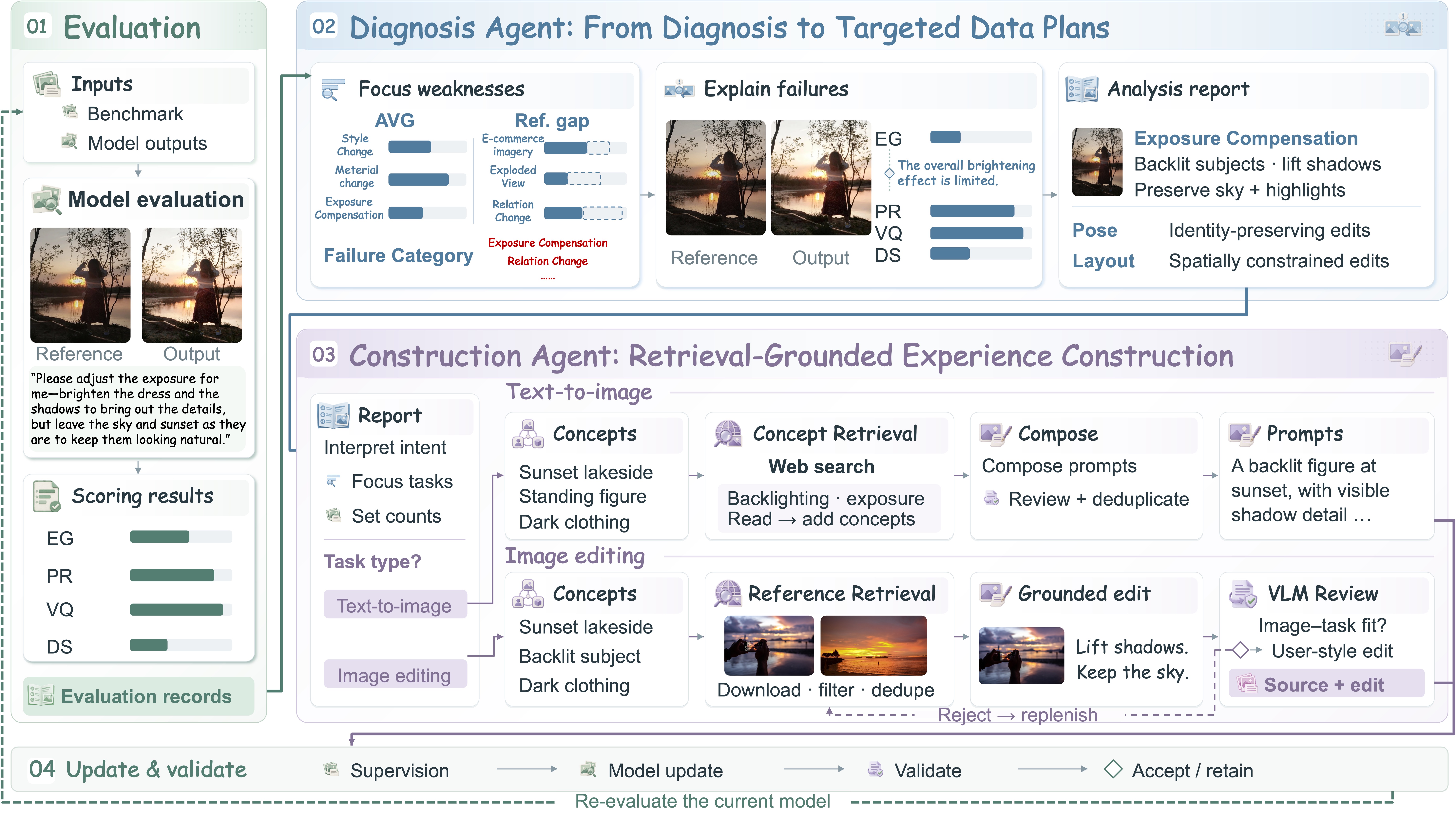}
\caption{\studio{} translates evaluation evidence into targeted training data. The Diagnosis Agent selects weaknesses and forms data prescriptions; the Construction Agent retrieves concepts or source images to synthesize generation and editing tasks. Supervised updates and re-evaluation close the loop. The exposure-editing example and score bars illustrate the workflow.}
\vspace{-3mm}
\label{fig:overview}
\end{figure}

\paragraph{Subset-specific aggregation.}
Single-image editing weights EG, PR, and VQ and subtracts the mean DS penalty over the applicable task set $T$, with a zero penalty when $T$ is empty:
\begin{equation}
S_{\mathrm{single}}=\operatorname{clip}_{[0,5]}\!\left(0.6S_{\mathrm{EG}}+0.2S_{\mathrm{PR}}+0.2S_{\mathrm{VQ}}-\frac{1}{|T|}\sum_{t\in T}\frac{5-S_{\mathrm{DS},t}}{4}\right).
\label{eq:single-score}
\end{equation}
Text-intensive editing evaluates the two sides separately and combines them when image-side requirements are present; otherwise, it uses only $S_{\mathrm{text}}$. Multi-reference editing combines goal fulfillment, reference-attribute preservation, and visual quality:
\begin{equation}
S_{\mathrm{text\text{-}edit}}=0.75S_{\mathrm{text}}+0.25S_{\mathrm{image}},\qquad
S_{\mathrm{multi}}=0.4S_{\mathrm{EG}}+0.4S_{\mathrm{PR}}+0.2S_{\mathrm{VQ}}.
\label{eq:text-score}
\end{equation}
Each subset's Overall is the mean of valid instance scores, counting each instance once. The retained item-level evidence directly supports capability diagnosis in \studio{}.

\section{WeGenEvoStudio}
\label{sec:studio}

\subsection{Unified Evaluation-to-Learning Framework}
\label{sec:state}

To translate the unmet requirements identified in Section~\ref{sec:benchmark} into learning objectives, we introduce \studio{}, a unified multi-agent harness centered on model-specific data prescriptions (Figure~\ref{fig:overview}). For model parameters $\theta_t$, evaluation records $\mathcal{E}_t$ contain instructions, source and output images, scores, and available rationales. Diagnosis extracts a capability profile $\Phi_t$ and a data prescription $\mathcal{T}_t$, which guide the construction of targeted tasks $\mathcal{X}_t$. These tasks are paired with generated targets to form a supervised training set $\mathcal{D}_t$, followed by parameter updates and re-evaluation:
\begin{equation}
\begin{aligned}
\theta_t &\xrightarrow{\text{evaluate}} \mathcal{E}_t
\xrightarrow{\text{diagnose}} (\Phi_t,\mathcal{T}_t)
\xrightarrow{\text{construct}} \mathcal{X}_t\\
&\xrightarrow{\text{supervise}} \mathcal{D}_t
\xrightarrow{\text{train}(\cdot;\lambda_t)} \theta_{t+1}
\xrightarrow{\text{re-evaluate}} \mathcal{E}_{t+1}.
\end{aligned}
\label{eq:evidence-to-data}
\end{equation}

\texttt{gemini-3-flash-preview} serves as the backbone LLM for agent planning and tool use. The underlying agentic harness encapsulates 68 atomic tools in a standard registry. Isolated subcontexts, restricted tool allowlists, and content fingerprints link evaluation records, data prescriptions, and model-checkpoint lineage. Pluggable feedback adapters support external public benchmarks: they accept coarse category scores from GEdit and ImgEdit, native rationales from WeGenBench, and fine-grained checklists from \bench{}, mapping external failure categories to a shared internal taxonomy. Re-evaluation of a successor $\theta_{t+1}$ can refresh prescription priorities, providing the infrastructure for an experience-acquisition loop oriented toward RSI~\citep{rsi_survey}. Appendix~\ref{app:studio-implementation} provides implementation details.

\subsection{From Failure Diagnosis to Targeted Data Prescriptions}
\label{sec:reflection}
\label{sec:planner}

Category averages can locate weak areas but do not explain the specific conditions that repeatedly trigger failures. As shown in Figure~\ref{fig:deep-analysis-agent} in the appendix, the Diagnosis Agent combines score-based category selection with case-level visual review. Given the target model's category score $\bar{S}_c(\theta_t)$ and an anchor $\theta_{\mathrm{anchor}}$ evaluated under the same protocol, it computes the relative deficit $\Delta_c^{\mathrm{rel}}=\bar{S}_c(\theta_{\mathrm{anchor}})-\bar{S}_c(\theta_t)$. Approximately 60\% of the slots prioritize the largest relative deficits, while the remaining 40\% target the lowest absolute scores. The initial selection is
\begin{equation}
\mathcal{C}_{\mathrm{seed}}=\operatorname{Deduplicate}\!\left(
\operatorname{Top}_{\lfloor0.6K\rfloor,c}\!\left(\Delta_c^{\mathrm{rel}}\right)
\cup
\operatorname{Bottom}_{\lceil0.4K\rceil,c}\!\left(\bar{S}_c(\theta_t)\right)
\right).
\label{eq:category-selection}
\end{equation}
After deduplication, reciprocal-rank ordering fills any remaining slots to obtain $\mathcal{C}_{\mathrm{focus}}$ with $K$ categories, retaining at least one category selected by absolute weakness. Without a compatible anchor, selection relies on the target model's own statistics.

Within each candidate category, the agent contrasts representative failures and successes, using VLM review of instructions and images to identify omitted operations, affected objects, and unintended changes. Cross-case synthesis yields the capability profile $\Phi_t$. The resulting structured prescription, $\mathcal{T}_t=\{(\psi(c),\mathcal{S}_c,\mathcal{R}_c^{\mathrm{edit}},\mathcal{R}_c^{\mathrm{pres}},N_c)\}_{c\in\mathcal{C}_{\mathrm{focus}}}$, maps each weak category to an internal taxonomy label $\psi(c)$ and specifies its quota $N_c$. When supported by the evidence, it also records source-image conditions $\mathcal{S}_c$ and editing and preservation requirements $(\mathcal{R}_c^{\mathrm{edit}},\mathcal{R}_c^{\mathrm{pres}})$. For example, failures during background replacement motivate a prescription for clearly visible product images and explicit preservation constraints. To prevent benchmark-instance leakage, individual failure cases remain in the diagnostic report as attribution evidence; only task labels, quotas, and abstract category notes are passed downstream, without reconstructing benchmark scenes.

\subsection{Retrieval-Grounded Experience Construction and Controlled Supervised Adaptation}
\label{sec:experience}

To address diagnosed weaknesses in new visual contexts, the Construction Agent maps target tasks to the shared taxonomy and allocates quotas across categories, without importing instruction templates from external benchmarks (Figure~\ref{fig:instruction-construction-agent} in the appendix). The editing branch retrieves unedited source images from a library or web search using category-specific queries, followed by format, resolution, and exact or perceptual duplicate checks. The text-to-image branch uses web-text retrieval to expand long-tail concepts and reduce dependence on high-frequency entities in the language model's prior. Each synthesis call grounds the target task and category notes in a new source image, explicitly specifying visible edit targets, intended transformations, and preservation conditions while rejecting incompatible sources. VLM review then checks feasibility and rewrites the instruction as a natural user request.

Given an approved source image and instruction, the downstream supervision pipeline invokes a high-capability generation backend to synthesize candidate target images. Visual-quality and task-alignment filters consistent with the construction stage reject candidates with structural artifacts or incomplete transformations, yielding a high-confidence training set $\mathcal{D}_t$ for the supervised update $\theta_{t+1}=\operatorname{Train}(\theta_t,\mathcal{D}_t;\lambda_t)$. In this design, the teacher supervision pipeline maintains per-example rendering quality, while upstream diagnostic prescriptions and retrieval grounding determine how the limited training budget is allocated across tasks. Appendices~\ref{app:construction-agent} and~\ref{app:construction-boundary} detail the construction interface and its task-level boundaries.

\begin{table}[t]
\centering
\caption{Quantitative comparison on \single{}. Bold and underlined scores denote the best and second best.}
\label{tab:wegen-single}
\label{tab:benchmark}
\fontsize{8pt}{8pt}\selectfont
\renewcommand{\arraystretch}{1.0}
\setlength{\tabcolsep}{4.5pt}
\begin{tabularx}{\linewidth}{@{}l|*{4}{>{\centering\arraybackslash}X}|>{\centering\arraybackslash}X}
\toprule
\textbf{Model Name} & EG {\normalsize$\uparrow$} & PR {\normalsize$\uparrow$} & VQ {\normalsize$\uparrow$} & DS {\normalsize$\uparrow$} & \textbf{Overall} {\normalsize$\uparrow$} \\
\midrule
\modelgroup{6}{Closed-source Models}
GPT-Image-2 & \hlbest{4.298} & 4.049 & \hlbest{4.463} & \hlsecond{4.182} & \hlbest{4.076} \\
Nano-Banana-Pro & 4.159 & 3.752 & 4.370 & 4.050 & 3.882 \\
Qwen-Image-3.0 & 4.120 & 3.469 & 4.331 & 4.126 & 3.813 \\
Seedream-5.0-Pro & \hlsecond{4.207} & \hlsecond{4.084} & \hlsecond{4.434} & \hlbest{4.206} & \hlsecond{4.029} \\
\midrule
\modelgroup{6}{Open-source Models}
HunyuanImage-3.0-Instruct-Distil & 4.020 & \hlbest{4.175} & 4.280 & 3.952 & 3.841 \\
HunyuanImage-3.0-Instruct-Distil-CoT & 4.071 & 4.065 & 4.265 & 4.008 & 3.862 \\
FireRed-Image-Edit-1.1 & 4.091 & 3.550 & 4.227 & 3.902 & 3.736 \\
JoyAI-Image-Edit & 4.042 & 3.547 & 4.169 & 3.794 & 3.667 \\
Qwen-Image-Edit-2511 & 3.963 & 3.567 & 4.095 & 3.708 & 3.588 \\
Boogu-Image-0.1-Edit & 3.850 & 3.718 & 4.092 & 3.768 & 3.564 \\
SenseNova-U1-8B-MoT-CoT & 3.948 & 3.636 & 3.932 & 3.605 & 3.535 \\
SenseNova-U1-8B-MoT & 4.030 & 3.408 & 3.917 & 3.674 & 3.552 \\
LongCat-Image-Edit & 3.761 & 3.439 & 4.003 & 3.643 & 3.410 \\
HiDream-O1-Image & 3.158 & 2.942 & 3.855 & 3.231 & 2.816 \\
FLUX.2 [dev] & 2.056 & 3.343 & 3.393 & 1.975 & 1.828 \\
\bottomrule
\end{tabularx}
\end{table}
\vspace{-3mm}

\section{Experiments}
\label{sec:experiments}

\subsection{Experimental Setup}
\label{sec:setup}

\paragraph{Benchmark evaluation.}
We evaluate 15 model configurations. Closed-source models comprise GPT-Image-2~\citep{gptimage2}, Nano-Banana-Pro~\citep{nanobananapro}, Qwen-Image-3.0~\citep{qwenimage3}, and Seedream-5.0-Pro~\citep{seedream5pro}. Open-source configurations comprise HunyuanImage-3.0-Instruct-Distil and its CoT variant~\citep{hunyuanimage3}, FireRed-Image-Edit-1.1~\citep{firered11}, JoyAI-Image-Edit~\citep{joyaiimage}, Qwen-Image-Edit-2511~\citep{qwenedit}, Boogu-Image-0.1-Edit~\citep{booguimage}, SenseNova-U1-8B-MoT and its CoT variant~\citep{sensenovau1}, LongCat-Image-Edit~\citep{longcatimage}, HiDream-O1-Image~\citep{hidreamo1}, and FLUX.2 [dev]~\citep{flux2}. All 15 configurations cover single-image and text-intensive editing, while six configurations supporting multiple input images cover multi-reference editing. All scores in Tables~\ref{tab:wegen-single}--\ref{tab:wegen-multi} use a fixed version of \texttt{gemini-3.1-pro-preview} and the shared-checklist protocol. Occasional invalid responses introduce small differences in valid-instance sets across models; very small aggregate gaps should therefore be interpreted as comparable performance. Diagnostic proportions are reported with 95\% confidence intervals from 2,000 source-image-cluster bootstrap replicates (Appendices~\ref{app:primary-task-analysis} and~\ref{app:diagnostic-analysis}).

\paragraph{Model evolution.}
We train separate rank-16 LoRA adapters for each backbone--benchmark pair, using Qwen-Image-2512 for text-to-image generation, Qwen-Image-Edit-2511 for editing, and FLUX.2 [dev] for both~\citep{lora,qwenimage2512,qwenedit,flux2}. Each experiment uses 25,000 synthetic examples, evenly allocated to the Top-$K=5$ diagnosed weak categories with 5,000 examples per category, and trains on 32 Cambricon MLU590 accelerators with a per-device batch size of 1. Experiments span four benchmarks with different feedback granularities: GEdit and ImgEdit provide coarse category scores~\citep{step1x,imgedit}, WeGenBench additionally provides native rationales~\citep{wegenbench}, and \bench{} supplies fine-grained checklist evidence. All experiments share the same retrieval-based constructor and use new source instances. Before-and-after comparisons are paired on common-valid instances (Appendix~\ref{app:experimental-details}). We use Gemini-3-flash as the orchestrator in WeGenEvoStudio.

\subsection{Scenario-Specific Capabilities and Fine-Grained Diagnosis}
\label{sec:quantitative-results}
\label{sec:benchmark-results}
\label{sec:benchmark-analysis}

\paragraph{Scenario-dependent strengths and distinct visual and textual capabilities.}
Tables~\ref{tab:wegen-single}--\ref{tab:wegen-multi} show that no single model dominates every scenario. GPT-Image-2 leads single-image editing at 4.076, closely followed by Seedream at 4.029, and multi-reference editing at 4.138. Seedream-5.0-Pro, however, surpasses it on text-intensive editing, scoring 4.291 versus 4.132. Decomposing Table~\ref{tab:wegen-text} shows that GPT-Image-2 still achieves higher image-side EG on text tasks (4.604 versus 4.362), whereas Seedream leads all four text-side dimensions. Its higher Overall thus reflects stronger textual execution, preservation, and presentation. Among open-source configurations, HunyuanImage-3.0-Instruct-Distil-CoT leads single-image and text-intensive Overall at 3.862 and 3.882, respectively. These results reveal a distinction between visual-semantic editing in continuous pixel space and typographic control over discrete characters: leadership on a single-image benchmark does not establish reliability on text-rich or multi-reference requests.

\begin{table}[t]
\centering
\caption{Quantitative comparison on \textedit{}. Bold and underlined scores denote the best and second best.}
\label{tab:wegen-text}
\fontsize{8pt}{8pt}\selectfont
\renewcommand{\arraystretch}{1.0}
\setlength{\tabcolsep}{1.0pt}
\begin{tabularx}{\linewidth}{@{}l|*{4}{>{\centering\arraybackslash}X}|*{4}{>{\centering\arraybackslash}X}|c}
\toprule
\raisebox{-5pt}[0pt][0pt]{\textbf{Model Name}} & \multicolumn{4}{c|}{\textbf{Text-side}} & \multicolumn{4}{c|}{\textbf{Image-side}} & \raisebox{-5pt}[0pt][0pt]{\textbf{Overall} {\normalsize$\uparrow$}} \\
\cmidrule(lr){2-5}\cmidrule(lr){6-9}
& EG {\normalsize$\uparrow$} & PR {\normalsize$\uparrow$} & VQ {\normalsize$\uparrow$} & DS {\normalsize$\uparrow$} & EG {\normalsize$\uparrow$} & PR {\normalsize$\uparrow$} & VQ {\normalsize$\uparrow$} & DS {\normalsize$\uparrow$} & \\
\midrule
\modelgroup{10}{Closed-source Models}
GPT-Image-2 & 4.289 & 4.031 & \hlsecond{4.329} & \hlsecond{4.399} & \hlbest{4.604} & 3.919 & \hlsecond{4.421} & \hlbest{4.452} & \hlsecond{4.132} \\
Nano-Banana-Pro & 4.295 & \hlsecond{4.092} & 4.310 & 4.322 & 4.396 & \hlsecond{4.208} & 4.232 & 4.306 & 4.116 \\
Qwen-Image-3.0 & \hlsecond{4.367} & 3.732 & \hlsecond{4.329} & 4.385 & 4.509 & 3.869 & 4.406 & 4.414 & 4.109 \\
Seedream-5.0-Pro & \hlbest{4.415} & \hlbest{4.327} & \hlbest{4.416} & \hlbest{4.504} & 4.362 & 4.207 & \hlbest{4.455} & \hlsecond{4.429} & \hlbest{4.291} \\
\midrule
\modelgroup{10}{Open-source Models}
SenseNova-U1-8B-MoT-CoT & 4.039 & 3.880 & 3.844 & 3.943 & 4.340 & 4.092 & 3.935 & 4.090 & 3.762 \\
HunyuanImage-3.0-Instruct-Distil & 3.914 & 3.801 & 3.917 & 4.133 & 4.361 & 4.177 & 4.109 & 4.214 & 3.742 \\
HunyuanImage-3.0-Instruct-Distil-CoT & 4.090 & 3.807 & 4.102 & 4.188 & 4.422 & \hlbest{4.258} & 4.194 & 4.323 & 3.882 \\
SenseNova-U1-8B-MoT & 4.013 & 3.809 & 3.902 & 3.924 & 4.408 & 3.830 & 3.895 & 4.099 & 3.733 \\
JoyAI-Image-Edit & 3.900 & 3.617 & 3.751 & 3.975 & \hlsecond{4.565} & 3.925 & 4.129 & 4.287 & 3.630 \\
FireRed-Image-Edit-1.1 & 3.712 & 3.308 & 3.646 & 3.890 & 4.497 & 3.622 & 3.932 & 4.245 & 3.423 \\
Qwen-Image-Edit-2511 & 3.658 & 3.583 & 3.655 & 3.820 & 4.384 & 3.830 & 3.969 & 4.112 & 3.429 \\
Boogu-Image-0.1-Edit & 3.842 & 3.180 & 3.732 & 3.915 & 4.412 & 3.459 & 3.898 & 4.197 & 3.482 \\
LongCat-Image-Edit & 3.027 & 3.560 & 3.385 & 3.320 & 4.187 & 3.639 & 3.789 & 3.949 & 2.901 \\
HiDream-O1-Image & 2.893 & 2.541 & 3.213 & 3.145 & 3.568 & 3.051 & 3.483 & 3.539 & 2.502 \\
FLUX.2 [dev] & 2.183 & 3.313 & 3.640 & 2.384 & 2.724 & 3.331 & 3.570 & 2.782 & 2.140 \\
\bottomrule
\end{tabularx}
\end{table}

\begin{table}[t]
\centering
\caption{Quantitative comparison on \multi{}. Bold and underlined scores denote the best and second best.}
\label{tab:wegen-multi}
\fontsize{8pt}{8pt}\selectfont
\renewcommand{\arraystretch}{1.0}
\setlength{\tabcolsep}{4.5pt}
\begin{tabularx}{\linewidth}{@{}l|*{3}{>{\centering\arraybackslash}X}|>{\centering\arraybackslash}X}
\toprule
\textbf{Model Name} & EG {\normalsize$\uparrow$} & PR {\normalsize$\uparrow$} & VQ {\normalsize$\uparrow$} & \textbf{Overall} {\normalsize$\uparrow$} \\
\midrule
\modelgroup{5}{Closed-source Models}
GPT-Image-2 & \hlbest{4.358} & \hlbest{4.134} & \hlsecond{3.707} & \hlbest{4.138} \\
Nano-Banana-Pro & 4.031 & 3.993 & 3.340 & 3.877 \\
Seedream-5.0-Pro & \hlsecond{4.270} & \hlsecond{4.093} & \hlbest{3.750} & \hlsecond{4.095} \\
\midrule
\modelgroup{5}{Open-source Models}
HunyuanImage-3.0-Instruct-Distil & 3.279 & 3.493 & 3.091 & 3.327 \\
FireRed-Image-Edit-1.1 & 2.714 & 2.977 & 3.055 & 2.887 \\
Qwen-Image-Edit-2511 & 2.589 & 2.884 & 3.093 & 2.808 \\
\bottomrule
\end{tabularx}
\end{table}

\paragraph{Structural bottlenecks in text rendering, dynamic interaction, and reference binding.}
Separating dimensions further exposes model weaknesses. JoyAI-Image-Edit achieves image-side EG of 4.565 on text-intensive tasks and 4.552 on posters, nearly matching GPT-Image-2 on the latter (Table~\ref{tab:wegen-text-primary-poster}), yet its text-side EG across text-intensive tasks is only 3.900. Successful visual modification therefore does not ensure correct textual execution. Across single-image task families (Appendix~\ref{app:primary-task-analysis}), the base Hunyuan model leads PR in six of eight families. On action and interaction editing (Table~\ref{tab:wegen-single-primary-action}), however, its high PR of 3.975 accompanies EG of only 3.640, compared with GPT-Image-2's 4.376. This pattern is consistent with conservative editing that preserves content while leaving requested changes incomplete. Its CoT variant raises interaction EG to 4.011, alleviating this execution bottleneck. Multi-reference editing (Table~\ref{tab:wegen-multi}) exposes a particularly pronounced gap for open-source models: the strongest evaluated open model, Hunyuan, scores 3.327 Overall, with EG and PR of 3.279 and 3.493, below GPT-Image-2's 4.358 and 4.134. On multi-reference text editing (Table~\ref{tab:wegen-multi-primary-text}), open-model EG ranges from just 1.839 to 2.700. These scenario and dimension splits distinguish weaknesses in dynamic interaction and cross-image attribute binding from general visual quality.

\paragraph{Atomic evidence beneath high aggregate scores.}
Strong aggregate performance still conceals many unmet requirements (Appendix~\ref{app:diagnostic-analysis}, Tables~\ref{tab:diagnostic-high-scores} and~\ref{tab:diagnostic-joint-sides}). Among GPT-Image-2's 644 single-image outputs with Overall $\ge4$, 315 contain at least one severely failed item scored $\le2$: 48.9\%, with a 95\% confidence interval of [45.1\%, 52.8\%]. Even among outputs scoring $\ge4.5$, this proportion remains 31.3\%. In multi-reference editing, 108 GPT-Image-2 outputs share the identical dimensional-and-Overall score vector $(5,5,3,4.6)$, yet the underlying checklists distinguish 44 outputs with a severe item failure from 64 without one. On 280 shared tasks with requirements on both the text and image sides, only 25.0\% of GPT-Image-2 outputs and 27.1\% of Seedream outputs satisfy every item on both sides at a score of $\ge4$. These findings show that dimensional aggregation loses consequential information when multiple constraints interact; atomic checklists recover local failures that remain hidden in aggregate scores.

\subsection{Feedback-Driven Adaptation and Capability Evolution}
\label{sec:evolution-results}
\label{sec:evolution-analysis}
\label{sec:qualitative-results}

\paragraph{Targeted gains across feedback formats and broader benchmark performance.}
Table~\ref{tab:evolution} demonstrates targeted gains under different feedback granularities, while Tables~\ref{tab:evolution-full}--\ref{tab:evolution-sensitivity} report full-benchmark performance, including categories outside the diagnostic targets, to assess broader behavior. On the report-selected Top-5 target sets, Qwen and FLUX.2 improve by 0.968 and 0.230 on WeGenBench, respectively, with full-benchmark gains of 0.670 and 0.124. On \single{}, Qwen gains 0.373 Overall (13.2\%), while FLUX.2 improves from 1.083 to 3.517. Even on GEdit and ImgEdit, which provide only coarse category scores, FLUX.2 gains 3.944 and 1.362 Overall on the target sets, respectively, accompanied by full-benchmark gains of 3.009 and 1.269 (Appendix~\ref{app:evolution-results}). These results demonstrate the framework's applicability across feedback formats and across generation and editing tasks.

\begin{table}[t]
\centering
\caption{Quantitative comparison on report-selected Top-5 categories across four benchmarks. FLUX.2 WeGenBench scores use 326 common-valid pairs. Target sets differ by backbone; green/red marks positive/negative $\Delta$.}
\label{tab:evolution}
\fontsize{8pt}{8pt}\selectfont
\renewcommand{\arraystretch}{1.12}
\setlength{\tabcolsep}{0.35pt}
\begin{tabular*}{\linewidth}{@{\extracolsep{\fill}}l*{13}{c}@{}}
\toprule
\raisebox{-13pt}[0pt][0pt]{\textbf{Model}} & \textbf{T2I} & \multicolumn{12}{c}{\textbf{Image editing}} \\
\cmidrule(lr){2-2}\cmidrule(lr){3-14}
& \shortstack{\textbf{WeGen}\\\textbf{Bench}} & \multicolumn{5}{c}{\textbf{\single{}}} & \multicolumn{3}{c}{\textbf{GEdit}} & \multicolumn{4}{c}{\textbf{ImgEdit}} \\
\cmidrule(lr){2-2}\cmidrule(lr){3-7}\cmidrule(lr){8-10}\cmidrule(lr){11-14}
& F $\uparrow$ & EG $\uparrow$ & PR $\uparrow$ & VQ $\uparrow$ & DS $\uparrow$ & F $\uparrow$ & SC $\uparrow$ & PQ $\uparrow$ & O $\uparrow$ & I $\uparrow$ & Q $\uparrow$ & D $\uparrow$ & F $\uparrow$ \\
\midrule
Qwen-Image & 7.700 & 3.244 & 2.646 & 3.500 & 3.565 & 2.834 & 8.515 & 7.131 & 7.511 & 4.621 & 4.374 & 4.400 & 4.465 \\
\addlinespace[1.5pt]
\textbf{+ WeGenEvoStudio} & 8.668 & 3.705 & 2.650 & 4.105 & 3.437 & 3.208 & 8.172 & 7.241 & 7.476 & 4.621 & 4.416 & 4.450 & 4.496 \\
\addlinespace[1.5pt]
$\Delta$ & \evogain{+0.968} & \evogain{+0.461} & \evogain{+0.003} & \evogain{+0.605} & \evoloss{-0.128} & \evogain{+0.373} & \evoloss{-0.343} & \evogain{+0.109} & \evoloss{-0.035} & +0.000 & \evogain{+0.042} & \evogain{+0.051} & \evogain{+0.031} \\
\midrule
FLUX.2 [dev] & 7.767 & 1.550 & 3.178 & 2.780 & 0.809 & 1.083 & 3.425 & 7.592 & 3.465 & 1.915 & 2.626 & 2.873 & 2.471 \\
\addlinespace[1.5pt]
\textbf{+ WeGenEvoStudio} & 7.997 & 3.767 & 3.865 & 4.414 & 3.380 & 3.517 & 7.949 & 8.032 & 7.409 & 3.821 & 3.722 & 3.955 & 3.833 \\
\addlinespace[1.5pt]
$\Delta$ & \evogain{+0.230} & \evogain{+2.217} & \evogain{+0.687} & \evogain{+1.634} & \evogain{+2.571} & \evogain{+2.434} & \evogain{+4.524} & \evogain{+0.440} & \evogain{+3.944} & \evogain{+1.906} & \evogain{+1.096} & \evogain{+1.082} & \evogain{+1.362} \\
\bottomrule
\end{tabular*}
\end{table}

\begin{figure}[t]
\centering
\includegraphics[width=\linewidth]{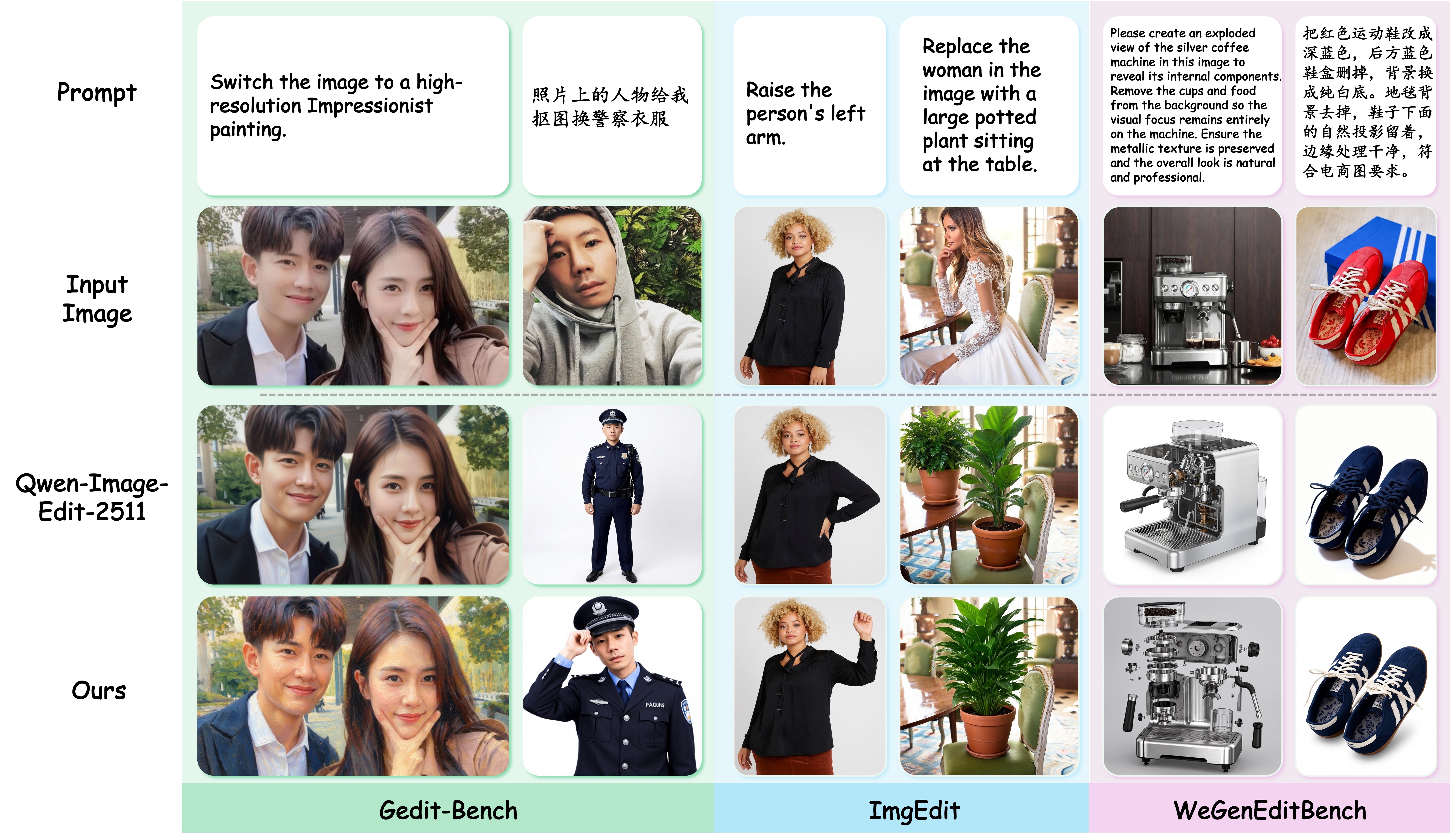}
\caption{Qualitative editing comparisons of Qwen-Image-Edit-2511 and its \studio{} adapters on GEdit, ImgEdit, and \bench{}. }
\label{fig:evolution-qualitative}
\vspace{-3mm}
\end{figure}

\paragraph{From diagnostic evidence to refined editing behavior.}
Figure~\ref{fig:evolution-qualitative} and Figures~\ref{fig:additional-generation}--\ref{fig:additional-flux-editing} in the appendix illustrate how targeted adaptation sharpens operation selection and edit-boundary control. Adapted Qwen raises the person's left arm instead of placing the hand at the hip, generates an exploded view of the coffee machine instead of a cutaway, and preserves white shoelaces and natural shadows while recoloring the shoes. These examples highlight the value of coupling intended transformations with explicit preservation boundaries in data prescriptions: the synthesis pipeline can encode both what should change and what should remain, providing targeted supervision for local behavioral errors. For the stronger Qwen backbone, the small gain on the full ImgEdit benchmark (+0.002) contrasts with the larger gain on its \single{} targets, motivating closer attention to the requirements captured by the training experience. The equal-budget web-retrieval ablation on WeGenBench (Appendix~\ref{app:web-search-ablation}, Table~\ref{tab:web-search-ablation}) further reveals concept-dependent retrieval effects. The small full-benchmark gain of 0.021 masks opposing category-level changes: retrieval improves counterintuitive concepts (+0.250), icons and symbols (+0.320), and intellectual-property concepts (+0.165), but reduces performance on animals by 0.292. One possible explanation is that retrieval contributes useful coverage for underrepresented concepts while introducing less useful variation for categories already well represented in the model's prior. The differing category responses motivate selective retrieval guided by diagnostic needs.

\paragraph{Backbone-dependent adaptation and cross-dimensional trade-offs.}
Table~\ref{tab:evolution} and Table~\ref{tab:evolution-full} reveal how targeted experience affects different backbones and capabilities. FLUX.2 [dev], which starts from lower scores on its selected single-image categories, improves across all four dimensions, including DS (+2.571). In contrast, Qwen's single-image gains concentrate in EG (+0.461) and VQ (+0.605), while PR remains stable and DS declines by 0.128. On its selected GEdit categories, PQ improves by 0.109 while SC declines by 0.343; full-benchmark Overall decreases by 0.190. This specialization pattern shows that stronger execution of salient transformations and better rendering do not automatically satisfy domain-specific rules or semantic requirements. Such trade-offs after a single adaptation round motivate re-evaluation and the design of multi-round RSI-oriented feedback loops. Once dimension-specific checklists reveal the updated capability profile, subsequent prescriptions can strengthen operation supervision for execution errors, tighten preservation constraints for unintended changes, and add specialized examples for domain-rule failures. Further experience acquisition can thereby respond to the model's remaining weaknesses.

\FloatBarrier
\section{Conclusion}
\label{sec:conclusion}

We introduced \bench{} and \studio{} to connect fine-grained image editing evaluation with targeted model improvement. Composite editing requests require feedback that distinguishes visual plausibility from complete requirement fulfillment and identifies what a model should learn next. Atomic checklists shared across models expose unmet local requirements, while the multi-agent harness translates failure diagnosis into targeted training data and supervised updates. In our evaluation, GPT-Image-2 leads single-image editing at 4.076 Overall, while Seedream-5.0-Pro leads text-intensive editing at 4.291. On their report-selected Top-5 single-image targets, FLUX.2 [dev] improves Overall from 1.083 to 3.517, and Qwen-Image-Edit-2511 gains 13.2\%. These targeted gains and cross-dimensional trade-offs demonstrate the value of interpretable, fine-grained evaluation as actionable training feedback. Our framework thus supports targeted experience acquisition and feedback-driven capability improvement within RSI-oriented learning loops. Extending this connection to a broader range of editing scenarios can guide the development of more reliable and adaptable image models.

\bibliography{references}
\bibliographystyle{plainnat}
\clearpage
\appendix
\begin{center}
{\Large\bfseries Supplementary Material}
\end{center}
\subsection*{Contents of the Appendix}
\newcommand{\appcontentsline}[3]{\noindent\hspace*{#1}\hyperref[#2]{#3}\nobreak\dotfill\nobreak\hyperref[#2]{\pageref*{#2}}\par}
\begingroup
\hypersetup{pdfborder={0 0 0}}
\small
\setlength{\parskip}{1pt}
\appcontentsline{0pt}{app:benchmark-details}{\textbf{\ref*{app:benchmark-details}\quad Benchmark Construction and Coverage}}
\appcontentsline{12pt}{app:benchmark-statistics}{\ref*{app:benchmark-statistics}\quad Application-Grounded Taxonomy}
\appcontentsline{12pt}{app:curation}{\ref*{app:curation}\quad Source-Grounded Curation}
\appcontentsline{12pt}{app:instruction-variants}{\ref*{app:instruction-variants}\quad Instruction and Record Consistency}
\appcontentsline{12pt}{app:benchmark-comparison}{\ref*{app:benchmark-comparison}\quad Coverage and Evaluation Resolution}
\appcontentsline{12pt}{app:primary-task-analysis}{\ref*{app:primary-task-analysis}\quad Task-Specific Strengths and Bottlenecks}
\vspace{3pt}
\appcontentsline{0pt}{app:evaluation-details}{\textbf{\ref*{app:evaluation-details}\quad Evaluation Protocol and Diagnostic Evidence}}
\appcontentsline{12pt}{app:evaluation}{\ref*{app:evaluation}\quad Shared Requirements and Item Judgments}
\appcontentsline{12pt}{app:scoring-details}{\ref*{app:scoring-details}\quad Scoring, Applicability, and Aggregation}
\appcontentsline{12pt}{app:human-audit}{\ref*{app:human-audit}\quad Human Audit and Shared Revisions}
\appcontentsline{12pt}{app:diagnostic-analysis}{\ref*{app:diagnostic-analysis}\quad Diagnostic Scope and Statistical Conventions}
\vspace{3pt}
\appcontentsline{0pt}{app:agents}{\textbf{\ref*{app:agents}\quad Agentic Diagnosis and Data Construction}}
\appcontentsline{12pt}{app:studio-implementation}{\ref*{app:studio-implementation}\quad Harness and Resource Contracts}
\appcontentsline{12pt}{app:diagnosis-agent}{\ref*{app:diagnosis-agent}\quad Diagnosis Agent: Evidence to Data Prescriptions}
\appcontentsline{12pt}{app:construction-agent}{\ref*{app:construction-agent}\quad Construction Agent: Retrieval-Grounded Task Synthesis}
\appcontentsline{12pt}{app:construction-boundary}{\ref*{app:construction-boundary}\quad Task-Level Feedback and Instance Boundaries}
\vspace{3pt}
\appcontentsline{0pt}{app:experimental-details}{\textbf{\ref*{app:experimental-details}\quad Experimental Protocol and Extended Analysis}}
\appcontentsline{12pt}{app:experimental-settings}{\ref*{app:experimental-settings}\quad Shared Training Settings and Evaluation Scope}
\appcontentsline{12pt}{sec:verification}{\ref*{sec:verification}\quad Updates and Evaluation Records}
\appcontentsline{12pt}{app:evolution-results}{\ref*{app:evolution-results}\quad Targeted Gains and Broader Behavior}
\appcontentsline{12pt}{app:web-search-ablation}{\ref*{app:web-search-ablation}\quad Web Search and Concept-Specific Effects}
\vspace{3pt}
\appcontentsline{0pt}{app:extended-tables}{\textbf{\ref*{app:extended-tables}\quad Supporting Results and Complete Taxonomy}}
\appcontentsline{12pt}{app:comparison-tables}{\ref*{app:comparison-tables}\quad Quantitative Comparison and Diagnostic Tables}
\appcontentsline{12pt}{app:primary-task-tables}{\ref*{app:primary-task-tables}\quad Primary-Task Quantitative Comparisons}
\appcontentsline{12pt}{app:taxonomy}{\ref*{app:taxonomy}\quad Complete Task Taxonomy}
\vspace{3pt}
\appcontentsline{0pt}{app:case-studies}{\textbf{\ref*{app:case-studies}\quad Qualitative Case Studies}}
\appcontentsline{12pt}{app:cases-single}{\ref*{app:cases-single}\quad \single{}}
\appcontentsline{12pt}{app:cases-text}{\ref*{app:cases-text}\quad \textedit{}}
\appcontentsline{12pt}{app:cases-multi}{\ref*{app:cases-multi}\quad \multi{}}
\appcontentsline{12pt}{app:evolution-qualitative-cases}{\ref*{app:evolution-qualitative-cases}\quad Qualitative Evidence of Model Adaptation}
\vspace{3pt}
\endgroup
\hypersetup{pdfborder={0 0 1}}
\clearpage

\section{Benchmark Construction and Coverage}
\label{app:benchmark-details}\label{app:protocol}
\subsection{Application-Grounded Taxonomy}
\label{app:benchmark-statistics}
\bench{} contains 2,740 instances: 1,008 single-image, 1,000 text-intensive, and 732 multi-reference tasks. Over 16,000 manually collected application requests and examples inform the taxonomy; automated construction and human refinement produce the evaluation records. The subsets contain 8/5/6 first-level families and 65/35/48 second-level types. The 148 types count subset--label pairs, not disjoint operations. Single-image and text-intensive instances may carry multiple labels; multi-reference instances have one primary label and two to four ordered inputs (382/185/165 cases). Table~\ref{tab:benchmark-statistics} and Appendix~\ref{app:taxonomy} give counts and definitions.
Instruction lengths use \texttt{tiktoken==0.12.0}, \texttt{o200k\_base}, and \texttt{encode\_ordinary} on the stored instruction without translation, trimming, or chat templates. The subset means are 79.2/124.6/146.3 tokens. Figure~\ref{fig:benchmark-statistics} additionally reports task-label multiplicity, input counts, and checklist sizes; multi-reference primary labels are excluded from the multiplicity comparison.

Tables~\ref{tab:taxonomy-single}--\ref{tab:taxonomy-multi} retain all 148 task types and their definitions. Family counts use original first-level annotations; task rows give operational definitions. Single/text records have 2,154/1,468 memberships, while all 732 multi-reference records have one primary label. Overlap denotes coverage, not extra instances. Single-image record 1006 carries an additional aesthetic-family annotation, explaining 320 original family members versus 319 reconstructed from child labels.

\subsection{Source-Grounded Curation}
\label{app:curation}\label{app:method-figures}
\begin{figure}[!t]
\centering
\includegraphics[width=\linewidth]{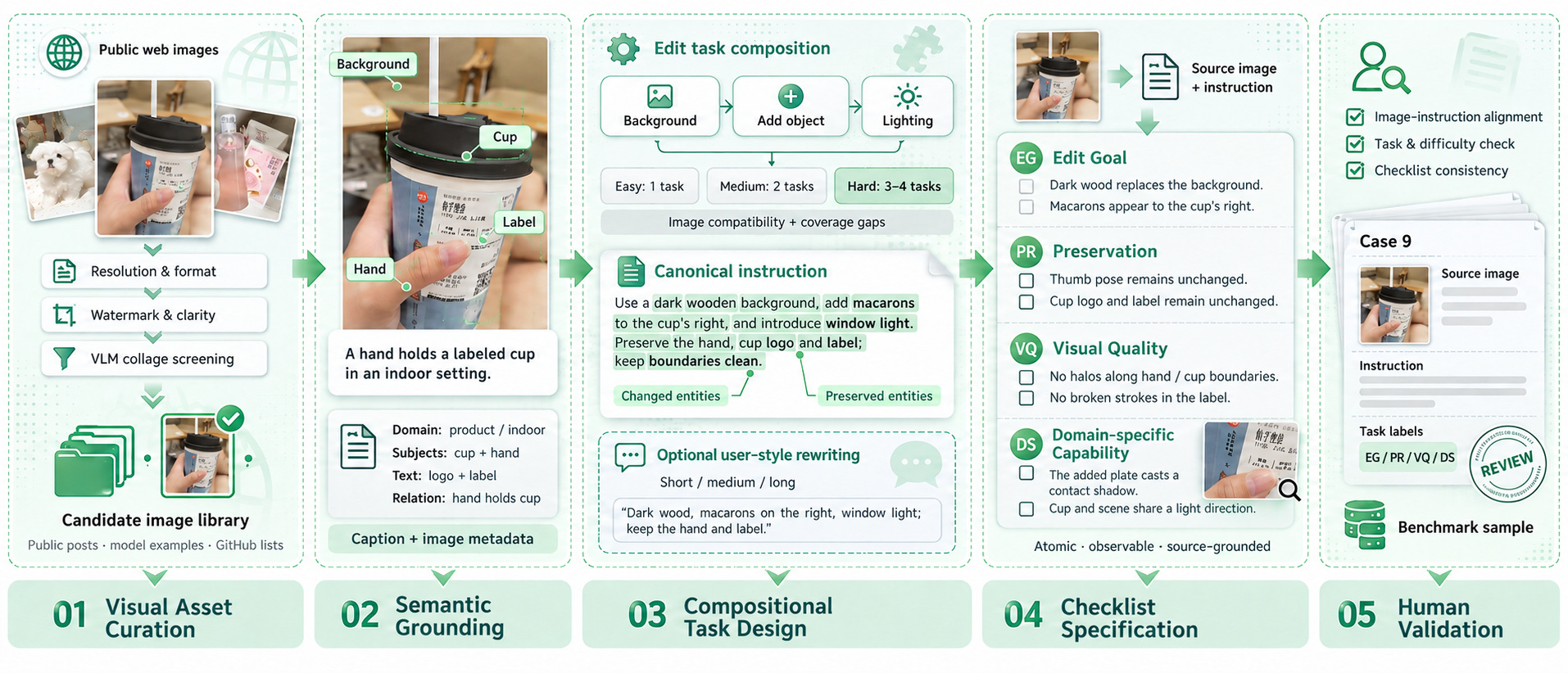}
\caption{Benchmark curation from visual assets to human-validated instructions and checklists. }
\label{fig:benchmark-pipeline}
\end{figure}

Figure~\ref{fig:benchmark-pipeline} shows five stages. \textbf{Visual Asset Curation} filters public images for format, resolution, clarity, and usable subjects or text. \textbf{Semantic Grounding} describes visible entities, OCR content, spatial relations, and reference roles. \textbf{Compositional Task Design} combines compatible labels with preservation, layout, and reference conditions, skipping unsupported targets. \textbf{Checklist Specification} converts the selected instruction and sources into observable requirements before output scoring. \textbf{Human Validation} checks feasibility, labels, image--instruction alignment, and checklist consistency; ambiguous or contradictory requirements are revised.
\subsection{Instruction and Record Consistency}
\label{app:instruction-variants}\label{app:record-schema}
Single-image and text-intensive generation uses proposal bands of one, two, or three to four task labels; text tasks may add at most one/two visual-edit labels in the medium/high bands. Short, medium, and long expressions can share a canonical intention, but only the selected expression is evaluated: unselected variants cannot add checklist obligations. Exports retain IDs, ordered sources, selected instructions, labels, checklists, and provenance. Text records separate the two sides; reference checks identify their assigned source.

\subsection{Coverage and Evaluation Resolution}
\label{app:benchmark-comparison}\label{app:protocol-comparison}
Tables~\ref{tab:benchmark-comparison} and~\ref{tab:benchmark-protocol-comparison} separate coverage from retained judgment granularity. GEdit/GEditBench v2 collect practical requests, ImgEdit includes multi-turn interaction, and Complex-Edit/ComplexBench-Edit examine composition and dependencies \citep{step1x,geditv2,imgedit,complexedit,complexbenchedit}. UnicBench adds instance-specific reasoning points, while REDEdit, RISEBench, and UniREditBench specialize text or reasoning assessment \citep{unicedit,firered,risebench,unireditbench}. \bench{} combines single-image, detailed text/layout, and multi-reference tasks with shared, sample-specific checks. Even one instruction can yield separate attribute, preservation, location, and source-binding judgments. This improves diagnostic resolution and auditability.

\paragraph{Comparison conventions.}
Inputs exclude judge-only ground-truth images, and taxonomies retain their native granularity. ImgEdit statistics use its 737-case Basic release, excluding UGE and multi-turn sequences. Token ranges use \texttt{o200k\_base}: Chinese for GEdit, UnicBench, and ours; English for ImgEdit. Dashes indicate unmeasured ranges. UnicBench supplies instance-specific reasoning points for 250 of 1,100 cases. Multi-dimensional evaluation includes non-VLM metrics.

\subsection{Task-Specific Strengths and Bottlenecks}
\label{app:primary-task-analysis}
Tables~\ref{tab:wegen-single-primary-local}--\ref{tab:wegen-multi-primary-aesthetic} report all 19 task families. Single/text categories overlap; averages use valid, applicable scores and per-case Overall. Valid-case sets differ across models, so small gaps are descriptive. Bold and underline rank the displayed scores across all models, including ties.

\paragraph{Competitive preservation does not imply complete execution.}
HunyuanImage-3.0-Instruct-Distil leads PR in six of eight single-image families. Its CoT variant ranks second overall on restoration (4.070), above Seedream-5.0-Pro (4.044); the base variant nearly matches Nano-Banana-Pro on cultural editing (3.957 versus 3.965). Yet on action/interaction editing, its leading PR of 3.975 accompanies EG of only 3.640, versus GPT-Image-2's 4.376. Open models can thus approach closed models on specific tasks or preservation, while still struggling to execute the requested transformation.

\paragraph{Text bottlenecks depend on the application.}
On posters/layout, JoyAI-Image-Edit nearly matches GPT-Image-2 in image-side EG (4.552 versus 4.567), but trails in text-side EG (3.534 versus 4.144). Strong visual editing therefore does not establish textual correctness. Hunyuan's CoT variant reaches 4.084 Overall on social-media editing, 0.143 below GPT-Image-2. For interfaces/documents/charts, however, the strongest open model, SenseNova-U1-8B-MoT-CoT, reaches only 3.724 versus Seedream's 4.392. The separate task and side scores locate this gap more precisely than a single text-editing average.

\paragraph{Multi-reference gaps are especially pronounced for text.}
Hunyuan leads the evaluated open models in all six multi-reference families, yet its Overall gap to the strongest closed model ranges from 0.580 on pets to 1.112 on text. Multi-reference text is its lowest-scoring family (2.878), with EG at 2.700; FireRed and Qwen-Image-Edit-2511 fall below 1.9 EG. These results identify reference-conditioned text editing as a shared weakness.

\section{Evaluation Protocol and Diagnostic Evidence}
\label{app:evaluation-details}
\subsection{Shared Requirements and Item Judgments}
\label{app:evaluation}
All three \bench{} subsets use Gemini 3.1 Pro Preview (\texttt{gemini-3.1-pro-preview}) as the judge. Checklists are generated from the selected instruction and source images before scoring and shared by all compared models within an evaluation version. Each item identifies an observable target, attribute, location, or reference assignment under EG/PR/VQ/DS. The judge receives the instruction, ordered sources, output, and checklist, and records a deduction $\delta_{i,j}\in[0,5]$, score $5-\delta_{i,j}$, and visual rationale for each item before assigning dimensional scores. Dimensional scores follow their rubrics and need not equal item means. Literal text and source indices remain explicit; invented requirements and redundant restatements are excluded.

\paragraph{Edit Goal (EG).}
EG measures whether each requested transformation is applied to the intended target with the specified attribute, quantity, and spatial relation. It covers object operations, requested text changes, and reference-conditioned composition. For a request to turn both boots and an ornament into silver metal, changing only the boots leaves an editing goal unmet.

\paragraph{Preservation (PR).}
PR measures whether content and attributes outside the requested changes remain faithful to the source. Checks cover protected identities, object geometry, existing text, layout, and background, as applicable. In multi-reference editing, PR checks the attributes assigned to each source. Preservation is judged relative to the instruction; a requested change is not itself a preservation error.

\paragraph{Visual Quality (VQ).}
VQ measures rendering quality and visual coherence, including clarity, structural integrity, natural boundaries, and freedom from blur, artifacts, or implausible seams. Text-side checks examine legibility and glyph integrity; image-side checks assess the surrounding visual content. A visually polished output may therefore receive high VQ while failing EG or PR requirements.

\paragraph{Domain-specific Capability (DS).}
DS measures correctness under task-specific criteria beyond generic execution, preservation, and rendering quality. Applicable checks address material behavior, physical or geometric relations, typographic hierarchy, or cultural semantics. In the metal-editing example, consistent metallic tones and reflections across both targets form a joint requirement: making one object reflective does not establish material consistency across the pair.

\subsection{Scoring, Applicability, and Aggregation}
\label{app:scoring-details}
For single-image instance $i$, $B_i=0.6S_{\mathrm{EG},i}+0.2S_{\mathrm{PR},i}+0.2S_{\mathrm{VQ},i}$ and $p_{i,t}=(5-S_{\mathrm{DS},i,t})/4$. Equation~\ref{eq:single-score} subtracts the mean applicable-task penalty and clips to $[0,5]$; task-conditioned summaries use the corresponding penalty. Empty DS sets incur no penalty and have no DS score. Empty EG/PR/VQ checklists receive 5; omitted dimension scores in compact responses fall back to item means. Missing required fields or mismatched item counts invalidate the response. Stored scores retain four decimals.
Text editing evaluates EG/PR/VQ/DS separately on the text and image sides, combining their scores with weights 0.75/0.25 only when image-side requirements exist; otherwise the text-side score is final. Multi-reference EG/PR/VQ correspond to native IF/RF/VQ and use weights 0.4/0.4/0.2 (Equation~\ref{eq:text-score}); no separate DS is annotated. Subset Overall averages valid instance scores once per instance, including unsuccessful edits. Inapplicable criteria and invalid judge responses are excluded from their respective denominators; task-label overlap does not duplicate instances in Overall. The weights define an operational rubric. High aggregate scores alongside low item scores demonstrate retained diagnostic detail.

\subsection{Human Audit and Shared Revisions}
\label{app:human-audit}
Three trained experts randomly audit 10\% of cases in each benchmark, checking whether requirements follow from the shown inputs and deductions are supported by outputs. Problematic checklists are corrected first; every compared model is re-evaluated on affected cases using the same revised checklist. Each version fixes the judge, prompt, input processing, checklist, and invalid-response policy. A score--item discrepancy alone is not proof of model failure: numerical aggregation, unsupported requirements, and judge inconsistency must be distinguished. The audit corrects identified problems but does not independently validate every judgment or quantify accuracy gains over holistic evaluation. Inter-rater agreement and error rates cannot be inferred from this corrective audit.

\subsection{Diagnostic Scope and Statistical Conventions}
\label{app:diagnostic-analysis}
The audit covers 15,214 model--task and 194,836 item records over 2,740 tasks; 262 inapplicable items are excluded. Single/text analyses include Qwen-Image-Edit-2511, FireRed-Image-Edit-1.1, HunyuanImage-3.0-Instruct-Distil-CoT, GPT-Image-2, Seedream-5.0-Pro, and JoyAI-Image-Edit. Multi-reference comparisons use Qwen, FireRed, GPT, and Seedream. High-scoring outputs have Overall $\geq4$; severe items score $\leq2$. A side passes only if every applicable item scores $\geq4$; empty image-side lists are not successes. Each output contributes once per diagnostic event.
For text editing, the six models are compared on the same 920 tasks, of which 280 have requirements on both the text and image sides. Proportion intervals use 2,000 source-image-cluster bootstrap replicates; multi-reference clustering uses ordered reference tuples, with first-source sensitivity checks. Diagnostic analyses use the item-level judgments, dimensional scores, and subset-specific Overall defined in Section~\ref{sec:metrics}. Main leaderboard and adaptation tables report point estimates. Tables~\ref{tab:diagnostic-high-scores} and~\ref{tab:diagnostic-joint-sides} report the diagnostic outcomes.

\paragraph{Recorded diagnostic outcomes.}\label{app:diagnostic-results}\label{app:aggregation-analysis}\label{app:joint-analysis}
Table~\ref{tab:diagnostic-high-scores} conditions on each model's high-scoring outputs rather than a shared ranking set. Within the single-image Overall $\geq4$ group, fixing 16 applicable items gives 50/104 severe-item cases (48.1\%, 95\% CI 38.4--57.7\%). Exact multi-reference score vector $(5,5,3,4.6)$ maps to both severe-item and no-severe-item outcomes: 44 with and 64 without severe items for GPT, and 30 with and 50 without for Seedream. At item threshold $\leq1$, neither has events among multi-reference outputs with Overall $\geq4.5$. Table~\ref{tab:diagnostic-joint-sides} uses the same 280 dual-side tasks.

\section{Agentic Diagnosis and Data Construction}
\label{app:agents}
\subsection{Harness and Resource Contracts}
\label{app:studio-implementation}\label{app:tool-tables}
The canonical registry exposes 68 tools (aliases excluded): benchmarking 12, analysis 17, construction 10, training 6, evolution 4, discovery 1, and system operations 18. \texttt{gemini-3-flash-preview} serves as the backbone LLM for the Diagnosis and Construction Agents, supporting planning and tool use. The harness supports four concurrent child agents and one delegation level by default. Each child has isolated context and a tool allowlist. Schema validation and committed resource references connect evaluation records, reports, datasets, training jobs, and checkpoints. Visual evidence is stored separately from original judgments and can be reused through content fingerprints.

\paragraph{Compatibility with native benchmark feedback.}
\label{app:feedback-compatibility}
\studio{} accepts benchmark-specific evaluation records through adapters, with category scores supporting task prioritization and native rationales or checklist items providing additional evidence when available. The shared interface preserves each benchmark's metric definitions and score direction; it does not require external benchmarks to adopt EG/PR/VQ/DS or our checklist format. This separates compatibility with an evaluation protocol from the granularity of the evidence it provides.

\subsection{Diagnosis Agent: Evidence to Data Prescriptions}
\label{app:diagnosis-agent}
\begin{figure}[!t]
\centering
\includegraphics[width=\linewidth]{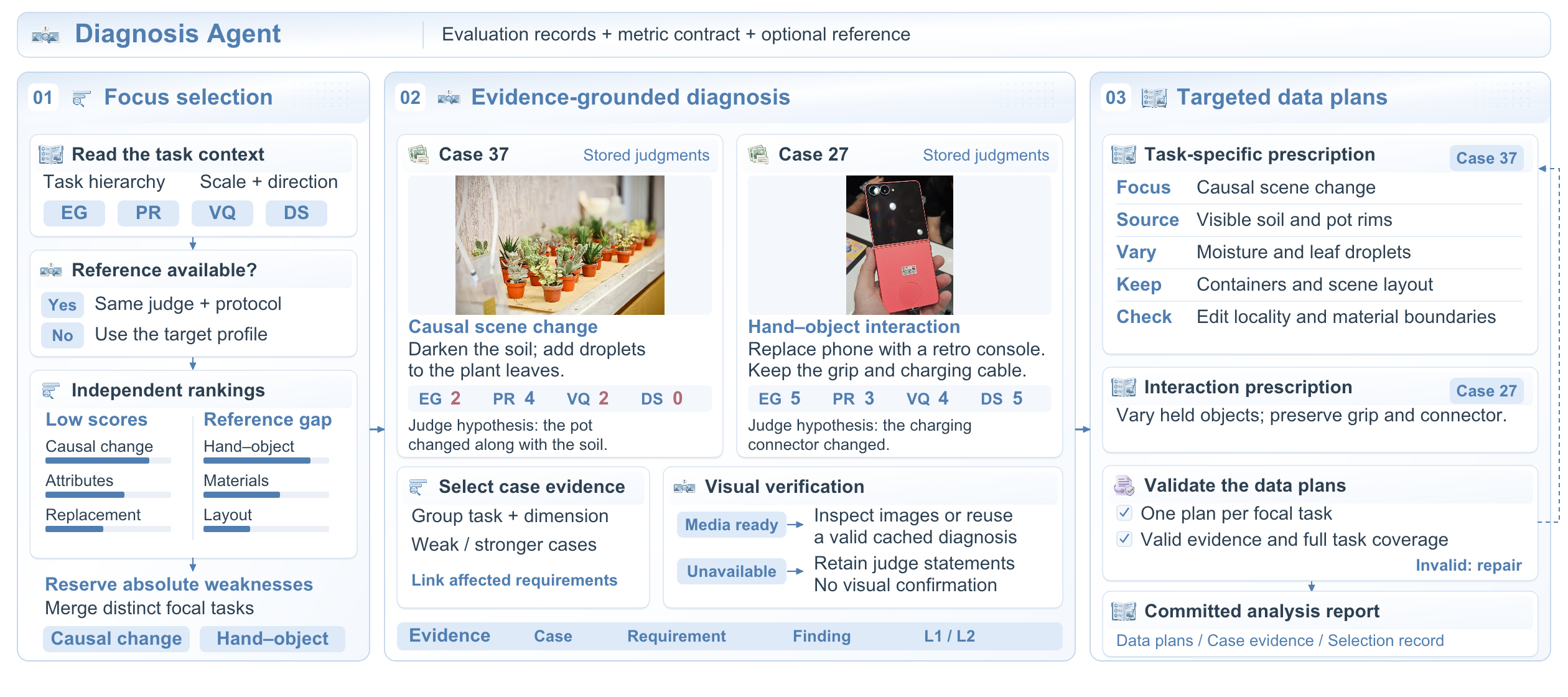}
\caption{Schematic of the Diagnosis Agent: category selection, visual review, failure grouping, and data prescriptions. }
\label{fig:deep-analysis-agent}
\end{figure}

Figure~\ref{fig:deep-analysis-agent} summarizes diagnosis. Adapters normalize category axes, score scales, and score direction. With a compatible anchor, selection allocates approximately 60\% of slots to relative deficits and 40\% to absolute weaknesses, with at least one absolute slot. Duplicate selections are removed and reciprocal-rank ordering fills remaining slots; absent relative deficits, selection uses absolute scores. This is slot allocation, not score interpolation. Selected categories bound case review: weak/successful contrasts distinguish recurring symptoms from isolated outputs. Instructions, sources, outputs, and available reasons remain linked to case IDs.

\paragraph{Diagnosis at the available feedback granularity.}
In our experiments, GEdit and ImgEdit guide task selection through category scores; WeGenBench additionally contributes its own rationales; \bench{} provides requirement-level scores and reasons. Category scores identify where to allocate data, while available explanations and visual review help specify what should change or remain intact. All routes produce task-level prescriptions, allowing the framework to operate with coarse feedback and incorporate finer evidence when supplied.
Failure groups retain affected IDs and score losses; visual inspection checks the reported defect. A data prescription names the capability to reinforce and supported source conditions, constraints, and acceptance criteria; reference and coverage checks precede commitment. We review 3 cases per category across 10 categories, with up to 30 visual reviews and 31 LLM calls. Model evolution selects $K=5$ task categories.
\subsection{Construction Agent: Retrieval-Grounded Task Synthesis}
\label{app:construction-agent}
\begin{figure}[!htbp]
\centering
\includegraphics[width=\linewidth]{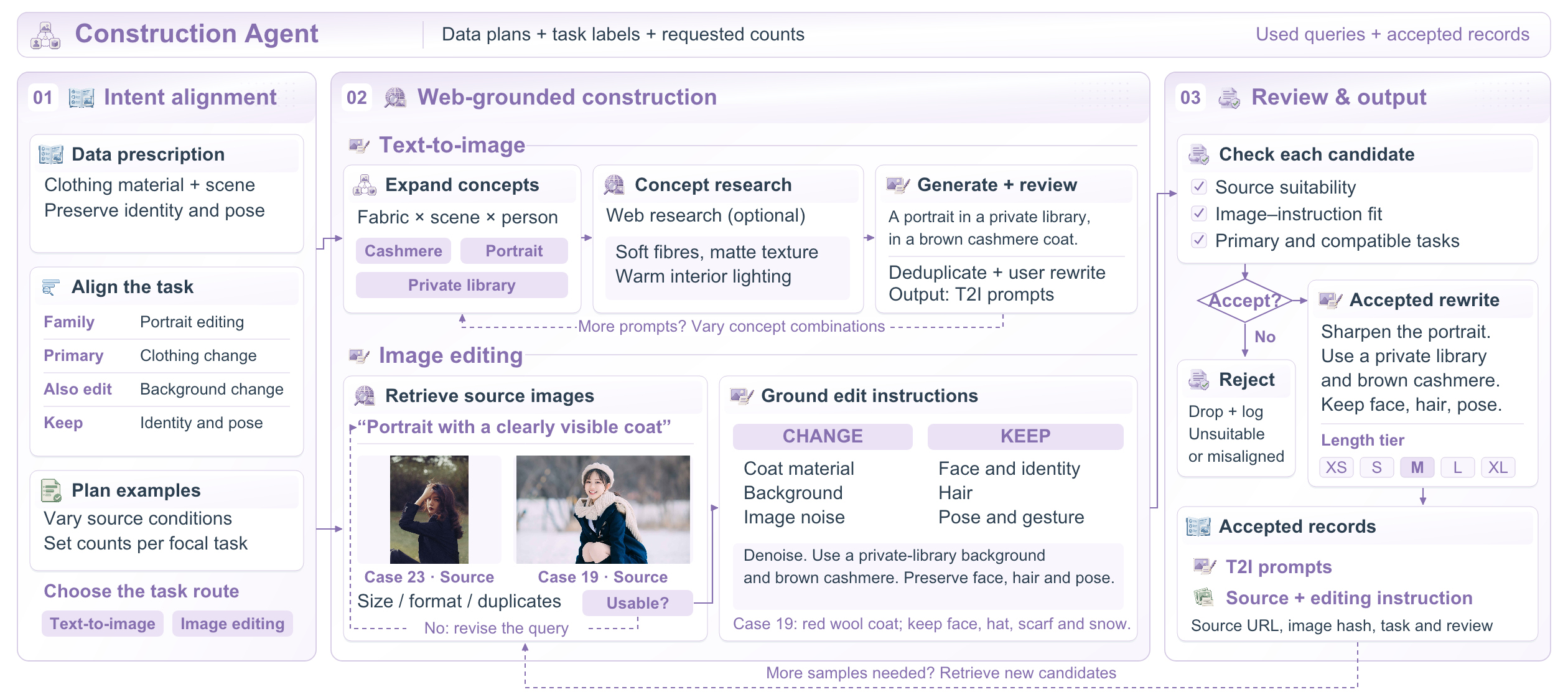}
\caption{Construction Agent: task alignment, concept research or image retrieval, grounded instruction generation, and review.  Accepted records retain source and instruction provenance; target supervision and training follow downstream.}
\label{fig:instruction-construction-agent}
\end{figure}

Figure~\ref{fig:instruction-construction-agent} shows task alignment, retrieval, grounded instruction generation, and review. Selected labels map to the reusable taxonomy and optional category notes; the instruction budget is allocated across these tasks. Web search is provided by Serper\footnote{\url{https://serper.dev/}}. Editing queries seek suitable unedited content, varying subjects and contexts. Acquired images pass format/resolution checks and registry-level exact-duplicate handling, with optional perceptual filtering. The generator receives an accepted source, metadata, task, and notes, then specifies feasible changes, protected content, and spatial conditions. Compatible secondary operations enrich requests; unsuitable sources are skipped.
VLM review checks category alignment and feasibility before rewriting canonical instructions into user-style requests. Accepted records retain both versions, source hashes, labels, filtering outcomes, and report/dataset identifiers. The downstream supervision pipeline supplies targets. For text-to-image generation, keyword expansion and optional web-text research provide diverse concepts and defining visual attributes without an editing source. The implemented adaptation branches cover generation and single-image editing.

\paragraph{A shared construction interface.}
Across feedback formats, the Construction Agent receives selected task labels, requested counts, and optional category notes. Score-guided plans specify task priorities; richer evidence can refine source conditions and preservation constraints. Both use the same retrieval, grounded generation, and review workflow. Thus, external benchmarks can guide data construction without supplying our sample-specific checklists or their own instruction-generation templates.
\subsection{Task-Level Feedback and Instance Boundaries}
\label{app:construction-boundary}
Cases justify capability selection within the diagnosis report; construction consumes task labels and optional category notes. Per-image calls do not receive evaluated input--output pairs as demonstrations or import external benchmarks' generation templates. Retrieved content determines the new scene, while feedback sets capability priorities. This is benchmark-informed task-level adaptation.

\section{Experimental Protocol and Extended Analysis}
\label{app:experimental-details}
\subsection{Shared Training Settings and Evaluation Scope}
\label{app:experimental-settings}\label{app:checkpoint-mapping}
The benchmark comparison evaluates 15 configurations on single-image/text editing and six on multi-reference editing. Evolution uses Qwen-Image-2512 for generation, Qwen-Image-Edit-2511 for editing, and FLUX.2 [dev] for both, with a separate rank-16 LoRA adapter per backbone--benchmark pair. The main evolution experiments follow a shared training protocol and each use 25,000 training examples, with 5,000 per selected task for $K=5$. The per-device training batch size is 1. The feedback adapters and shared construction interface in Appendix~\ref{app:feedback-compatibility} support category scores from GEdit/ImgEdit, native rationales from WeGenBench, and checklist evidence from \bench{}. These experiments test the utility of the framework across native feedback formats.
Report-selected evaluation counts for Qwen/FLUX.2 are 277/326 (WeGenBench), 540/473 (GEdit), 433/425 (ImgEdit), and 82/78 (single-image). Report-selected evaluations use deduplicated category unions and differ by backbone; FLUX.2 WeGenBench comparisons use common-valid pairs. GEdit averages $O_i=\sqrt{\mathrm{SC}_i\mathrm{PQ}_i}$ per case; ImgEdit's detail score D differs from our DS. Deltas use unrounded means.

\paragraph{Hardware and external services.}
All local experiments, including model training and inference, run on a 32-card Cambricon MLU590 system. Closed-source image generation, API-based VLM evaluation, and web retrieval are accessed through external APIs; their provider-side hardware is outside this local configuration.

\subsection{Updates and Evaluation Records}
\label{sec:verification}\label{app:protocol-qualifications}
Constructed tasks are paired with target supervision into $D_t$; the backend updates the parent checkpoint and records the data revision and training recipe $\lambda_t$:
\begin{equation}
\widetilde{\theta}_{t+1}=\operatorname{Train}(\theta_t,D_t;\lambda_t).
\label{eq:update}
\end{equation}
Re-evaluation links the updated model to revised capability priorities:
\begin{equation}
\theta_t\longrightarrow\Phi_t\longrightarrow\mathcal{T}_t\longrightarrow\mathcal{X}_t\longrightarrow D_t\longrightarrow\theta_{t+1}\longrightarrow\Phi_{t+1}.
\label{eq:evolution-loop}
\end{equation}
Run records retain parentage, supervision, filtering, inference settings, and evaluation versions. Validation supports target and retention checks.
FLUX.2 WeGenBench uses Qwen3.8-Flash-Next with the deduction prompt on 326 target and 995 full-benchmark common-valid pairs. FLUX.2 GEdit uses GPT-4.1 VIEScore~\citep{viescore} on 473 target and 1,210 full-benchmark common-valid pairs. The \bench{} single-image comparison shares cases and judge identity.

\subsection{Targeted Gains and Broader Behavior}
\label{app:evolution-results}\label{app:targeted-evolution}\label{app:full-evolution}\label{app:gedit-category-results}\label{app:cases-evolution}
Table~\ref{tab:evolution-full} separates targeted gains from full-benchmark changes: Qwen gains 0.670 WeGenBench F but loses 0.190 GEdit Overall; FLUX.2 gains 0.124 and 3.009, respectively. All five FLUX.2 GEdit targets improve, with larger SC than PQ gains (Table~\ref{tab:flux-gedit-categories}); motion change has the largest Overall increase (+5.295). Selected Qwen examples in Figure~\ref{fig:evolution-qualitative} show better operation choice and local preservation.
On WeGenBench, FLUX.2 improves Final from 7.767 to 7.997 on the report-selected Top-5 set ($\Delta=+0.230$) and from 8.352 to 8.475 on the full-benchmark set ($\Delta=+0.124$), using 326 and 995 common-valid pairs, respectively (Table~\ref{tab:evolution-sensitivity}).

\subsection{Web Search and Concept-Specific Effects}
\label{app:web-search-ablation}
Qwen-Image-2512 search-enabled and search-disabled adapters each train on 25,000 examples for 2,346 steps using 32 Cambricon MLU590 accelerators and a per-device batch size of 1. Both receive one Qwen3.8-Flash-Next evaluation pass. Table~\ref{tab:web-search-ablation} keeps the original 277-case Top-5 aggregate while displaying four category columns. Search changes target/full F by +0.018/+0.021, improving counterintuitive concepts, symbols, and IP but reducing the animal score. With matched training-data size and step schedules, these heterogeneous changes indicate concept-dependent retrieval effects.

\section{Supporting Results and Complete Taxonomy}
\label{app:extended-tables}
\subsection{Quantitative Comparison and Diagnostic Tables}
\label{app:comparison-tables}
\begin{table}[!htbp]
\centering\small
\caption{Statistics of the three source files. Length is measured in tokens using \texttt{o200k\_base}. Types are subset-qualified.}
\label{tab:benchmark-statistics}
\begin{tabular}{lrrr}
\toprule
Statistic & Single-image & Text-intensive & Multi-reference\\
\midrule
Instances & 1,008 & 1,000 & 732\\
Task families & 8 & 5 & 6\\
Task types & 65 & 35 & 48\\
Mean instruction length & 79.2 & 124.6 & 146.3\\
Median instruction length & 60.5 & 126 & 116.5\\
Instruction length range & 12--255 & 10--344 & 30--439\\
Mean task labels & 2.14 & 1.47 & ---\\
Multi-label instances (\%) & 70.6 & 31.2 & ---\\
Mean input images & 1.00 & 1.00 & 2.70\\
\bottomrule
\end{tabular}
\end{table}

\begin{table}[!htbp]
\centering
\caption{Quantitative comparison on image-editing benchmarks: task coverage. Conventions are specified in Appendix~\ref{app:benchmark-comparison}.}
\label{tab:benchmark-comparison}
\begingroup
\fontsize{9pt}{11pt}\selectfont
\setlength{\tabcolsep}{3pt}
\renewcommand{\arraystretch}{1.12}
\begin{tabularx}{\linewidth}{@{}p{0.26\linewidth}p{0.18\linewidth}cYc@{}}
\toprule
\textbf{Benchmark} & \textbf{Taxonomy} & \textbf{Tokens} & \textbf{Task emphasis} & \textbf{Inputs} \\
\midrule
GEdit-Bench~\citep{step1x} & 11 categories & 2--38 & General editing & 1 \\
ImgEdit~\citep{imgedit} & 9 Basic types & 4--60 & Single/multi-turn & 1 \\
REDEdit~\citep{firered} & 15 categories & --- & General/text & 1 \\
RISEBench~\citep{risebench} & 4 families / 16 types & --- & Implicit reasoning & 1 \\
UnicBench~\citep{unicedit} & 4 families / 22 types & 6--42 & General/reasoning & 1 \\
UniREditBench~\citep{unireditbench} & 8 dimensions / 18 subtypes & --- & Real/game reasoning & 1 \\
Complex-Edit~\citep{complexedit} & 9 groups / 24 operations & --- & 1--8 operations & 1 \\
ComplexBench-Edit~\citep{complexbenchedit} & 3 dependency levels & --- & Parallel/chained edits & 1 \\
GEditBench v2~\citep{geditv2} & 22 tasks + open set & --- & Hybrid/open-set & 1 \\
\midrule
\textbf{WeGenEditBench} & \textbf{19 families / 148 types} & \textbf{10--439} & \textbf{Single/text/multi} & \textbf{1--4} \\
\bottomrule
\end{tabularx}
\endgroup
\end{table}

\begin{table}[!htbp]
\centering
\caption{Comparison of evaluation protocols and diagnostic outputs. Conventions are specified in Appendix~\ref{app:benchmark-comparison}.}
\label{tab:benchmark-protocol-comparison}
\begingroup
\fontsize{9pt}{11pt}\selectfont
\setlength{\tabcolsep}{3pt}
\renewcommand{\arraystretch}{1.12}
\begin{tabularx}{\linewidth}{@{}p{0.25\linewidth}c p{0.23\linewidth}Y@{}}
\toprule
\textbf{Benchmark} & \textbf{Multi-dim.} & \textbf{Verification criteria} & \textbf{Retained judgment unit} \\
\midrule
GEdit-Bench & Yes & General rubric & SC/PQ scores; explanations in VIEScore \\
ImgEdit & Yes & Task-specific rubrics & Dimension scores + brief reason \\
REDEdit & Yes & Task/text criteria; OCR & Dimension scores; text-error metrics \\
RISEBench & Yes & Reasoning guidelines & Reasoning, consistency, plausibility \\
UnicBench & Yes & Instance reasoning points & Dimension scores + reason \\
UniREditBench & Yes & Text + GT image references & Reference-aided dimension scores \\
Complex-Edit & Yes & Detailed IF/IP/PQ rubrics & Scores per evaluation dimension \\
ComplexBench-Edit & Yes & Edit-type criteria; chain rules & Per-instruction scores; region consistency \\
GEditBench v2 & Yes & IF/VQ/VC comparisons & Pairwise dimensional preferences \\
\midrule
\textbf{WeGenEditBench} & \textbf{Yes} & \textbf{Shared instance checklist} & \textbf{Per-requirement score, deduction, and rationale} \\
\bottomrule
\end{tabularx}
\endgroup
\end{table}

\begin{table}[!htbp]
\centering
\caption{GPT-Image-2 outputs with at least one recorded item score $\leq2$, conditional on the specified Overall threshold. Counts refer to model outputs; intervals are 95\% cluster-bootstrap intervals.}
\label{tab:diagnostic-high-scores}
\small
\setlength{\tabcolsep}{4pt}
\begin{tabular*}{\linewidth}{@{\extracolsep{\fill}}l|c|ccc@{}}
\toprule
\textbf{Subset} & \textbf{Overall} & \textbf{Count} & \textbf{Rate} & \textbf{95\% CI} \\
\midrule
Single-image & $\geq4$ & 315/644 & 48.9\% & [45.1, 52.8]\% \\
Text-intensive & $\geq4$ & 171/666 & 25.7\% & [22.4, 29.0]\% \\
Multi-reference & $\geq4$ & 119/503 & 23.7\% & [19.9, 27.2]\% \\
\midrule
Single-image & $\geq4.5$ & 101/323 & 31.3\% & [26.5, 36.6]\% \\
\bottomrule
\end{tabular*}
\end{table}

\begin{table}[!htbp]
\centering
\caption{Joint satisfaction of text-side and image-side requirements on 280 shared tasks. A side passes only when every applicable item scores at least 4. Counts within each row sum to 280.}
\label{tab:diagnostic-joint-sides}
\small
\setlength{\tabcolsep}{3pt}
\begin{tabular*}{\linewidth}{@{\extracolsep{\fill}}l|cccc@{}}
\toprule
\textbf{Model} & \textbf{Both sides} & \textbf{Text only} & \textbf{Image only} & \textbf{Neither} \\
\midrule
GPT-Image-2 & 70 (25.0\%) & 29 & 77 & 104 \\
Seedream-5.0-Pro & 76 (27.1\%) & 49 & 66 & 89 \\
\bottomrule
\end{tabular*}
\end{table}

\begin{table}[!htbp]
\centering
\caption{Full-benchmark results for the adapters in Table~\ref{tab:evolution}. FLUX.2 WeGenBench scores use 995 common-valid pairs.}
\label{tab:evolution-full}
\fontsize{8.5pt}{10.3pt}\selectfont
\setlength{\tabcolsep}{1.3pt}
\renewcommand{\arraystretch}{1.10}
\begin{tabular*}{\linewidth}{@{\extracolsep{\fill}}l*{8}{c}@{}}
\toprule
\textbf{Model} & \textbf{WeGenBench} & \multicolumn{3}{c}{\textbf{GEdit}} & \multicolumn{4}{c}{\textbf{ImgEdit}} \\
\cmidrule(lr){2-2}\cmidrule(lr){3-5}\cmidrule(lr){6-9}
& F $\uparrow$ & SC $\uparrow$ & PQ $\uparrow$ & O $\uparrow$ & I $\uparrow$ & Q $\uparrow$ & D $\uparrow$ & F $\uparrow$ \\
\midrule
Qwen-Image & 8.170 & 8.774 & 7.676 & 7.962 & 4.704 & 4.438 & 4.490 & 4.544 \\
\addlinespace[1.5pt]
\textbf{+ WeGenEvoStudio} & 8.840 & 8.338 & 7.597 & 7.772 & 4.688 & 4.474 & 4.476 & 4.546 \\
\addlinespace[1.5pt]
$\Delta$ & \evogain{+0.670} & \evoloss{-0.436} & \evoloss{-0.079} & \evoloss{-0.190} & \evoloss{-0.016} & \evogain{+0.035} & \evoloss{-0.014} & \evogain{+0.002} \\
\midrule
FLUX.2 [dev] & 8.352 & 4.388 & 7.621 & 4.395 & 2.323 & 3.124 & 3.207 & 2.885 \\
\addlinespace[1.5pt]
\textbf{+ WeGenEvoStudio} & 8.475 & 7.785 & 8.132 & 7.404 & 4.170 & 4.099 & 4.190 & 4.153 \\
\addlinespace[1.5pt]
$\Delta$ & \evogain{+0.124} & \evogain{+3.398} & \evogain{+0.511} & \evogain{+3.009} & \evogain{+1.846} & \evogain{+0.976} & \evogain{+0.984} & \evogain{+1.269} \\
\bottomrule
\end{tabular*}
\end{table}

\begin{table}[!htbp]
\centering
\caption{FLUX.2 [dev] results after 2,346 training steps on the five report-selected GEdit categories, evaluated with GPT-4.1 VIEScore. Scores refer to the adapted model, and $\Delta$ denotes the change from its paired baseline.}
\label{tab:flux-gedit-categories}
\small
\setlength{\tabcolsep}{3pt}
\renewcommand{\arraystretch}{1.1}
\begin{tabular*}{\linewidth}{@{\extracolsep{\fill}}l|*{3}{c}|*{3}{c}@{}}
\toprule
\textbf{Category} & SC & PQ & O & $\Delta$ SC & $\Delta$ PQ & $\Delta$ O \\
\midrule
\texttt{motion\_change} & 8.114 & 8.430 & 7.907 & \evogain{+5.747} & \evogain{+0.848} & \evogain{+5.295} \\
\texttt{subject-remove} & 7.272 & 8.316 & 6.911 & \evogain{+3.991} & \evogain{+0.351} & \evogain{+3.706} \\
\texttt{color\_alter} & 9.275 & 7.812 & 8.308 & \evogain{+5.463} & \evogain{+0.050} & \evogain{+4.371} \\
\texttt{material\_alter} & 7.625 & 7.600 & 7.002 & \evogain{+4.438} & \evogain{+0.575} & \evogain{+3.678} \\
\texttt{subject-replace} & 7.817 & 7.933 & 7.226 & \evogain{+3.658} & \evogain{+0.425} & \evogain{+3.173} \\
\bottomrule
\end{tabular*}
\end{table}

\begin{table}[!htbp]
\centering
\caption{FLUX.2 [dev] WeGenBench Final scores on report-selected Top-5 and full-benchmark common-valid pairs. Baseline and adapted checkpoints share the same evaluated cases.}
\label{tab:evolution-sensitivity}
\small
\setlength{\tabcolsep}{4pt}
\begin{tabular*}{\linewidth}{@{\extracolsep{\fill}}lrrrr@{}}
\toprule
\textbf{Scope} & \textbf{Pairs} & \textbf{Before} & \textbf{After} & {\normalsize$\Delta$} \\
\midrule
Report Top-5 & 326 & 7.767 & 7.997 & \evogain{+0.230} \\
\midrule
Full benchmark & 995 & 8.352 & 8.475 & \evogain{+0.124} \\
\bottomrule
\end{tabular*}
\end{table}

\begin{table}[!htbp]
\centering
\caption{Web-search comparison on WeGenBench using Qwen-Image-2512. All scores are F $\uparrow$ (1--10). $\Delta$ is search-enabled minus search-disabled. Top-5 includes all five selected categories (277 cases), although only four category columns are displayed.}
\label{tab:web-search-ablation}
\small
\setlength{\tabcolsep}{2pt}
\renewcommand{\arraystretch}{1.18}
\begin{tabularx}{\linewidth}{@{}p{0.20\linewidth}|*{4}{>{\centering\arraybackslash}X}|*{2}{>{\centering\arraybackslash}X}@{}}
\toprule
\textbf{Setting} & \shortstack{Counter-\\intuitive} & \shortstack{Icons /\\symbols} & Animals & IP & Top-5 & Full \\
\midrule
Baseline & 7.150 & 7.540 & 8.000 & 7.810 & 7.700 & 8.170 \\
Without web search & 8.450 & 8.440 & \textbf{8.903} & 8.506 & 8.650 & 8.819 \\
With web search & \textbf{8.700} & \textbf{8.760} & 8.611 & \textbf{8.671} & \textbf{8.668} & \textbf{8.840} \\
\midrule
$\Delta$ & \evogain{+0.250} & \evogain{+0.320} & \evoloss{-0.292} & \evogain{+0.165} & \evogain{+0.018} & \evogain{+0.021} \\
\bottomrule
\end{tabularx}
\end{table}

\clearpage
\subsection{Primary-Task Quantitative Comparisons}
\label{app:primary-task-tables}
\subsubsection{Single-Image Editing}
\label{app:primary-tables-single}
\begin{table}[!htbp]
\centering
\caption{Quantitative comparison on \single{}: Basic Local Editing. Bold and underline denote the best and second-best scores.}
\label{tab:wegen-single-primary-local}
\fontsize{8.5pt}{10pt}\selectfont
\renewcommand{\arraystretch}{1.0}
\setlength{\tabcolsep}{4.5pt}
\begin{tabularx}{\linewidth}{@{}l|*{4}{>{\centering\arraybackslash}X}|>{\centering\arraybackslash}X@{}}
\toprule
\textbf{Model Name} & EG$\uparrow$ & PR$\uparrow$ & VQ$\uparrow$ & DS$\uparrow$ & \textbf{Overall}$\uparrow$ \\
\midrule
\rowcolor{TableTint}\multicolumn{6}{c}{\itshape Closed-source Models} \\
GPT-Image-2 & \textbf{4.424} & 3.966 & \textbf{4.445} & \underline{4.225} & \textbf{4.143} \\
Nano-Banana-Pro & 4.317 & 3.640 & 4.402 & 4.056 & 3.963 \\
Qwen-Image-3.0 & 4.228 & 3.517 & 4.359 & 4.125 & 3.893 \\
Seedream-5.0-Pro & \underline{4.328} & 3.985 & \underline{4.424} & \textbf{4.251} & \underline{4.091} \\
\midrule
\rowcolor{TableTint}\multicolumn{6}{c}{\itshape Open-source Models} \\
HunyuanImage-3.0-Instruct-Distil & 4.101 & \textbf{4.147} & 4.271 & 4.002 & 3.895 \\
HunyuanImage-3.0-Instruct-Distil-CoT & 4.197 & \underline{4.030} & 4.239 & 4.042 & 3.933 \\
FireRed-Image-Edit-1.1 & 4.171 & 3.496 & 4.224 & 3.934 & 3.780 \\
JoyAI-Image-Edit & 4.118 & 3.520 & 4.172 & 3.869 & 3.727 \\
Qwen-Image-Edit-2511 & 4.043 & 3.495 & 4.101 & 3.761 & 3.636 \\
Boogu-Image-0.1-Edit & 4.000 & 3.626 & 4.094 & 3.763 & 3.635 \\
SenseNova-U1-8B-MoT-CoT & 4.071 & 3.610 & 3.943 & 3.632 & 3.614 \\
SenseNova-U1-8B-MoT & 4.147 & 3.317 & 3.948 & 3.723 & 3.622 \\
LongCat-Image-Edit & 3.938 & 3.371 & 4.011 & 3.670 & 3.513 \\
HiDream-O1-Image & 3.286 & 2.962 & 3.896 & 3.199 & 2.897 \\
FLUX.2 [dev] & 1.984 & 3.311 & 3.307 & 1.753 & 1.707 \\
\bottomrule
\end{tabularx}
\end{table}

\begin{table}[!htbp]
\centering
\caption{Quantitative comparison on \single{}: Aesthetic and Photographic Editing. Bold and underline denote the best and second-best scores.}
\label{tab:wegen-single-primary-aesthetic}
\fontsize{8.5pt}{10pt}\selectfont
\renewcommand{\arraystretch}{1.0}
\setlength{\tabcolsep}{4.5pt}
\begin{tabularx}{\linewidth}{@{}l|*{4}{>{\centering\arraybackslash}X}|>{\centering\arraybackslash}X@{}}
\toprule
\textbf{Model Name} & EG$\uparrow$ & PR$\uparrow$ & VQ$\uparrow$ & DS$\uparrow$ & \textbf{Overall}$\uparrow$ \\
\midrule
\rowcolor{TableTint}\multicolumn{6}{c}{\itshape Closed-source Models} \\
GPT-Image-2 & \textbf{4.096} & 4.068 & \textbf{4.494} & \textbf{4.215} & \textbf{3.974} \\
Nano-Banana-Pro & 3.983 & 3.880 & 4.424 & 4.010 & 3.803 \\
Qwen-Image-3.0 & 3.916 & 3.485 & 4.324 & 4.149 & 3.699 \\
Seedream-5.0-Pro & 4.035 & \underline{4.163} & \underline{4.441} & \underline{4.189} & \underline{3.939} \\
\midrule
\rowcolor{TableTint}\multicolumn{6}{c}{\itshape Open-source Models} \\
HunyuanImage-3.0-Instruct-Distil & 3.917 & \textbf{4.234} & 4.357 & 4.018 & 3.823 \\
HunyuanImage-3.0-Instruct-Distil-CoT & 3.985 & 4.093 & 4.319 & 4.098 & 3.848 \\
FireRed-Image-Edit-1.1 & \underline{4.038} & 3.670 & 4.268 & 4.012 & 3.764 \\
JoyAI-Image-Edit & 4.016 & 3.539 & 4.187 & 3.824 & 3.661 \\
Qwen-Image-Edit-2511 & 3.893 & 3.720 & 4.075 & 3.768 & 3.587 \\
Boogu-Image-0.1-Edit & 3.778 & 3.720 & 4.064 & 3.806 & 3.525 \\
SenseNova-U1-8B-MoT-CoT & 3.779 & 3.651 & 3.953 & 3.628 & 3.445 \\
SenseNova-U1-8B-MoT & 3.963 & 3.495 & 3.911 & 3.761 & 3.549 \\
LongCat-Image-Edit & 3.562 & 3.509 & 4.087 & 3.749 & 3.344 \\
HiDream-O1-Image & 3.184 & 3.010 & 3.849 & 3.399 & 2.885 \\
FLUX.2 [dev] & 2.215 & 3.298 & 3.628 & 2.302 & 2.044 \\
\bottomrule
\end{tabularx}
\end{table}

\clearpage
\begin{table}[!htbp]
\centering
\caption{Quantitative comparison on \single{}: Portrait Editing. Bold and underline denote the best and second-best scores.}
\label{tab:wegen-single-primary-portrait}
\fontsize{8.5pt}{10pt}\selectfont
\renewcommand{\arraystretch}{1.0}
\setlength{\tabcolsep}{4.5pt}
\begin{tabularx}{\linewidth}{@{}l|*{4}{>{\centering\arraybackslash}X}|>{\centering\arraybackslash}X@{}}
\toprule
\textbf{Model Name} & EG$\uparrow$ & PR$\uparrow$ & VQ$\uparrow$ & DS$\uparrow$ & \textbf{Overall}$\uparrow$ \\
\midrule
\rowcolor{TableTint}\multicolumn{6}{c}{\itshape Closed-source Models} \\
GPT-Image-2 & \textbf{4.443} & \underline{4.156} & \textbf{4.464} & \textbf{4.328} & \textbf{4.222} \\
Nano-Banana-Pro & \underline{4.390} & 3.696 & 4.350 & 4.226 & 4.050 \\
Qwen-Image-3.0 & 4.332 & 3.451 & 4.337 & 4.279 & 3.977 \\
Seedream-5.0-Pro & 4.376 & 4.125 & \underline{4.402} & \underline{4.325} & \underline{4.162} \\
\midrule
\rowcolor{TableTint}\multicolumn{6}{c}{\itshape Open-source Models} \\
HunyuanImage-3.0-Instruct-Distil & 4.272 & \textbf{4.231} & 4.306 & 4.071 & 4.038 \\
HunyuanImage-3.0-Instruct-Distil-CoT & 4.193 & 4.095 & 4.305 & 4.162 & 3.986 \\
FireRed-Image-Edit-1.1 & 4.192 & 3.506 & 4.186 & 4.030 & 3.811 \\
JoyAI-Image-Edit & 4.195 & 3.505 & 4.177 & 3.880 & 3.774 \\
Qwen-Image-Edit-2511 & 4.263 & 3.449 & 4.186 & 3.937 & 3.819 \\
Boogu-Image-0.1-Edit & 4.094 & 3.618 & 4.161 & 3.945 & 3.749 \\
SenseNova-U1-8B-MoT-CoT & 4.240 & 3.556 & 4.017 & 3.756 & 3.748 \\
SenseNova-U1-8B-MoT & 4.276 & 3.359 & 4.016 & 3.834 & 3.749 \\
LongCat-Image-Edit & 4.096 & 3.461 & 4.120 & 3.949 & 3.711 \\
HiDream-O1-Image & 3.496 & 2.793 & 3.975 & 3.589 & 3.100 \\
FLUX.2 [dev] & 1.995 & 3.341 & 3.352 & 1.990 & 1.785 \\
\bottomrule
\end{tabularx}
\end{table}

\begin{table}[!htbp]
\centering
\caption{Quantitative comparison on \single{}: Reasoning-based Editing. Bold and underline denote the best and second-best scores.}
\label{tab:wegen-single-primary-reasoning}
\fontsize{8.5pt}{10pt}\selectfont
\renewcommand{\arraystretch}{1.0}
\setlength{\tabcolsep}{4.5pt}
\begin{tabularx}{\linewidth}{@{}l|*{4}{>{\centering\arraybackslash}X}|>{\centering\arraybackslash}X@{}}
\toprule
\textbf{Model Name} & EG$\uparrow$ & PR$\uparrow$ & VQ$\uparrow$ & DS$\uparrow$ & \textbf{Overall}$\uparrow$ \\
\midrule
\rowcolor{TableTint}\multicolumn{6}{c}{\itshape Closed-source Models} \\
GPT-Image-2 & \textbf{4.228} & 4.134 & \textbf{4.433} & \underline{4.143} & \textbf{4.036} \\
Nano-Banana-Pro & 4.138 & 3.745 & 4.423 & 3.992 & 3.864 \\
Qwen-Image-3.0 & 4.153 & 3.627 & 4.385 & 3.923 & 3.825 \\
Seedream-5.0-Pro & \underline{4.155} & 4.074 & \underline{4.429} & \textbf{4.184} & \underline{3.990} \\
\midrule
\rowcolor{TableTint}\multicolumn{6}{c}{\itshape Open-source Models} \\
HunyuanImage-3.0-Instruct-Distil & 3.822 & \textbf{4.205} & 4.346 & 3.846 & 3.715 \\
HunyuanImage-3.0-Instruct-Distil-CoT & 3.845 & \underline{4.155} & 4.302 & 4.021 & 3.754 \\
FireRed-Image-Edit-1.1 & 3.907 & 3.698 & 4.249 & 3.790 & 3.631 \\
JoyAI-Image-Edit & 4.026 & 3.582 & 4.213 & 3.722 & 3.655 \\
Qwen-Image-Edit-2511 & 3.758 & 3.585 & 4.114 & 3.573 & 3.438 \\
Boogu-Image-0.1-Edit & 3.763 & 3.834 & 4.153 & 3.687 & 3.527 \\
SenseNova-U1-8B-MoT-CoT & 4.057 & 3.723 & 4.012 & 3.595 & 3.631 \\
SenseNova-U1-8B-MoT & 4.029 & 3.357 & 4.080 & 3.729 & 3.589 \\
LongCat-Image-Edit & 3.586 & 3.507 & 3.966 & 3.593 & 3.303 \\
HiDream-O1-Image & 3.120 & 3.055 & 3.862 & 3.171 & 2.802 \\
FLUX.2 [dev] & 1.773 & 3.405 & 3.288 & 1.738 & 1.592 \\
\bottomrule
\end{tabularx}
\end{table}

\clearpage
\begin{table}[!htbp]
\centering
\caption{Quantitative comparison on \single{}: Product and Commercial Editing. Bold and underline denote the best and second-best scores.}
\label{tab:wegen-single-primary-commercial}
\fontsize{8.5pt}{10pt}\selectfont
\renewcommand{\arraystretch}{1.0}
\setlength{\tabcolsep}{4.5pt}
\begin{tabularx}{\linewidth}{@{}l|*{4}{>{\centering\arraybackslash}X}|>{\centering\arraybackslash}X@{}}
\toprule
\textbf{Model Name} & EG$\uparrow$ & PR$\uparrow$ & VQ$\uparrow$ & DS$\uparrow$ & \textbf{Overall}$\uparrow$ \\
\midrule
\rowcolor{TableTint}\multicolumn{6}{c}{\itshape Closed-source Models} \\
GPT-Image-2 & 4.117 & 3.372 & \underline{4.314} & 4.125 & 3.789 \\
Nano-Banana-Pro & \textbf{4.223} & 3.269 & 4.234 & 4.088 & \underline{3.806} \\
Qwen-Image-3.0 & \underline{4.190} & 3.216 & 4.230 & \underline{4.153} & 3.792 \\
Seedream-5.0-Pro & 4.003 & \underline{3.806} & \textbf{4.471} & \textbf{4.264} & \textbf{3.874} \\
\midrule
\rowcolor{TableTint}\multicolumn{6}{c}{\itshape Open-source Models} \\
HunyuanImage-3.0-Instruct-Distil & 3.976 & \textbf{3.838} & 4.189 & 3.901 & 3.716 \\
HunyuanImage-3.0-Instruct-Distil-CoT & 3.912 & 3.460 & 4.042 & 3.790 & 3.556 \\
FireRed-Image-Edit-1.1 & 4.002 & 2.866 & 4.036 & 3.701 & 3.457 \\
JoyAI-Image-Edit & 3.926 & 3.226 & 4.135 & 3.706 & 3.505 \\
Qwen-Image-Edit-2511 & 3.881 & 2.865 & 3.999 & 3.451 & 3.321 \\
Boogu-Image-0.1-Edit & 3.792 & 3.241 & 3.939 & 3.551 & 3.349 \\
SenseNova-U1-8B-MoT-CoT & 3.647 & 3.336 & 3.733 & 3.523 & 3.242 \\
SenseNova-U1-8B-MoT & 3.742 & 2.837 & 3.598 & 3.425 & 3.141 \\
LongCat-Image-Edit & 3.464 & 2.505 & 3.486 & 3.127 & 2.848 \\
HiDream-O1-Image & 2.346 & 2.726 & 3.653 & 2.482 & 2.070 \\
FLUX.2 [dev] & 2.366 & 3.422 & 3.389 & 2.242 & 2.099 \\
\bottomrule
\end{tabularx}
\end{table}

\begin{table}[!htbp]
\centering
\caption{Quantitative comparison on \single{}: Restoration and Low-level Enhancement. Bold and underline denote the best and second-best scores.}
\label{tab:wegen-single-primary-restoration}
\fontsize{8.5pt}{10pt}\selectfont
\renewcommand{\arraystretch}{1.0}
\setlength{\tabcolsep}{4.5pt}
\begin{tabularx}{\linewidth}{@{}l|*{4}{>{\centering\arraybackslash}X}|>{\centering\arraybackslash}X@{}}
\toprule
\textbf{Model Name} & EG$\uparrow$ & PR$\uparrow$ & VQ$\uparrow$ & DS$\uparrow$ & \textbf{Overall}$\uparrow$ \\
\midrule
\rowcolor{TableTint}\multicolumn{6}{c}{\itshape Closed-source Models} \\
GPT-Image-2 & \textbf{4.347} & \textbf{4.425} & \textbf{4.565} & \textbf{4.462} & \textbf{4.272} \\
Nano-Banana-Pro & 3.612 & 4.003 & \underline{4.550} & \underline{4.392} & 3.726 \\
Qwen-Image-3.0 & 3.672 & 3.950 & 4.313 & 4.290 & 3.678 \\
Seedream-5.0-Pro & 4.117 & 4.202 & 4.465 & 4.362 & 4.044 \\
\midrule
\rowcolor{TableTint}\multicolumn{6}{c}{\itshape Open-source Models} \\
HunyuanImage-3.0-Instruct-Distil & 4.002 & \underline{4.304} & 4.328 & 4.218 & 3.932 \\
HunyuanImage-3.0-Instruct-Distil-CoT & \underline{4.224} & 4.300 & 4.444 & 4.146 & \underline{4.070} \\
FireRed-Image-Edit-1.1 & 4.083 & 3.897 & 4.323 & 3.982 & 3.839 \\
JoyAI-Image-Edit & 4.060 & 3.843 & 4.215 & 3.998 & 3.797 \\
Qwen-Image-Edit-2511 & 3.392 & 4.261 & 4.157 & 3.974 & 3.462 \\
Boogu-Image-0.1-Edit & 3.322 & 4.010 & 4.211 & 3.835 & 3.346 \\
SenseNova-U1-8B-MoT-CoT & 3.583 & 4.027 & 4.057 & 4.066 & 3.536 \\
SenseNova-U1-8B-MoT & 3.429 & 4.006 & 4.115 & 3.835 & 3.390 \\
LongCat-Image-Edit & 3.230 & 3.988 & 4.194 & 3.923 & 3.308 \\
HiDream-O1-Image & 3.260 & 3.149 & 3.681 & 3.747 & 3.009 \\
FLUX.2 [dev] & 2.678 & 3.620 & 3.749 & 3.233 & 2.639 \\
\bottomrule
\end{tabularx}
\end{table}

\clearpage
\begin{table}[!htbp]
\centering
\caption{Quantitative comparison on \single{}: Action and Interaction Editing. Bold and underline denote the best and second-best scores.}
\label{tab:wegen-single-primary-action}
\fontsize{8.5pt}{10pt}\selectfont
\renewcommand{\arraystretch}{1.0}
\setlength{\tabcolsep}{4.5pt}
\begin{tabularx}{\linewidth}{@{}l|*{4}{>{\centering\arraybackslash}X}|>{\centering\arraybackslash}X@{}}
\toprule
\textbf{Model Name} & EG$\uparrow$ & PR$\uparrow$ & VQ$\uparrow$ & DS$\uparrow$ & \textbf{Overall}$\uparrow$ \\
\midrule
\rowcolor{TableTint}\multicolumn{6}{c}{\itshape Closed-source Models} \\
GPT-Image-2 & \textbf{4.376} & 3.662 & \textbf{4.364} & \underline{4.009} & \textbf{3.983} \\
Nano-Banana-Pro & \underline{4.353} & 3.317 & 4.190 & 3.762 & 3.804 \\
Qwen-Image-3.0 & 4.190 & 3.047 & 4.198 & 3.910 & 3.691 \\
Seedream-5.0-Pro & 4.225 & 3.613 & \underline{4.312} & \textbf{4.047} & \underline{3.882} \\
\midrule
\rowcolor{TableTint}\multicolumn{6}{c}{\itshape Open-source Models} \\
HunyuanImage-3.0-Instruct-Distil & 3.640 & \textbf{3.975} & 4.133 & 3.338 & 3.390 \\
HunyuanImage-3.0-Instruct-Distil-CoT & 4.011 & \underline{3.751} & 4.060 & 3.777 & 3.663 \\
FireRed-Image-Edit-1.1 & 4.013 & 3.000 & 4.202 & 3.763 & 3.539 \\
JoyAI-Image-Edit & 4.092 & 2.951 & 4.150 & 3.824 & 3.582 \\
Qwen-Image-Edit-2511 & 4.045 & 2.941 & 4.013 & 3.797 & 3.517 \\
Boogu-Image-0.1-Edit & 3.880 & 3.338 & 3.979 & 3.653 & 3.455 \\
SenseNova-U1-8B-MoT-CoT & 3.873 & 3.106 & 3.976 & 3.414 & 3.344 \\
SenseNova-U1-8B-MoT & 4.054 & 2.741 & 3.927 & 3.295 & 3.340 \\
LongCat-Image-Edit & 4.133 & 2.954 & 3.980 & 3.504 & 3.492 \\
HiDream-O1-Image & 3.334 & 2.746 & 3.749 & 3.072 & 2.817 \\
FLUX.2 [dev] & 1.781 & 3.221 & 3.180 & 1.130 & 1.392 \\
\bottomrule
\end{tabularx}
\end{table}

\begin{table}[!htbp]
\centering
\caption{Quantitative comparison on \single{}: Traditional Chinese Cultural Editing. Bold and underline denote the best and second-best scores.}
\label{tab:wegen-single-primary-culture}
\fontsize{8.5pt}{10pt}\selectfont
\renewcommand{\arraystretch}{1.0}
\setlength{\tabcolsep}{4.5pt}
\begin{tabularx}{\linewidth}{@{}l|*{4}{>{\centering\arraybackslash}X}|>{\centering\arraybackslash}X@{}}
\toprule
\textbf{Model Name} & EG$\uparrow$ & PR$\uparrow$ & VQ$\uparrow$ & DS$\uparrow$ & \textbf{Overall}$\uparrow$ \\
\midrule
\rowcolor{TableTint}\multicolumn{6}{c}{\itshape Closed-source Models} \\
GPT-Image-2 & 4.450 & 4.105 & 3.802 & 3.671 & 3.919 \\
Nano-Banana-Pro & \textbf{4.566} & 3.703 & \underline{3.960} & 3.774 & \underline{3.965} \\
Qwen-Image-3.0 & \underline{4.496} & 3.264 & 3.850 & 3.880 & 3.840 \\
Seedream-5.0-Pro & 4.410 & \textbf{4.340} & \textbf{4.097} & \textbf{4.082} & \textbf{4.104} \\
\midrule
\rowcolor{TableTint}\multicolumn{6}{c}{\itshape Open-source Models} \\
HunyuanImage-3.0-Instruct-Distil & 4.320 & \underline{4.311} & 3.939 & 3.862 & 3.957 \\
HunyuanImage-3.0-Instruct-Distil-CoT & 4.284 & 4.187 & 3.803 & 3.677 & 3.838 \\
FireRed-Image-Edit-1.1 & 4.233 & 3.818 & 3.764 & 3.457 & 3.670 \\
JoyAI-Image-Edit & 4.431 & 3.636 & 3.896 & 3.574 & 3.808 \\
Qwen-Image-Edit-2511 & 4.255 & 3.875 & 3.873 & 3.254 & 3.666 \\
Boogu-Image-0.1-Edit & 4.236 & 4.029 & 3.702 & \underline{3.982} & 3.833 \\
SenseNova-U1-8B-MoT-CoT & 4.203 & 3.883 & 3.467 & 3.090 & 3.514 \\
SenseNova-U1-8B-MoT & 4.216 & 3.796 & 3.352 & 3.538 & 3.594 \\
LongCat-Image-Edit & 3.918 & 3.835 & 3.491 & 2.787 & 3.263 \\
HiDream-O1-Image & 3.365 & 2.914 & 3.446 & 2.875 & 2.760 \\
FLUX.2 [dev] & 1.815 & 3.118 & 2.727 & 0.698 & 1.183 \\
\bottomrule
\end{tabularx}
\end{table}

\clearpage
\subsubsection{Text-Intensive Editing}
\label{app:primary-tables-text}
\begin{table}[!htbp]
\centering
\caption{Quantitative comparison on \textedit{}: Basic Text Editing. Bold and underline denote the best and second-best scores.}
\label{tab:wegen-text-primary-basic-text}
\fontsize{8pt}{9pt}\selectfont
\renewcommand{\arraystretch}{1.0}
\setlength{\tabcolsep}{1.0pt}
\begin{tabularx}{\linewidth}{@{}l|*{4}{>{\centering\arraybackslash}X}|*{4}{>{\centering\arraybackslash}X}|c}
\toprule
\raisebox{-5pt}[0pt][0pt]{\textbf{Model Name}} & \multicolumn{4}{c|}{\textbf{Text-side}} & \multicolumn{4}{c|}{\textbf{Image-side}} & \raisebox{-5pt}[0pt][0pt]{\textbf{Overall} {\normalsize$\uparrow$}} \\
\cmidrule(lr){2-5}\cmidrule(lr){6-9}
& EG {\normalsize$\uparrow$} & PR {\normalsize$\uparrow$} & VQ {\normalsize$\uparrow$} & DS {\normalsize$\uparrow$} & EG {\normalsize$\uparrow$} & PR {\normalsize$\uparrow$} & VQ {\normalsize$\uparrow$} & DS {\normalsize$\uparrow$} & \\
\midrule
\rowcolor{TableTint}\multicolumn{10}{c}{\itshape Closed-source Models} \\
GPT-Image-2 & 4.249 & 4.050 & \underline{4.308} & 4.356 & \textbf{4.594} & 3.924 & \underline{4.409} & \textbf{4.446} & 4.109 \\
Nano-Banana-Pro & 4.317 & \underline{4.129} & 4.303 & 4.305 & 4.392 & \underline{4.220} & 4.227 & 4.295 & \underline{4.141} \\
Qwen-Image-3.0 & \underline{4.350} & 3.692 & 4.289 & \underline{4.395} & 4.507 & 3.887 & 4.408 & 4.410 & 4.096 \\
Seedream-5.0-Pro & \textbf{4.366} & \textbf{4.327} & \textbf{4.412} & \textbf{4.475} & 4.352 & 4.218 & \textbf{4.447} & \underline{4.428} & \textbf{4.260} \\
\midrule
\rowcolor{TableTint}\multicolumn{10}{c}{\itshape Open-source Models} \\
SenseNova-U1-8B-MoT-CoT & 4.022 & 3.950 & 3.832 & 3.965 & 4.352 & 4.105 & 3.941 & 4.089 & 3.790 \\
HunyuanImage-3.0-Instruct-Distil & 3.917 & 3.868 & 3.894 & 4.120 & 4.362 & 4.174 & 4.111 & 4.220 & 3.776 \\
HunyuanImage-3.0-Instruct-Distil-CoT & 4.112 & 3.914 & 4.084 & 4.177 & 4.418 & \textbf{4.261} & 4.195 & 4.324 & 3.929 \\
SenseNova-U1-8B-MoT & 4.029 & 3.890 & 3.931 & 3.962 & 4.415 & 3.840 & 3.892 & 4.094 & 3.792 \\
JoyAI-Image-Edit & 3.934 & 3.716 & 3.754 & 4.024 & \underline{4.582} & 3.920 & 4.129 & 4.275 & 3.710 \\
FireRed-Image-Edit-1.1 & 3.785 & 3.444 & 3.673 & 3.972 & 4.498 & 3.624 & 3.930 & 4.237 & 3.549 \\
Qwen-Image-Edit-2511 & 3.727 & 3.639 & 3.683 & 3.888 & 4.383 & 3.829 & 3.962 & 4.108 & 3.532 \\
Boogu-Image-0.1-Edit & 3.876 & 3.330 & 3.786 & 3.968 & 4.432 & 3.470 & 3.902 & 4.202 & 3.578 \\
LongCat-Image-Edit & 3.044 & 3.702 & 3.432 & 3.315 & 4.195 & 3.655 & 3.787 & 3.958 & 2.994 \\
HiDream-O1-Image & 2.864 & 2.584 & 3.237 & 3.121 & 3.554 & 3.052 & 3.477 & 3.535 & 2.513 \\
FLUX.2 [dev] & 2.205 & 3.429 & 3.677 & 2.391 & 2.717 & 3.339 & 3.566 & 2.776 & 2.214 \\
\bottomrule
\end{tabularx}
\end{table}

\begin{table}[!htbp]
\centering
\caption{Quantitative comparison on \textedit{}: Posters and Layout. Bold and underline denote the best and second-best scores.}
\label{tab:wegen-text-primary-poster}
\fontsize{8pt}{9pt}\selectfont
\renewcommand{\arraystretch}{1.0}
\setlength{\tabcolsep}{1.0pt}
\begin{tabularx}{\linewidth}{@{}l|*{4}{>{\centering\arraybackslash}X}|*{4}{>{\centering\arraybackslash}X}|c}
\toprule
\raisebox{-5pt}[0pt][0pt]{\textbf{Model Name}} & \multicolumn{4}{c|}{\textbf{Text-side}} & \multicolumn{4}{c|}{\textbf{Image-side}} & \raisebox{-5pt}[0pt][0pt]{\textbf{Overall} {\normalsize$\uparrow$}} \\
\cmidrule(lr){2-5}\cmidrule(lr){6-9}
& EG {\normalsize$\uparrow$} & PR {\normalsize$\uparrow$} & VQ {\normalsize$\uparrow$} & DS {\normalsize$\uparrow$} & EG {\normalsize$\uparrow$} & PR {\normalsize$\uparrow$} & VQ {\normalsize$\uparrow$} & DS {\normalsize$\uparrow$} & \\
\midrule
\rowcolor{TableTint}\multicolumn{10}{c}{\itshape Closed-source Models} \\
GPT-Image-2 & 4.144 & 3.817 & \underline{4.372} & \underline{4.439} & \textbf{4.567} & 3.908 & \underline{4.483} & \textbf{4.471} & \underline{4.038} \\
Nano-Banana-Pro & 4.032 & \underline{4.032} & 4.326 & 4.177 & 4.304 & \underline{4.208} & 4.216 & 4.268 & 3.913 \\
Qwen-Image-3.0 & \underline{4.257} & 3.509 & \textbf{4.477} & 4.289 & 4.528 & 3.854 & 4.463 & 4.390 & 4.015 \\
Seedream-5.0-Pro & \textbf{4.288} & \textbf{4.152} & 4.316 & \textbf{4.496} & 4.260 & 4.187 & \textbf{4.528} & \underline{4.407} & \textbf{4.152} \\
\midrule
\rowcolor{TableTint}\multicolumn{10}{c}{\itshape Open-source Models} \\
SenseNova-U1-8B-MoT-CoT & 3.715 & 3.795 & 3.846 & 3.817 & 4.264 & 4.200 & 4.000 & 4.032 & 3.532 \\
HunyuanImage-3.0-Instruct-Distil & 3.593 & 3.685 & 3.466 & 3.922 & 4.280 & 4.080 & 4.024 & 4.160 & 3.424 \\
HunyuanImage-3.0-Instruct-Distil-CoT & 3.860 & 3.735 & 3.900 & 4.099 & 4.360 & \textbf{4.256} & 4.224 & 4.340 & 3.695 \\
SenseNova-U1-8B-MoT & 3.697 & 3.671 & 3.688 & 3.807 & 4.304 & 3.992 & 4.056 & 4.184 & 3.471 \\
JoyAI-Image-Edit & 3.534 & 3.557 & 3.733 & 3.900 & \underline{4.552} & 4.152 & 4.200 & 4.324 & 3.428 \\
FireRed-Image-Edit-1.1 & 3.462 & 3.215 & 3.480 & 3.862 & 4.456 & 3.632 & 3.960 & 4.312 & 3.251 \\
Qwen-Image-Edit-2511 & 3.299 & 3.425 & 3.348 & 3.631 & 4.352 & 3.840 & 3.944 & 4.140 & 3.115 \\
Boogu-Image-0.1-Edit & 3.633 & 3.260 & 3.715 & 3.901 & 4.464 & 3.376 & 3.960 & 4.180 & 3.389 \\
LongCat-Image-Edit & 2.643 & 3.388 & 3.122 & 3.235 & 4.112 & 3.704 & 3.840 & 3.900 & 2.624 \\
HiDream-O1-Image & 2.914 & 2.708 & 3.299 & 3.282 & 3.440 & 3.272 & 3.528 & 3.620 & 2.589 \\
FLUX.2 [dev] & 1.995 & 3.457 & 3.561 & 2.235 & 2.744 & 3.440 & 3.520 & 2.816 & 1.995 \\
\bottomrule
\end{tabularx}
\end{table}

\clearpage
\begin{table}[!htbp]
\centering
\caption{Quantitative comparison on \textedit{}: Interfaces, Documents, and Charts. Bold and underline denote the best and second-best scores.}
\label{tab:wegen-text-primary-document}
\fontsize{8pt}{9pt}\selectfont
\renewcommand{\arraystretch}{1.0}
\setlength{\tabcolsep}{1.0pt}
\begin{tabularx}{\linewidth}{@{}l|*{4}{>{\centering\arraybackslash}X}|*{4}{>{\centering\arraybackslash}X}|c}
\toprule
\raisebox{-5pt}[0pt][0pt]{\textbf{Model Name}} & \multicolumn{4}{c|}{\textbf{Text-side}} & \multicolumn{4}{c|}{\textbf{Image-side}} & \raisebox{-5pt}[0pt][0pt]{\textbf{Overall} {\normalsize$\uparrow$}} \\
\cmidrule(lr){2-5}\cmidrule(lr){6-9}
& EG {\normalsize$\uparrow$} & PR {\normalsize$\uparrow$} & VQ {\normalsize$\uparrow$} & DS {\normalsize$\uparrow$} & EG {\normalsize$\uparrow$} & PR {\normalsize$\uparrow$} & VQ {\normalsize$\uparrow$} & DS {\normalsize$\uparrow$} & \\
\midrule
\rowcolor{TableTint}\multicolumn{10}{c}{\itshape Closed-source Models} \\
GPT-Image-2 & 4.341 & \underline{4.120} & 4.329 & 4.423 & \textbf{4.639} & \underline{4.139} & \textbf{4.500} & \textbf{4.583} & \underline{4.187} \\
Nano-Banana-Pro & 4.260 & 3.949 & \underline{4.333} & \underline{4.434} & \underline{4.595} & \textbf{4.351} & 4.351 & 4.351 & 4.103 \\
Qwen-Image-3.0 & \underline{4.385} & 3.856 & 4.299 & 4.297 & \textbf{4.639} & 3.833 & 4.167 & 4.500 & 4.106 \\
Seedream-5.0-Pro & \textbf{4.535} & \textbf{4.371} & \textbf{4.435} & \textbf{4.576} & 4.541 & 4.027 & \underline{4.486} & \underline{4.568} & \textbf{4.392} \\
\midrule
\rowcolor{TableTint}\multicolumn{10}{c}{\itshape Open-source Models} \\
SenseNova-U1-8B-MoT-CoT & 4.040 & 3.780 & 3.791 & 3.934 & 4.432 & 3.730 & 3.649 & 4.243 & 3.724 \\
HunyuanImage-3.0-Instruct-Distil & 3.655 & 3.605 & 3.831 & 4.044 & 4.270 & 4.027 & 4.000 & 4.027 & 3.505 \\
HunyuanImage-3.0-Instruct-Distil-CoT & 3.842 & 3.520 & 4.000 & 4.248 & 4.378 & 3.919 & 3.811 & 4.257 & 3.670 \\
SenseNova-U1-8B-MoT & 3.938 & 3.751 & 3.780 & 3.926 & 4.405 & 3.595 & 3.595 & 4.108 & 3.646 \\
JoyAI-Image-Edit & 3.678 & 3.379 & 3.452 & 3.787 & 4.351 & 3.811 & 3.649 & 4.338 & 3.336 \\
FireRed-Image-Edit-1.1 & 3.446 & 3.017 & 3.390 & 3.566 & 4.459 & 3.216 & 3.405 & 3.878 & 3.059 \\
Qwen-Image-Edit-2511 & 3.497 & 3.407 & 3.503 & 3.752 & 4.243 & 3.432 & 3.486 & 4.041 & 3.236 \\
Boogu-Image-0.1-Edit & 3.599 & 2.621 & 3.226 & 3.641 & 4.297 & 3.243 & 3.459 & 3.973 & 3.051 \\
LongCat-Image-Edit & 2.836 & 3.203 & 3.028 & 3.168 & 4.027 & 3.216 & 3.243 & 3.959 & 2.588 \\
HiDream-O1-Image & 2.582 & 2.249 & 2.836 & 2.875 & 3.703 & 2.649 & 2.973 & 3.108 & 2.119 \\
FLUX.2 [dev] & 2.254 & 3.073 & 3.706 & 2.605 & 2.568 & 3.243 & 3.649 & 2.554 & 2.159 \\
\bottomrule
\end{tabularx}
\end{table}

\begin{table}[!htbp]
\centering
\caption{Quantitative comparison on \textedit{}: Commercial Text--Image Editing. Bold and underline denote the best and second-best scores.}
\label{tab:wegen-text-primary-commercial}
\fontsize{8pt}{9pt}\selectfont
\renewcommand{\arraystretch}{1.0}
\setlength{\tabcolsep}{1.0pt}
\begin{tabularx}{\linewidth}{@{}l|*{4}{>{\centering\arraybackslash}X}|*{4}{>{\centering\arraybackslash}X}|c}
\toprule
\raisebox{-5pt}[0pt][0pt]{\textbf{Model Name}} & \multicolumn{4}{c|}{\textbf{Text-side}} & \multicolumn{4}{c|}{\textbf{Image-side}} & \raisebox{-5pt}[0pt][0pt]{\textbf{Overall} {\normalsize$\uparrow$}} \\
\cmidrule(lr){2-5}\cmidrule(lr){6-9}
& EG {\normalsize$\uparrow$} & PR {\normalsize$\uparrow$} & VQ {\normalsize$\uparrow$} & DS {\normalsize$\uparrow$} & EG {\normalsize$\uparrow$} & PR {\normalsize$\uparrow$} & VQ {\normalsize$\uparrow$} & DS {\normalsize$\uparrow$} & \\
\midrule
\rowcolor{TableTint}\multicolumn{10}{c}{\itshape Closed-source Models} \\
GPT-Image-2 & \underline{4.448} & 3.676 & 4.295 & 4.423 & \textbf{4.782} & 3.818 & \textbf{4.473} & \textbf{4.518} & \underline{4.160} \\
Nano-Banana-Pro & 4.382 & \underline{4.027} & 4.282 & 4.408 & 4.508 & 3.847 & 4.220 & \underline{4.432} & 4.139 \\
Qwen-Image-3.0 & 4.418 & 3.455 & \textbf{4.391} & \underline{4.515} & 4.508 & 3.763 & 4.322 & 4.356 & 4.139 \\
Seedream-5.0-Pro & \textbf{4.472} & \textbf{4.241} & \underline{4.380} & \textbf{4.578} & 4.474 & 3.895 & \underline{4.351} & 4.368 & \textbf{4.296} \\
\midrule
\rowcolor{TableTint}\multicolumn{10}{c}{\itshape Open-source Models} \\
SenseNova-U1-8B-MoT-CoT & 4.045 & 3.573 & 3.800 & 3.933 & 4.373 & 4.034 & 4.068 & 4.102 & 3.700 \\
HunyuanImage-3.0-Instruct-Distil & 4.083 & 3.780 & 4.083 & 4.270 & 4.458 & \underline{4.051} & 4.169 & 4.347 & 3.906 \\
HunyuanImage-3.0-Instruct-Distil-CoT & 4.211 & 3.642 & 4.183 & 4.259 & 4.576 & \textbf{4.254} & 4.220 & 4.314 & 3.960 \\
SenseNova-U1-8B-MoT & 4.164 & 3.618 & 3.873 & 3.927 & 4.492 & 3.678 & 3.814 & 3.992 & 3.767 \\
JoyAI-Image-Edit & 4.191 & 3.609 & 3.845 & 4.102 & 4.576 & 3.763 & 4.271 & 4.203 & 3.835 \\
FireRed-Image-Edit-1.1 & 3.745 & 3.382 & 3.600 & 4.037 & \underline{4.610} & 3.390 & 3.949 & 4.347 & 3.500 \\
Qwen-Image-Edit-2511 & 3.745 & 3.527 & 3.682 & 3.942 & 4.508 & 3.763 & 4.068 & 4.127 & 3.516 \\
Boogu-Image-0.1-Edit & 3.955 & 3.236 & 3.773 & 4.194 & 4.254 & 3.119 & 3.898 & 4.331 & 3.623 \\
LongCat-Image-Edit & 3.255 & 3.436 & 3.564 & 3.705 & 4.424 & 3.492 & 4.034 & 3.890 & 3.147 \\
HiDream-O1-Image & 3.227 & 3.000 & 3.545 & 3.641 & 3.661 & 2.915 & 3.390 & 3.475 & 2.955 \\
FLUX.2 [dev] & 1.964 & 3.418 & 3.555 & 2.154 & 3.051 & 3.237 & 3.542 & 2.873 & 1.967 \\
\bottomrule
\end{tabularx}
\end{table}

\clearpage
\begin{table}[!htbp]
\centering
\caption{Quantitative comparison on \textedit{}: Social-media Text--Image Editing. Bold and underline denote the best and second-best scores.}
\label{tab:wegen-text-primary-social}
\fontsize{8pt}{9pt}\selectfont
\renewcommand{\arraystretch}{1.0}
\setlength{\tabcolsep}{1.0pt}
\begin{tabularx}{\linewidth}{@{}l|*{4}{>{\centering\arraybackslash}X}|*{4}{>{\centering\arraybackslash}X}|c}
\toprule
\raisebox{-5pt}[0pt][0pt]{\textbf{Model Name}} & \multicolumn{4}{c|}{\textbf{Text-side}} & \multicolumn{4}{c|}{\textbf{Image-side}} & \raisebox{-5pt}[0pt][0pt]{\textbf{Overall} {\normalsize$\uparrow$}} \\
\cmidrule(lr){2-5}\cmidrule(lr){6-9}
& EG {\normalsize$\uparrow$} & PR {\normalsize$\uparrow$} & VQ {\normalsize$\uparrow$} & DS {\normalsize$\uparrow$} & EG {\normalsize$\uparrow$} & PR {\normalsize$\uparrow$} & VQ {\normalsize$\uparrow$} & DS {\normalsize$\uparrow$} & \\
\midrule
\rowcolor{TableTint}\multicolumn{10}{c}{\itshape Closed-source Models} \\
GPT-Image-2 & \textbf{4.457} & 3.837 & \underline{4.511} & 4.454 & \textbf{4.571} & 3.839 & 4.304 & 4.348 & \textbf{4.227} \\
Nano-Banana-Pro & 4.274 & 4.105 & 4.274 & \underline{4.541} & 4.228 & \underline{4.404} & 4.018 & 4.351 & 4.127 \\
Qwen-Image-3.0 & \underline{4.448} & 3.615 & \textbf{4.521} & \textbf{4.549} & 4.534 & 3.810 & \textbf{4.483} & \textbf{4.466} & \underline{4.209} \\
Seedream-5.0-Pro & 4.333 & \textbf{4.312} & 4.387 & 4.449 & 4.175 & \underline{4.404} & \underline{4.368} & \underline{4.395} & 4.195 \\
\midrule
\rowcolor{TableTint}\multicolumn{10}{c}{\itshape Open-source Models} \\
SenseNova-U1-8B-MoT-CoT & 4.042 & 3.990 & 4.177 & 4.021 & 4.345 & 4.069 & 3.879 & 4.164 & 3.858 \\
HunyuanImage-3.0-Instruct-Distil & 3.906 & 4.073 & 3.875 & 4.311 & 4.259 & 4.345 & 4.121 & 4.241 & 3.824 \\
HunyuanImage-3.0-Instruct-Distil-CoT & 4.260 & \underline{4.219} & 4.260 & 4.207 & 4.379 & \textbf{4.534} & 4.293 & 4.241 & 4.084 \\
SenseNova-U1-8B-MoT & 3.979 & 3.917 & 4.125 & 4.046 & 4.534 & 3.759 & 3.828 & 4.121 & 3.790 \\
JoyAI-Image-Edit & 3.844 & 3.938 & 4.073 & 4.182 & \underline{4.552} & 3.741 & 4.190 & 4.250 & 3.780 \\
FireRed-Image-Edit-1.1 & 3.469 & 3.417 & 3.719 & 4.100 & 4.448 & 3.690 & 4.207 & 4.267 & 3.404 \\
Qwen-Image-Edit-2511 & 3.510 & 3.792 & 3.750 & 3.953 & 4.379 & 3.966 & 4.241 & 4.190 & 3.454 \\
Boogu-Image-0.1-Edit & 3.896 & 3.646 & 4.052 & 3.982 & 4.328 & 3.828 & 4.034 & 4.207 & 3.682 \\
LongCat-Image-Edit & 3.417 & 3.646 & 3.542 & 3.886 & 4.172 & 3.655 & 3.810 & 4.069 & 3.295 \\
HiDream-O1-Image & 3.240 & 2.948 & 3.688 & 3.675 & 3.638 & 2.793 & 3.603 & 3.414 & 2.984 \\
FLUX.2 [dev] & 1.968 & 3.421 & 3.653 & 2.163 & 2.684 & 3.421 & 3.561 & 2.816 & 1.965 \\
\bottomrule
\end{tabularx}
\end{table}

\clearpage
\subsubsection{Multi-Reference Editing}
\label{app:primary-tables-multi}
\begin{table}[!htbp]
\centering
\caption{Quantitative comparison on \multi{}: Portrait Editing. Bold and underline denote the best and second-best scores.}
\label{tab:wegen-multi-primary-portrait}
\fontsize{8.5pt}{10pt}\selectfont
\renewcommand{\arraystretch}{1.0}
\setlength{\tabcolsep}{4.5pt}
\begin{tabularx}{\linewidth}{@{}l|*{3}{>{\centering\arraybackslash}X}|>{\centering\arraybackslash}X@{}}
\toprule
\textbf{Model Name} & EG$\uparrow$ & PR$\uparrow$ & VQ$\uparrow$ & \textbf{Overall}$\uparrow$ \\
\midrule
\rowcolor{TableTint}\multicolumn{5}{c}{\itshape Closed-source Models} \\
GPT-Image-2 & \underline{4.315} & \underline{4.107} & \underline{3.658} & \underline{4.101} \\
Nano-Banana-Pro & 4.284 & \textbf{4.122} & 3.236 & 4.009 \\
Seedream-5.0-Pro & \textbf{4.374} & 4.094 & \textbf{3.835} & \textbf{4.154} \\
\midrule
\rowcolor{TableTint}\multicolumn{5}{c}{\itshape Open-source Models} \\
HunyuanImage-3.0-Instruct-Distil & 3.496 & 3.656 & 3.216 & 3.504 \\
FireRed-Image-Edit-1.1 & 2.980 & 2.987 & 3.107 & 3.008 \\
Qwen-Image-Edit-2511 & 2.740 & 2.887 & 3.160 & 2.883 \\
\bottomrule
\end{tabularx}
\end{table}

\begin{table}[!htbp]
\centering
\caption{Quantitative comparison on \multi{}: Product and Commercial Editing. Bold and underline denote the best and second-best scores.}
\label{tab:wegen-multi-primary-commercial}
\fontsize{8.5pt}{10pt}\selectfont
\renewcommand{\arraystretch}{1.0}
\setlength{\tabcolsep}{4.5pt}
\begin{tabularx}{\linewidth}{@{}l|*{3}{>{\centering\arraybackslash}X}|>{\centering\arraybackslash}X@{}}
\toprule
\textbf{Model Name} & EG$\uparrow$ & PR$\uparrow$ & VQ$\uparrow$ & \textbf{Overall}$\uparrow$ \\
\midrule
\rowcolor{TableTint}\multicolumn{5}{c}{\itshape Closed-source Models} \\
GPT-Image-2 & \underline{4.220} & \textbf{4.068} & \underline{3.339} & \underline{3.983} \\
Nano-Banana-Pro & 3.795 & 3.889 & 3.043 & 3.682 \\
Seedream-5.0-Pro & \textbf{4.250} & \underline{4.056} & \textbf{3.361} & \textbf{3.994} \\
\midrule
\rowcolor{TableTint}\multicolumn{5}{c}{\itshape Open-source Models} \\
HunyuanImage-3.0-Instruct-Distil & 3.198 & 3.089 & 2.861 & 3.087 \\
FireRed-Image-Edit-1.1 & 2.442 & 2.775 & 2.858 & 2.658 \\
Qwen-Image-Edit-2511 & 2.550 & 2.767 & 2.875 & 2.702 \\
\bottomrule
\end{tabularx}
\end{table}

\begin{table}[!htbp]
\centering
\caption{Quantitative comparison on \multi{}: Text Editing. Bold and underline denote the best and second-best scores.}
\label{tab:wegen-multi-primary-text}
\fontsize{8.5pt}{10pt}\selectfont
\renewcommand{\arraystretch}{1.0}
\setlength{\tabcolsep}{4.5pt}
\begin{tabularx}{\linewidth}{@{}l|*{3}{>{\centering\arraybackslash}X}|>{\centering\arraybackslash}X@{}}
\toprule
\textbf{Model Name} & EG$\uparrow$ & PR$\uparrow$ & VQ$\uparrow$ & \textbf{Overall}$\uparrow$ \\
\midrule
\rowcolor{TableTint}\multicolumn{5}{c}{\itshape Closed-source Models} \\
GPT-Image-2 & \textbf{4.147} & \textbf{3.983} & \underline{3.690} & \textbf{3.990} \\
Nano-Banana-Pro & 3.664 & 3.779 & 3.319 & 3.641 \\
Seedream-5.0-Pro & \underline{4.018} & \underline{3.936} & \textbf{3.697} & \underline{3.921} \\
\midrule
\rowcolor{TableTint}\multicolumn{5}{c}{\itshape Open-source Models} \\
HunyuanImage-3.0-Instruct-Distil & 2.700 & 3.130 & 2.730 & 2.878 \\
FireRed-Image-Edit-1.1 & 1.839 & 2.703 & 2.695 & 2.356 \\
Qwen-Image-Edit-2511 & 1.873 & 2.508 & 2.754 & 2.303 \\
\bottomrule
\end{tabularx}
\end{table}

\clearpage
\begin{table}[!htbp]
\centering
\caption{Quantitative comparison on \multi{}: Scene and Spatial Editing. Bold and underline denote the best and second-best scores.}
\label{tab:wegen-multi-primary-scene}
\fontsize{8.5pt}{10pt}\selectfont
\renewcommand{\arraystretch}{1.0}
\setlength{\tabcolsep}{4.5pt}
\begin{tabularx}{\linewidth}{@{}l|*{3}{>{\centering\arraybackslash}X}|>{\centering\arraybackslash}X@{}}
\toprule
\textbf{Model Name} & EG$\uparrow$ & PR$\uparrow$ & VQ$\uparrow$ & \textbf{Overall}$\uparrow$ \\
\midrule
\rowcolor{TableTint}\multicolumn{5}{c}{\itshape Closed-source Models} \\
GPT-Image-2 & \textbf{4.453} & \textbf{4.222} & \textbf{3.795} & \textbf{4.229} \\
Nano-Banana-Pro & 4.060 & 3.966 & 3.368 & 3.884 \\
Seedream-5.0-Pro & \underline{4.272} & \underline{4.123} & \underline{3.658} & \underline{4.089} \\
\midrule
\rowcolor{TableTint}\multicolumn{5}{c}{\itshape Open-source Models} \\
HunyuanImage-3.0-Instruct-Distil & 3.209 & 3.549 & 2.978 & 3.299 \\
FireRed-Image-Edit-1.1 & 3.026 & 3.316 & 3.043 & 3.145 \\
Qwen-Image-Edit-2511 & 2.880 & 3.179 & 3.145 & 3.053 \\
\bottomrule
\end{tabularx}
\end{table}

\begin{table}[!htbp]
\centering
\caption{Quantitative comparison on \multi{}: Pet Editing. Bold and underline denote the best and second-best scores.}
\label{tab:wegen-multi-primary-pet}
\fontsize{8.5pt}{10pt}\selectfont
\renewcommand{\arraystretch}{1.0}
\setlength{\tabcolsep}{4.5pt}
\begin{tabularx}{\linewidth}{@{}l|*{3}{>{\centering\arraybackslash}X}|>{\centering\arraybackslash}X@{}}
\toprule
\textbf{Model Name} & EG$\uparrow$ & PR$\uparrow$ & VQ$\uparrow$ & \textbf{Overall}$\uparrow$ \\
\midrule
\rowcolor{TableTint}\multicolumn{5}{c}{\itshape Closed-source Models} \\
GPT-Image-2 & \textbf{4.579} & \textbf{4.211} & \underline{3.833} & \textbf{4.282} \\
Nano-Banana-Pro & 4.357 & 4.018 & 3.384 & 4.027 \\
Seedream-5.0-Pro & \underline{4.463} & \underline{4.176} & \textbf{3.926} & \underline{4.241} \\
\midrule
\rowcolor{TableTint}\multicolumn{5}{c}{\itshape Open-source Models} \\
HunyuanImage-3.0-Instruct-Distil & 3.787 & 3.734 & 3.468 & 3.702 \\
FireRed-Image-Edit-1.1 & 2.912 & 2.895 & 3.316 & 2.986 \\
Qwen-Image-Edit-2511 & 2.816 & 2.974 & 3.421 & 3.000 \\
\bottomrule
\end{tabularx}
\end{table}

\begin{table}[!htbp]
\centering
\caption{Quantitative comparison on \multi{}: Aesthetic and Photographic Editing. Bold and underline denote the best and second-best scores.}
\label{tab:wegen-multi-primary-aesthetic}
\fontsize{8.5pt}{10pt}\selectfont
\renewcommand{\arraystretch}{1.0}
\setlength{\tabcolsep}{4.5pt}
\begin{tabularx}{\linewidth}{@{}l|*{3}{>{\centering\arraybackslash}X}|>{\centering\arraybackslash}X@{}}
\toprule
\textbf{Model Name} & EG$\uparrow$ & PR$\uparrow$ & VQ$\uparrow$ & \textbf{Overall}$\uparrow$ \\
\midrule
\rowcolor{TableTint}\multicolumn{5}{c}{\itshape Closed-source Models} \\
GPT-Image-2 & \textbf{4.455} & \textbf{4.223} & \underline{3.955} & \textbf{4.263} \\
Nano-Banana-Pro & 3.955 & 4.153 & 3.739 & 3.991 \\
Seedream-5.0-Pro & \underline{4.216} & \underline{4.171} & \textbf{4.000} & \underline{4.155} \\
\midrule
\rowcolor{TableTint}\multicolumn{5}{c}{\itshape Open-source Models} \\
HunyuanImage-3.0-Instruct-Distil & 3.255 & 3.798 & 3.287 & 3.479 \\
FireRed-Image-Edit-1.1 & 3.044 & 3.195 & 3.319 & 3.159 \\
Qwen-Image-Edit-2511 & 2.646 & 3.000 & 3.204 & 2.899 \\
\bottomrule
\end{tabularx}
\end{table}

\clearpage
\subsection{Complete Task Taxonomy}
\label{app:taxonomy}
\subsubsection{Single-Image Task Taxonomy}
\label{app:taxonomy-single}

\begingroup
\fontsize{8.5}{10}\selectfont
\setlength{\tabcolsep}{4pt}
\renewcommand{\arraystretch}{1.0}
\setlength{\LTleft}{0pt}
\setlength{\LTright}{0pt}
\begin{longtable}{@{}>{\raggedright\arraybackslash}p{0.30\linewidth}>{\raggedright\arraybackslash}p{\dimexpr0.70\linewidth-2\tabcolsep\relax}@{}}
\multicolumn{2}{@{}p{\textwidth}@{}}{\small\TableCaptionLabel{Table~\thetable}\enspace Single-Image Task Taxonomy. Family headings give unique covered instances; rows define the second-level tasks. The subset has 1,008 instances and 65 task labels.}\label{tab:taxonomy-single}\\[3pt]
\toprule
\textbf{Second-level task} & \textbf{Operational definition} \\
\midrule
\endfirsthead
\multicolumn{2}{@{}l}{\small\textit{Table~\ref{tab:taxonomy-single} continued.}}\\
\toprule
\textbf{Second-level task} & \textbf{Operational definition} \\
\midrule
\endhead
\midrule
\multicolumn{2}{r@{}}{\small\textit{Continued on next page.}}\\
\endfoot
\bottomrule
\endlastfoot
\multicolumn{2}{@{}p{\linewidth}@{}}{\textbf{Basic local editing} \hfill 631 unique instances}\\*
Object addition & Insert a specified object with correct location, scale, and occlusion. \\
Object removal & Remove a target, reconstruct its background, and preserve surrounding content. \\
Object replacement & Replace a target while respecting its scene role. \\
Object count change & Change the specified objects to the requested count. \\
Local attribute editing & Change target color or attributes within the requested scope. \\
Material replacement & Change target material while retaining geometry and identity. \\
Motion-state editing & Change motion state and depict requested motion cues. \\
Texture-detail editing & Modify surface patterns or details while preserving carrier structure. \\
Size or shape adjustment & Adjust target proportions or shape with coherent surrounding geometry. \\
Background replacement & Replace the background while preserving foreground identity and boundaries. \\
Subject extraction & Extract the target with clean boundaries and the requested background. \\
Image outpainting & Extend the canvas with compatible scene content. \\
Regional repainting & Regenerate a specified region while preserving content outside it. \\
Political-symbol editing & Insert or modify requested political symbols within the source scene. \\
\midrule
\multicolumn{2}{@{}p{\linewidth}@{}}{\textbf{Portrait editing} \hfill 309 unique instances}\\*
Portrait retouching & Retouch facial appearance while preserving identity and natural structure. \\
Makeup editing & Modify makeup while preserving facial identity and anatomy. \\
Hairstyle change & Change hairstyle while retaining other required subject attributes. \\
Hair-color change & Recolor hair while preserving its shape, face, and surroundings. \\
Facial-expression editing & Change facial expression while preserving identity and facial structure. \\
Clothing-style editing & Restyle clothing from one input while preserving the person. \\
Pose editing & Change body or limb configuration with plausible anatomy. \\
Formal or professional portrait & Present a portrait in the requested formal or professional style. \\
Portrait or role-play styling & Style a portrait for a specified role, character, or concept. \\
Themed portrait styling & Coordinate attire, props, and setting for a portrait theme. \\
Photo-print or album rendering & Render source content as a photograph, instant print, or album. \\
Animalized or creative face editing & Integrate requested animal-like or creative facial transformations. \\
Age editing & Change apparent age while preserving identity cues and scene content. \\
Surreal portrait editing & Add surreal elements while preserving the intended portrait composition. \\
Named-character editing & Introduce or evoke a specified recognizable character or persona. \\
\midrule
\multicolumn{2}{@{}p{\linewidth}@{}}{\textbf{Aesthetic and photographic editing} \hfill 320 unique instances}\\*
Global stylization & Change global style while preserving requested semantic content. \\
Color-tone adjustment & Adjust palette or grading while preserving required scene details. \\
Relighting & Modify illumination, highlights, and shadows to match requested lighting. \\
Depth-of-field editing & Adjust focus and background blur around the intended subject. \\
Framing or zoom adjustment & Change framing or shot scale while preserving scene relationships. \\
Photographic-view refinement & Refine viewpoint or composition while maintaining scene plausibility. \\
Subject-view rotation & Rotate the target view while preserving recognizable structure. \\
Material-detail enhancement & Enhance texture or gloss while preserving material identity. \\
Sketch-to-image conversion & Render a sketch into an image consistent with its structure. \\
Drawing colorization or rendering & Color or render artwork while retaining contours and composition. \\
Fisheye selfie rendering & Create a selfie composition with coherent fisheye distortion. \\
\midrule
\multicolumn{2}{@{}p{\linewidth}@{}}{\textbf{Product and commercial editing} \hfill 117 unique instances}\\*
White-background product image & Place a product on white while preserving its details. \\
Transparent-background product image & Extract a product onto transparency with clean contours. \\
Product-in-scene rendering & Restage a product in context while preserving its appearance. \\
Studio product photography & Create studio product photography with controlled lighting and materials. \\
Exploded product view & Separate or expose product components in the requested explanatory arrangement. \\
Creative product advertising & Create a stylized advertisement or object-inspired product arrangement. \\
Dieline-to-package rendering & Render a flat packaging layout as a coherent 3D package. \\
\midrule
\multicolumn{2}{@{}p{\linewidth}@{}}{\textbf{Action and interaction editing} \hfill 64 unique instances}\\*
Person--person interaction & Edit interpersonal interactions while preserving roles and contact relationships. \\
Sports or dynamic action & Depict a requested dynamic action with coherent body configuration. \\
Person--scene interaction & Edit person--scene interactions, including support and contact. \\
Hand--object interaction & Change a held object or grasp with plausible finger contact. \\
\midrule
\multicolumn{2}{@{}p{\linewidth}@{}}{\textbf{Reasoning-intensive editing} \hfill 164 unique instances}\\*
Spatial-reasoning editing & Execute edits requiring spatial relations or relative-position reasoning. \\
Coordinated multi-object editing & Coordinate changes to multiple objects under a shared objective. \\
Relational editing & Change a relationship between entities, not just their appearance. \\
Physical-state editing & Change physical state and depict its visible consequences. \\
Causal-scene editing & Depict an event and its visually coherent consequences. \\
Idiom-grounded editing & Translate colloquial or idiomatic intent into a visual edit. \\
Composite-operation editing & Coordinate multiple transformations and their interacting constraints. \\
Negative or preservation constraints & Honor explicit prohibitions or invariances while executing edits. \\
\midrule
\multicolumn{2}{@{}p{\linewidth}@{}}{\textbf{Restoration and low-level enhancement} \hfill 67 unique instances}\\*
Old-photo restoration & Repair old-photo degradation while retaining original subject content. \\
Deblurring & Reduce blur while preserving structure and avoiding unsupported details. \\
Denoising & Suppress noise while preserving meaningful texture and edges. \\
Exposure correction & Correct exposure while maintaining coherent color and contrast. \\
Super-resolution enhancement & Enhance resolution and detail while preserving image content. \\
\midrule
\multicolumn{2}{@{}p{\linewidth}@{}}{\textbf{Traditional Chinese cultural editing} \hfill 33 unique instances}\\*
Traditional Chinese cultural editing & Edit requested traditional cultural motifs, attire, objects, or styles. \\
\addlinespace[3pt]
\end{longtable}
\endgroup

\subsubsection{Text-Intensive Task Taxonomy}
\label{app:taxonomy-text}

\begingroup
\fontsize{8.5}{10}\selectfont
\setlength{\tabcolsep}{4pt}
\renewcommand{\arraystretch}{1.0}
\setlength{\LTleft}{0pt}
\setlength{\LTright}{0pt}
\begin{longtable}{@{}>{\raggedright\arraybackslash}p{0.30\linewidth}>{\raggedright\arraybackslash}p{\dimexpr0.70\linewidth-2\tabcolsep\relax}@{}}
\multicolumn{2}{@{}p{\textwidth}@{}}{\small\TableCaptionLabel{Table~\thetable}\enspace Text-Intensive Task Taxonomy. Family headings give unique covered instances; rows define the second-level tasks. The subset has 1,000 instances and 35 task labels.}\label{tab:taxonomy-text}\\[3pt]
\toprule
\textbf{Second-level task} & \textbf{Operational definition} \\
\midrule
\endfirsthead
\multicolumn{2}{@{}l}{\small\textit{Table~\ref{tab:taxonomy-text} continued.}}\\
\toprule
\textbf{Second-level task} & \textbf{Operational definition} \\
\midrule
\endhead
\midrule
\multicolumn{2}{r@{}}{\small\textit{Continued on next page.}}\\
\endfoot
\bottomrule
\endlastfoot
\multicolumn{2}{@{}p{\linewidth}@{}}{\textbf{Basic text editing} \hfill 678 unique instances}\\*
Text-content replacement & Replace specified wording while preserving placement and surrounding content. \\
Text insertion & Insert wording without obscuring protected visual information. \\
Text removal & Remove text, reconstruct its carrier, and preserve non-target content. \\
Typography preservation & Change wording while preserving specified typography and arrangement. \\
Typography-attribute editing & Change typography while preserving wording and required structure. \\
Multilingual text replacement & Replace text in another language and adapt its layout. \\
Perspective-aware text editing & Edit surface text with correct perspective, curvature, lighting, and material. \\
Logo or symbol replacement & Replace a logo, icon, or symbol within its carrier. \\
Text-effect design & Apply decorative text effects while preserving character legibility. \\
Watermark insertion or removal & Add or remove a watermark while preserving required scene content. \\
Typographical correction & Correct erroneous characters while preserving surrounding text and presentation. \\
\midrule
\multicolumn{2}{@{}p{\linewidth}@{}}{\textbf{Interfaces, documents, and charts} \hfill 177 unique instances}\\*
Interface-copy editing & Edit interface copy while preserving components, spacing, and alignment. \\
Table or document editing & Edit document fields or cells while retaining unrelated content and rules. \\
Chart-label editing & Edit chart labels while preserving the plot unless otherwise requested. \\
Visual-question completion & Solve a depicted question and write the answer where specified. \\
Quantity or price comparison & Compare visible quantities and mark or write the requested result. \\
Interface mockup design & Design an interface with specified copy, components, and hierarchy. \\
\midrule
\multicolumn{2}{@{}p{\linewidth}@{}}{\textbf{Commercial text--image design} \hfill 110 unique instances}\\*
Product-advertising poster & Create a product poster with specified copy, layout, and product fidelity. \\
Packaging-text editing & Edit packaging text while preserving geometry and non-target information. \\
Product-feature labels & Add product labels or callouts without obscuring protected details. \\
E-commerce detail module & Organize product information into an illustrated detail-page module. \\
\midrule
\multicolumn{2}{@{}p{\linewidth}@{}}{\textbf{Posters and visual layouts} \hfill 221 unique instances}\\*
Xiaohongshu-style cover & Create a social cover with prominent titles and a clear subject. \\
Event or course poster & Organize event or course information into a poster hierarchy. \\
Seasonal marketing poster & Compose seasonal imagery and promotional copy. \\
Magazine-cover design & Arrange cover titles and supporting text around the source subject. \\
Infographic design & Organize information into visual sections, labels, and explanations. \\
Timeline graphic & Arrange events or stages sequentially with associated text and imagery. \\
Map or world-layout labeling & Create or edit place names, regions, legends, or map annotations. \\
Before--after comparison layout & Present requested states in a labeled comparison layout. \\
Poster-layout rearrangement & Rearrange poster hierarchy while preserving key visuals and copy. \\
Format or platform adaptation & Adapt format while retaining important content and readability. \\
\midrule
\multicolumn{2}{@{}p{\linewidth}@{}}{\textbf{Social-media graphics} \hfill 96 unique instances}\\*
Venue or recommendation graphic & Create scene-grounded recommendation graphics with titles, tags, or highlights. \\
Outfit-description graphic & Add outfit labels or commentary without obscuring clothing. \\
Meme or sticker design & Combine a subject with wording matching its expression. \\
Pet portrait or identity card & Create a pet portrait card with names or descriptive text. \\
\addlinespace[3pt]
\end{longtable}
\endgroup

\subsubsection{Multi-Reference Task Taxonomy}
\label{app:taxonomy-multi}

\begingroup
\fontsize{8.5}{10}\selectfont
\setlength{\tabcolsep}{4pt}
\renewcommand{\arraystretch}{1.0}
\setlength{\LTleft}{0pt}
\setlength{\LTright}{0pt}
\begin{longtable}{@{}>{\raggedright\arraybackslash}p{0.30\linewidth}>{\raggedright\arraybackslash}p{\dimexpr0.70\linewidth-2\tabcolsep\relax}@{}}
\multicolumn{2}{@{}p{\textwidth}@{}}{\small\TableCaptionLabel{Table~\thetable}\enspace Multi-Reference Task Taxonomy. Family headings give unique covered instances; rows define the second-level tasks. The subset has 732 instances and 48 task labels.}\label{tab:taxonomy-multi}\\[3pt]
\toprule
\textbf{Second-level task} & \textbf{Operational definition} \\
\midrule
\endfirsthead
\multicolumn{2}{@{}l}{\small\textit{Table~\ref{tab:taxonomy-multi} continued.}}\\
\toprule
\textbf{Second-level task} & \textbf{Operational definition} \\
\midrule
\endhead
\midrule
\multicolumn{2}{r@{}}{\small\textit{Continued on next page.}}\\
\endfoot
\bottomrule
\endlastfoot
\multicolumn{2}{@{}p{\linewidth}@{}}{\textbf{Portrait editing} \hfill 150 unique instances}\\*
Face transfer & Transfer facial identity while preserving required body, pose, and scene. \\
Reference-based clothing transfer & Transfer reference clothing while preserving the target person's identity. \\
Hairstyle transfer & Transfer hairstyle without substituting unrelated identity attributes. \\
Expression transfer & Transfer expression while preserving the target person's identity. \\
Jewelry try-on & Fit reference jewelry with correct scale, contact, and occlusion. \\
Makeup and styling transfer & Transfer specified reference makeup or styling to the target. \\
Multi-person interaction & Compose referenced people with correct identities, roles, and interactions. \\
Person--scene composition & Place a referenced person with coherent pose, contact, and perspective. \\
Person--pet interaction & Compose a person and pet with correct reference assignments. \\
Surreal portrait composition & Combine portraits and surreal elements with correct subject assignments. \\
\midrule
\multicolumn{2}{@{}p{\linewidth}@{}}{\textbf{Product and commercial editing} \hfill 120 unique instances}\\*
Artwork-to-packaging transfer & Transfer artwork or text onto packaging with correct folds and perspective. \\
Character-themed commercial display & Integrate referenced characters or branding into a commercial display. \\
Product--scene composition & Compose a reference product while preserving identity and physical placement. \\
Model--product--scene composition & Bind person, product, and scene to their designated references. \\
Dieline-guided package rendering & Render flat artwork and reference information as a 3D package. \\
Image-to-device display transfer & Insert imagery into a device display with correct crop and perspective. \\
Branding transfer & Apply reference branding while preserving its visual identity. \\
Multi-product composition & Arrange referenced products with correct identities and relationships. \\
\midrule
\multicolumn{2}{@{}p{\linewidth}@{}}{\textbf{Scene and spatial editing} \hfill 117 unique instances}\\*
Scene-element addition & Add reference elements at requested target-scene locations. \\
Scene-element replacement & Replace scene elements with reference content while preserving surroundings. \\
Reference-based background replacement & Transfer a reference background while retaining the required foreground. \\
Scene-viewpoint transformation & Reconstruct a scene from a reference-specified viewpoint. \\
Scene-style transfer & Transfer scene style while preserving target content and structure. \\
Weather or season transfer & Transfer weather or season while preserving identities and layout. \\
Scene-lighting or time transfer & Transfer reference lighting or time-of-day appearance. \\
Plan-to-scene rendering & Render a scene consistent with a reference map or floor plan. \\
\midrule
\multicolumn{2}{@{}p{\linewidth}@{}}{\textbf{Aesthetic and photographic editing} \hfill 113 unique instances}\\*
Reference-style transfer & Transfer reference style while preserving requested target content. \\
Reference-guided line-art coloring & Color line art from references while respecting its structure. \\
Reference-material transfer & Transfer reference material appearance onto a target object. \\
Reference-guided relighting & Transfer illumination with plausible lighting on the target. \\
Photographic-viewpoint transfer & Transfer photographic viewpoint while preserving required subject details. \\
Composition transfer & Transfer arrangement or framing while preserving target subjects. \\
Color-palette transfer & Transfer a reference palette while preserving required content. \\
Lens or focus-effect transfer & Transfer focus or lens effects while preserving focal content. \\
\midrule
\multicolumn{2}{@{}p{\linewidth}@{}}{\textbf{Pet editing} \hfill 114 unique instances}\\*
Pet portrait styling & Style a pet portrait while preserving the pet's identity. \\
Pet-pattern transfer & Transfer reference markings to the specified pet region. \\
Pet clothing transfer & Transfer pet clothing with coherent fit and preserved identity. \\
Pet-to-toy rendering & Render a pet as a reference-guided toy retaining identity cues. \\
Pet--scene composition & Compose a pet and scene with coherent scale, lighting, and contact. \\
Pet-expression transfer & Transfer pet expression while preserving identity and anatomy. \\
Pet-pose transfer & Transfer pet pose while preserving identity. \\
\midrule
\multicolumn{2}{@{}p{\linewidth}@{}}{\textbf{Text editing} \hfill 118 unique instances}\\*
Full-image watermark transfer & Transfer a watermark with the requested coverage and appearance. \\
Interface or screen-text transfer & Transfer interface or display text with correct perspective and alignment. \\
Reference-guided text--image layout & Rearrange text and images using designated layout references. \\
Reference-guided text rendering & Render reference-guided text while preserving correct characters. \\
Reference-based meme generation & Combine reference content and wording into an expressive meme. \\
Multilingual text transfer & Transfer or translate reference text and adapt its presentation. \\
Typeface-reference transfer & Transfer reference typography while retaining requested wording. \\
\addlinespace[3pt]
\end{longtable}
\endgroup

\clearpage
\section{Qualitative Case Studies}
\label{app:case-studies}
\label{app:case-format}
Sections~\ref{app:cases-single}--\ref{app:cases-multi} present 32 benchmark cases in the order \single{} (12), \textedit{} (12), and \multi{} (8), with at most two panels per page. Each panel includes the shown instruction, source/reference images, and outputs from the labeled models. Multi-reference inputs retain their ordering so that identity, appearance, and position can be checked against the intended source.

\subsection{\single{}: Single-Image Editing Cases}
\label{app:cases-single}
Inspect the requested transformation together with non-target objects, local geometry, and surrounding appearance.
\begin{figure}[!htbp]
\centering
\setlength{\abovecaptionskip}{4pt}
\setlength{\belowcaptionskip}{2pt}
\includegraphics[page=1,trim=12bp 12bp 12bp 12bp,clip,width=\linewidth]{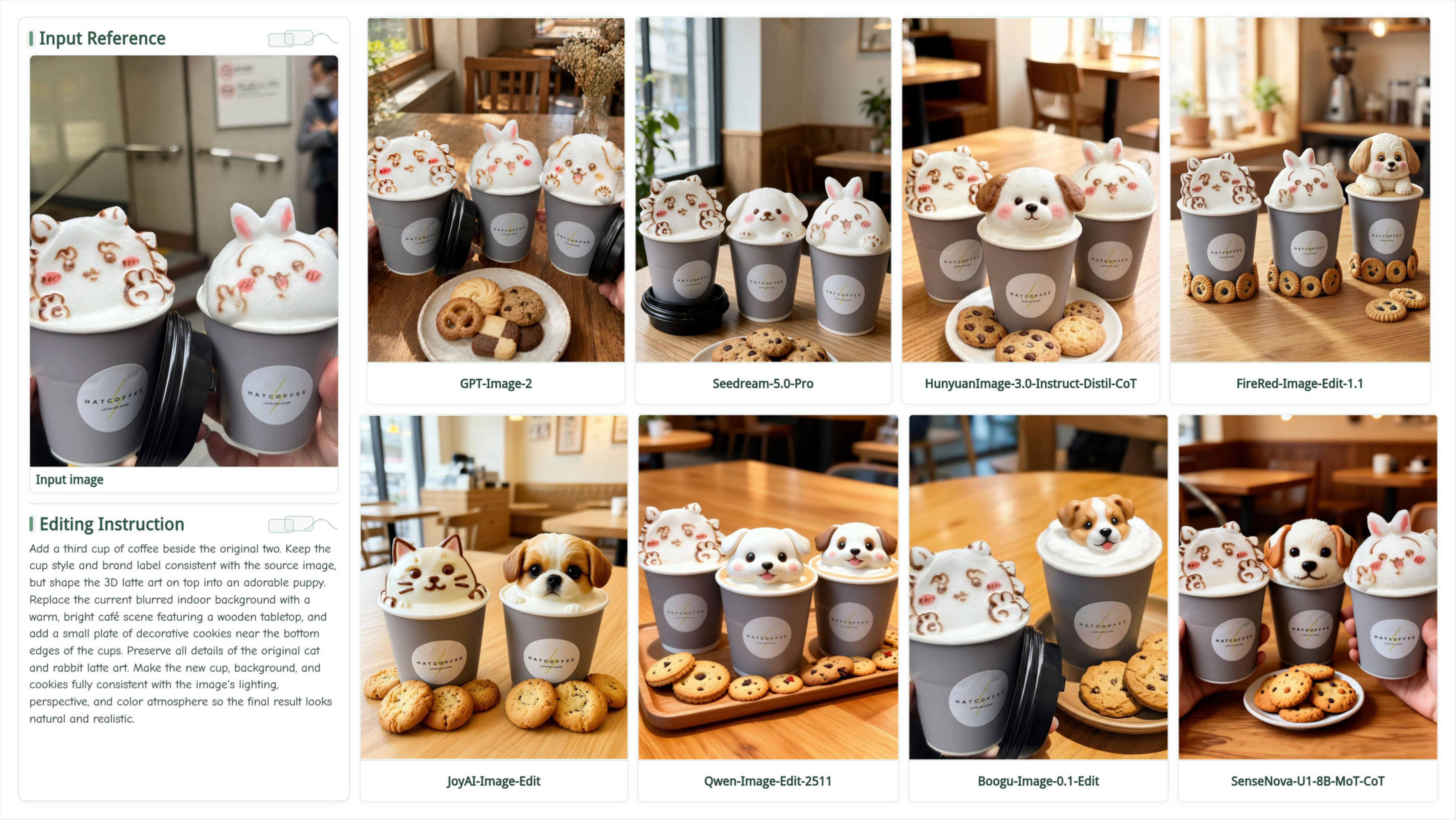}
\caption{\single{}: object addition (case 350). The panel shows the input image(s), editing instruction, and outputs from the labeled models.}
\label{fig:sup-s01}
\end{figure}

\begin{figure}[!htbp]
\centering
\setlength{\abovecaptionskip}{4pt}
\setlength{\belowcaptionskip}{2pt}
\includegraphics[page=2,trim=12bp 12bp 12bp 12bp,clip,width=\linewidth]{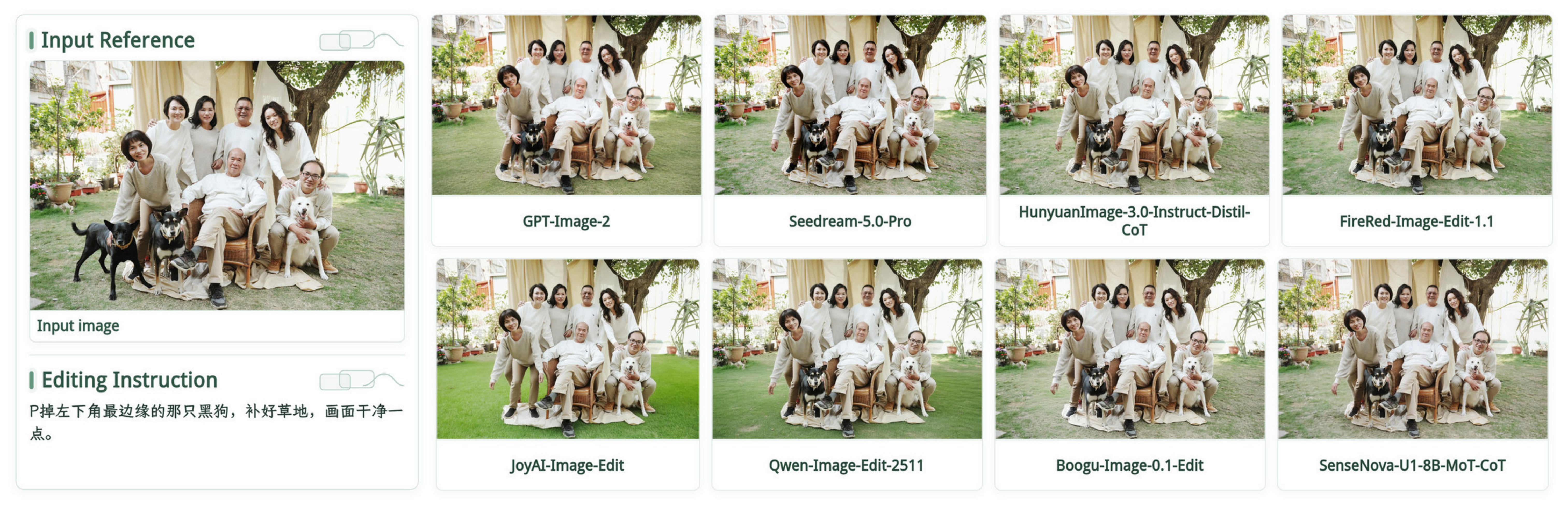}
\caption{\single{}: object removal (case 110). The panel shows the input image(s), editing instruction, and outputs from the labeled models.}
\label{fig:sup-s02}
\end{figure}

\begin{figure}[!htbp]
\centering
\setlength{\abovecaptionskip}{4pt}
\setlength{\belowcaptionskip}{2pt}
\includegraphics[page=3,trim=12bp 12bp 12bp 12bp,clip,width=\linewidth]{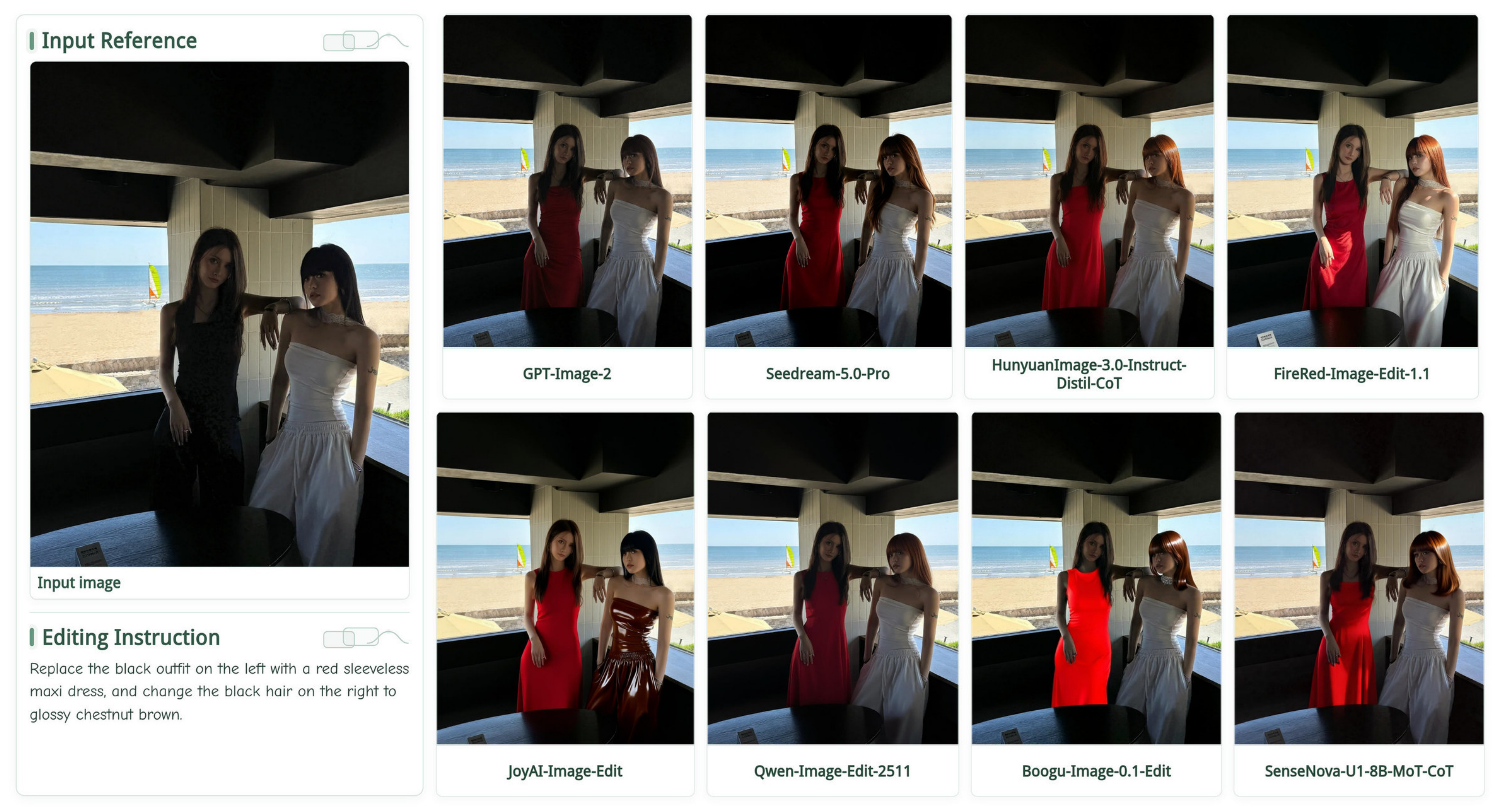}
\caption{\single{}: clothing-style editing (case 700). The panel shows the input image(s), editing instruction, and outputs from the labeled models.}
\label{fig:sup-s03}
\end{figure}

\begin{figure}[!htbp]
\centering
\setlength{\abovecaptionskip}{4pt}
\setlength{\belowcaptionskip}{2pt}
\includegraphics[page=4,trim=12bp 12bp 12bp 12bp,clip,width=\linewidth]{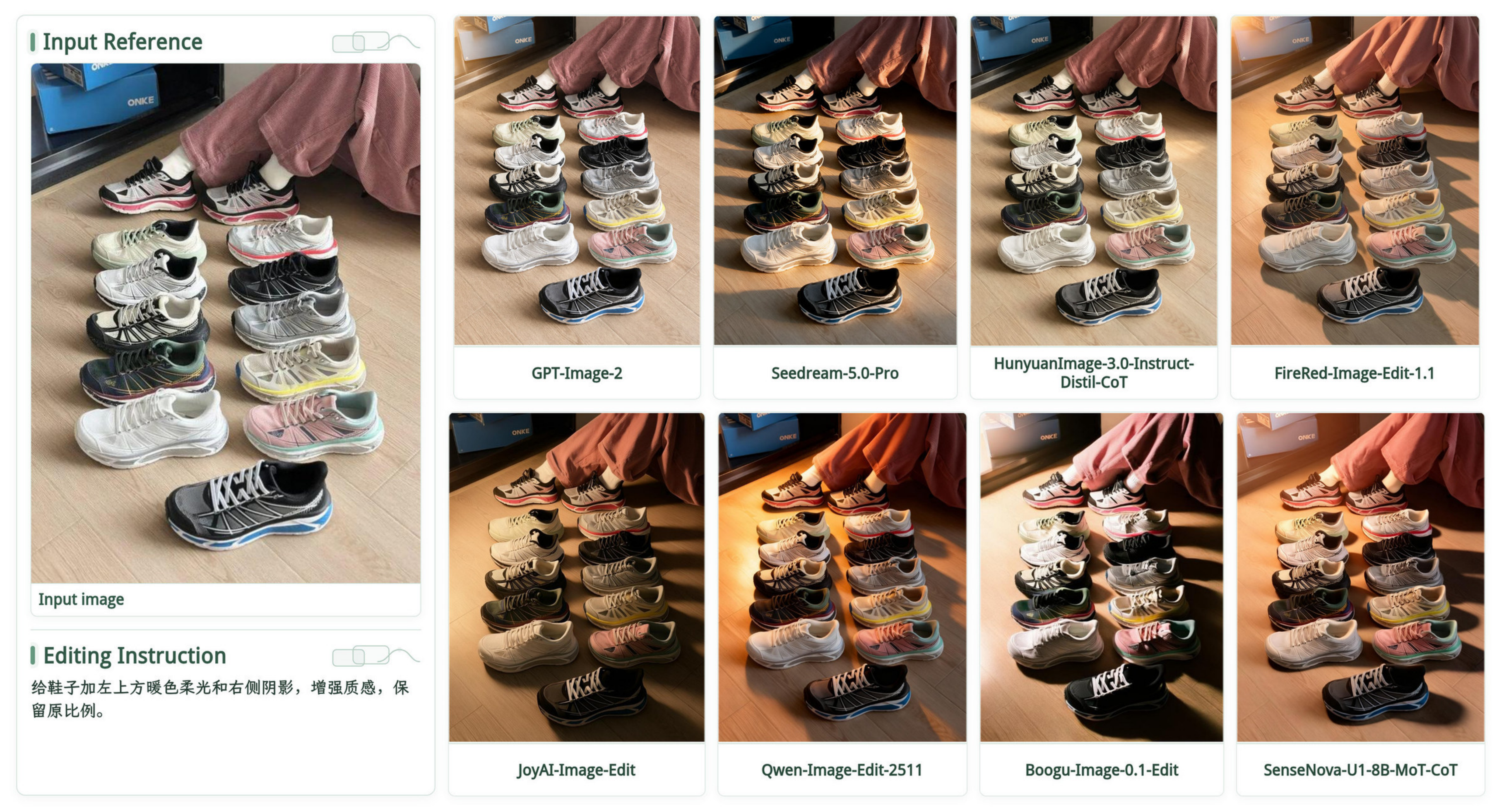}
\caption{\single{}: lighting redesign (case 118). The panel shows the input image(s), editing instruction, and outputs from the labeled models.}
\label{fig:sup-s06}
\end{figure}

\begin{figure}[!htbp]
\centering
\setlength{\abovecaptionskip}{4pt}
\setlength{\belowcaptionskip}{2pt}
\includegraphics[page=5,trim=12bp 12bp 12bp 12bp,clip,width=\linewidth]{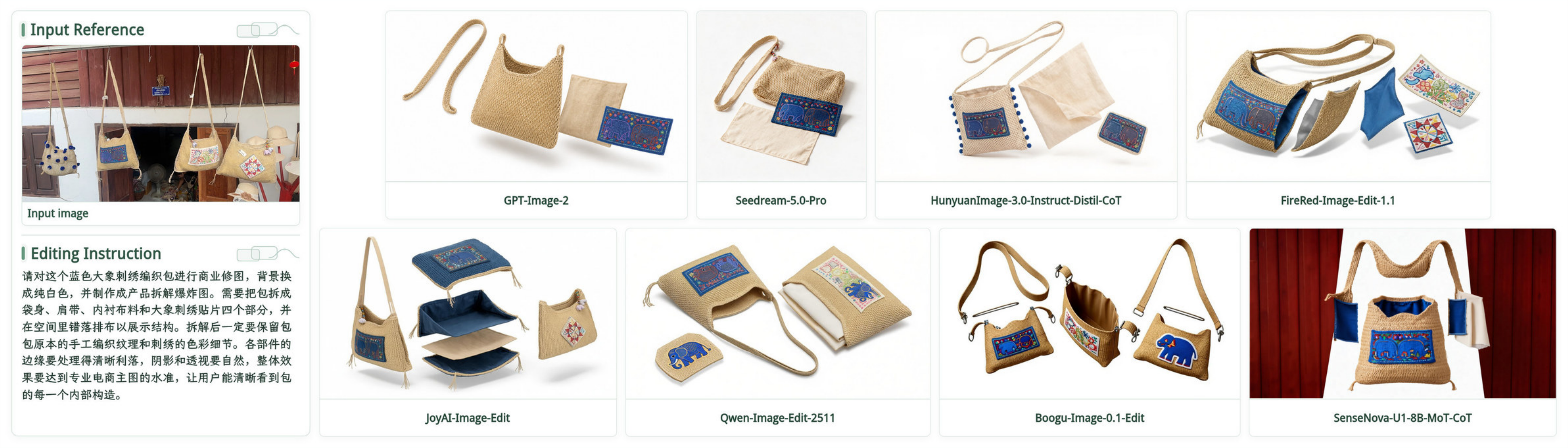}
\caption{\single{}: exploded product view (case 359). The panel shows the input image(s), editing instruction, and outputs from the labeled models.}
\label{fig:sup-s07}
\end{figure}

\begin{figure}[!htbp]
\centering
\setlength{\abovecaptionskip}{4pt}
\setlength{\belowcaptionskip}{2pt}
\includegraphics[page=6,trim=12bp 12bp 12bp 12bp,clip,width=\linewidth]{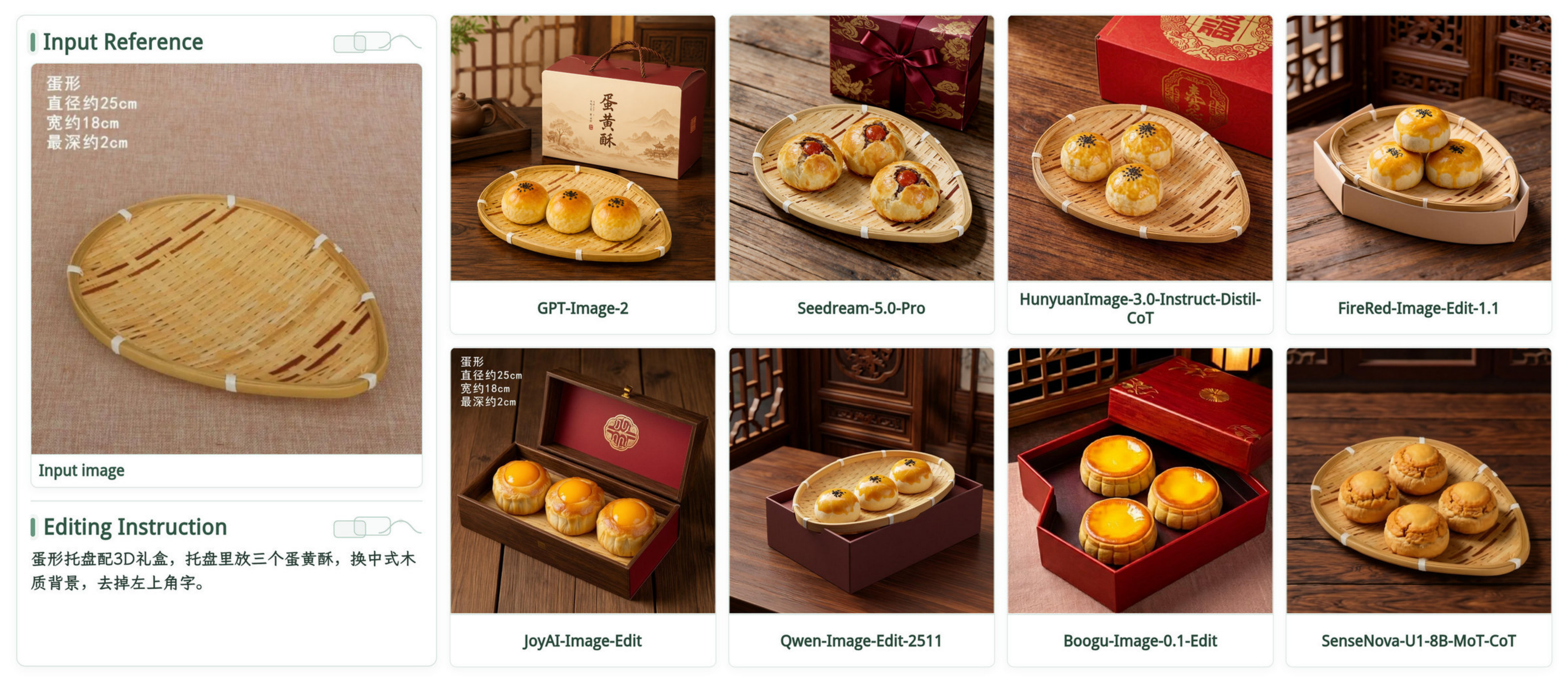}
\caption{\single{}: dieline-to-package rendering (case 495). The panel shows the input image(s), editing instruction, and outputs from the labeled models.}
\label{fig:sup-s08}
\end{figure}

\begin{figure}[!htbp]
\centering
\setlength{\abovecaptionskip}{4pt}
\setlength{\belowcaptionskip}{2pt}
\includegraphics[page=7,trim=12bp 12bp 12bp 12bp,clip,width=\linewidth]{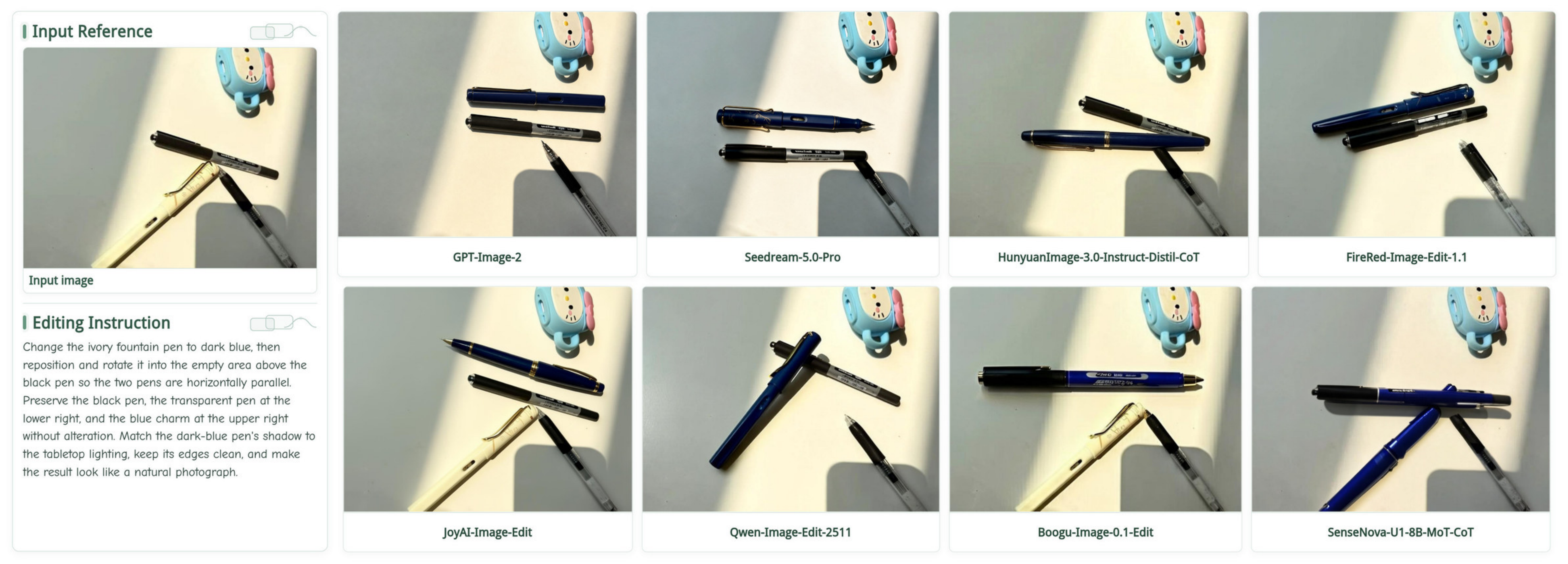}
\caption{\single{}: spatial-reasoning editing (case 142). The panel shows the input image(s), editing instruction, and outputs from the labeled models.}
\label{fig:sup-s09}
\end{figure}

\begin{figure}[!htbp]
\centering
\setlength{\abovecaptionskip}{4pt}
\setlength{\belowcaptionskip}{2pt}
\includegraphics[page=8,trim=12bp 12bp 12bp 12bp,clip,width=\linewidth]{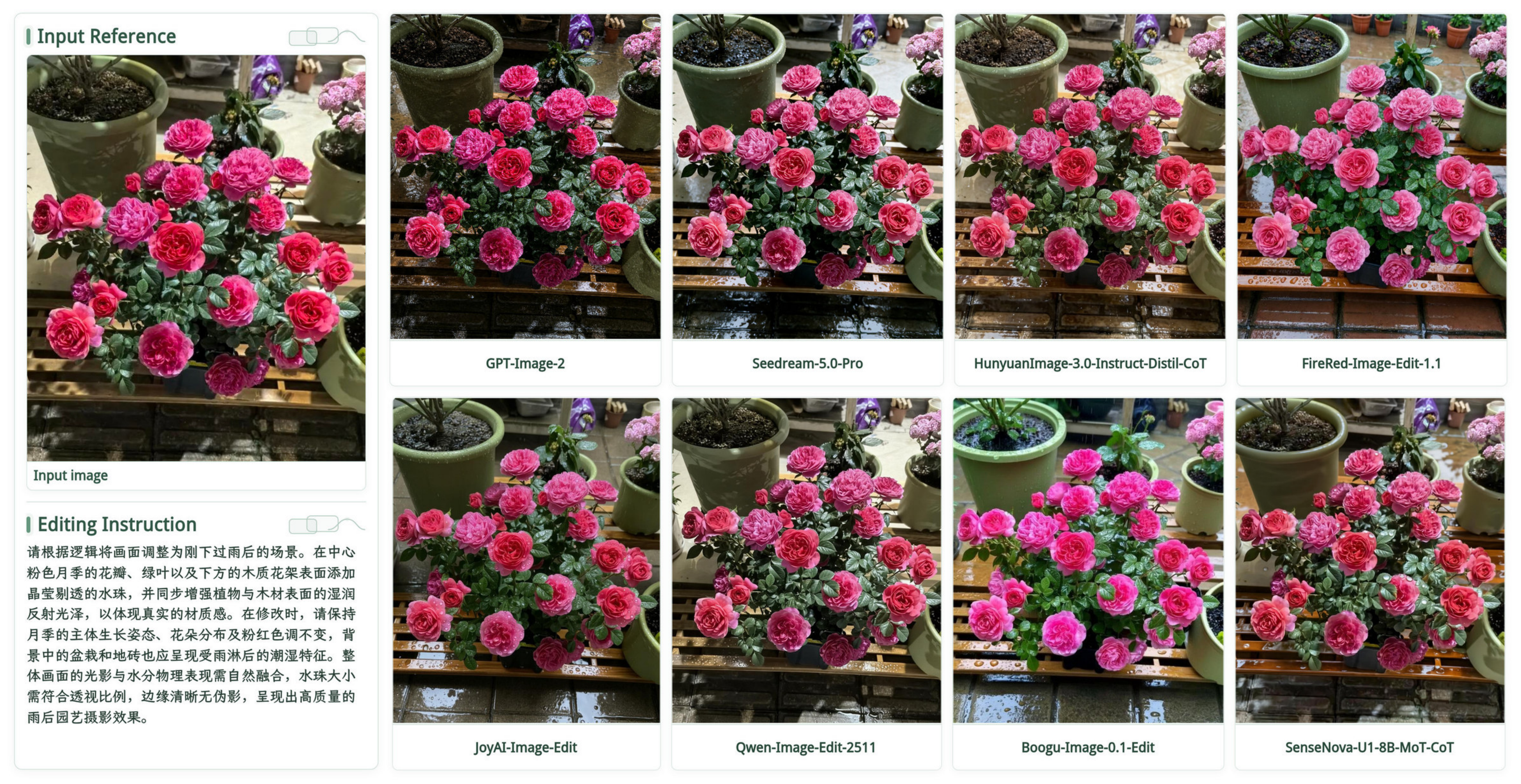}
\caption{\single{}: causal-scene editing (case 272). The panel shows the input image(s), editing instruction, and outputs from the labeled models.}
\label{fig:sup-s10}
\end{figure}

\begin{figure}[!htbp]
\centering
\setlength{\abovecaptionskip}{4pt}
\setlength{\belowcaptionskip}{2pt}
\includegraphics[page=9,trim=12bp 12bp 12bp 12bp,clip,width=\linewidth]{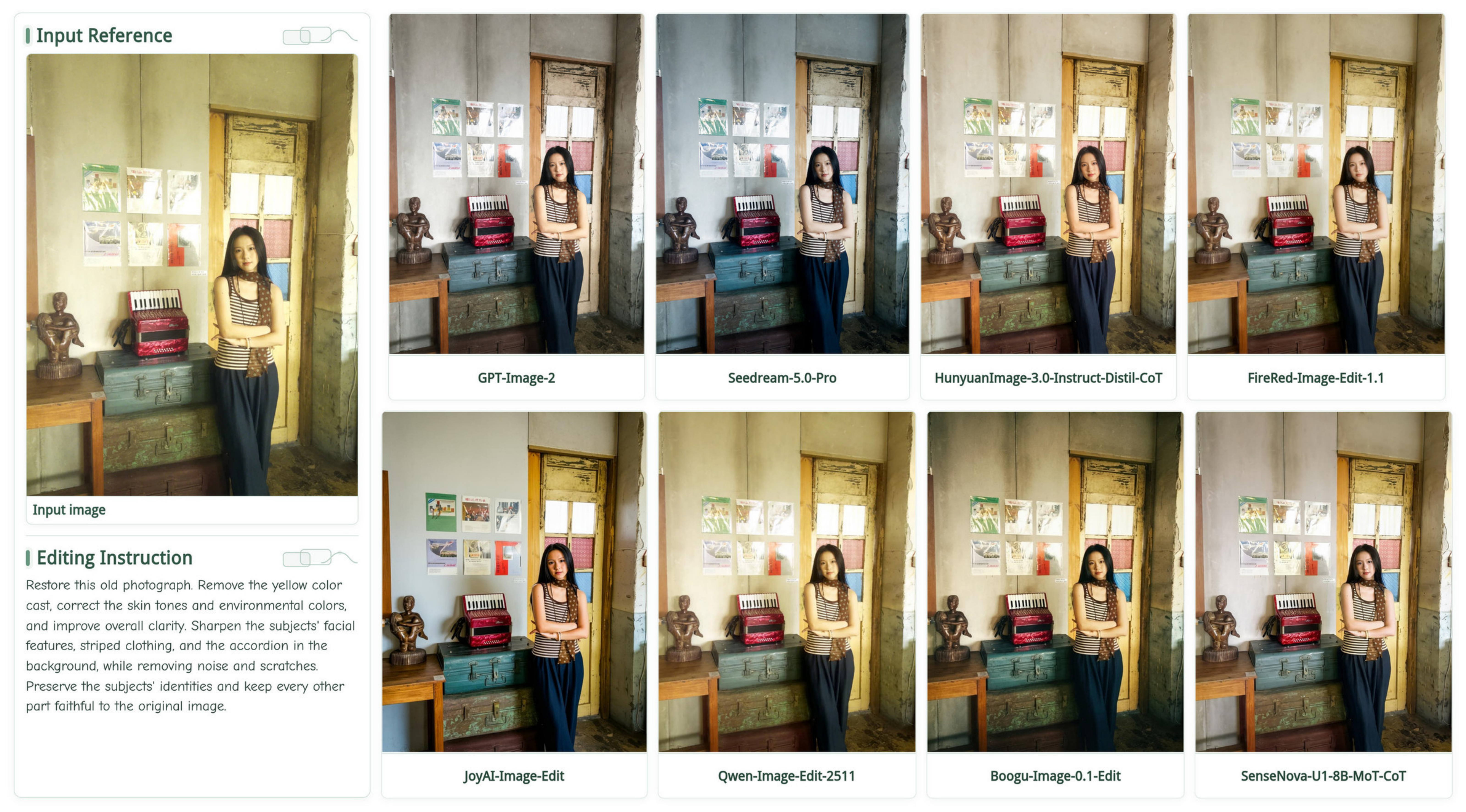}
\caption{\single{}: old-photo restoration (case 356). The panel shows the input image(s), editing instruction, and outputs from the labeled models.}
\label{fig:sup-s11}
\end{figure}

\begin{figure}[!htbp]
\centering
\setlength{\abovecaptionskip}{4pt}
\setlength{\belowcaptionskip}{2pt}
\includegraphics[page=10,trim=12bp 12bp 12bp 12bp,clip,width=\linewidth]{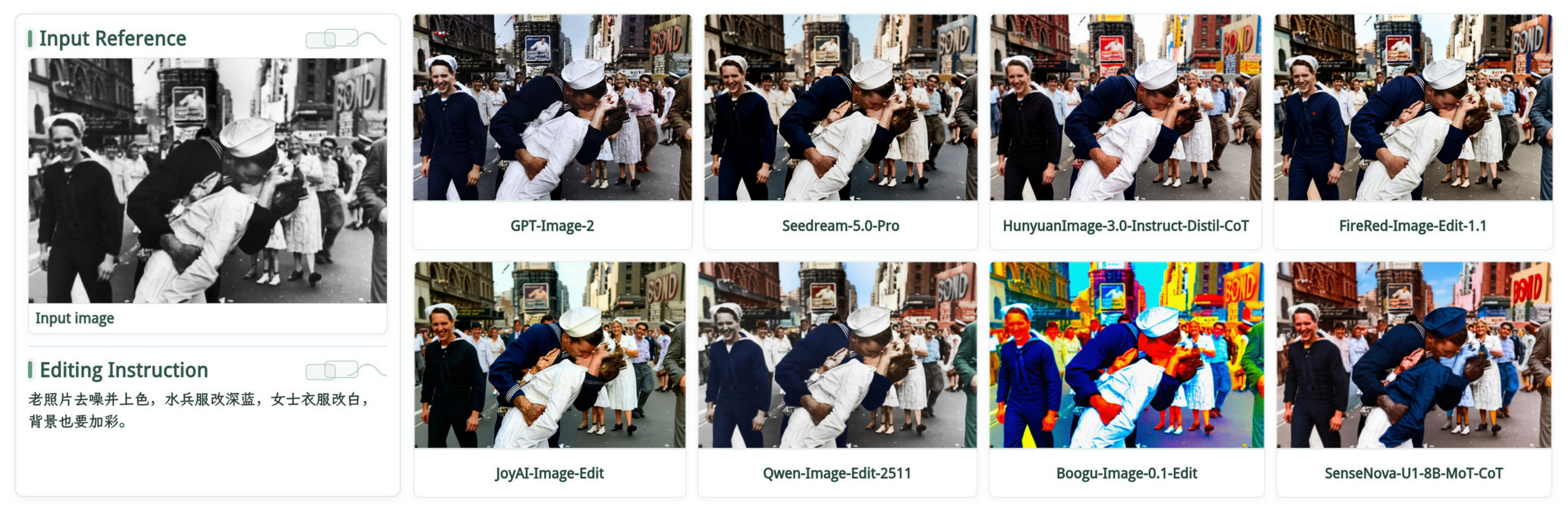}
\caption{\single{}: drawing colorization (case 241). The panel shows the input image(s), editing instruction, and outputs from the labeled models.}
\label{fig:sup-s12}
\end{figure}

\begin{figure}[!htbp]
\centering
\setlength{\abovecaptionskip}{4pt}
\setlength{\belowcaptionskip}{2pt}
\includegraphics[page=11,trim=12bp 12bp 12bp 12bp,clip,width=\linewidth]{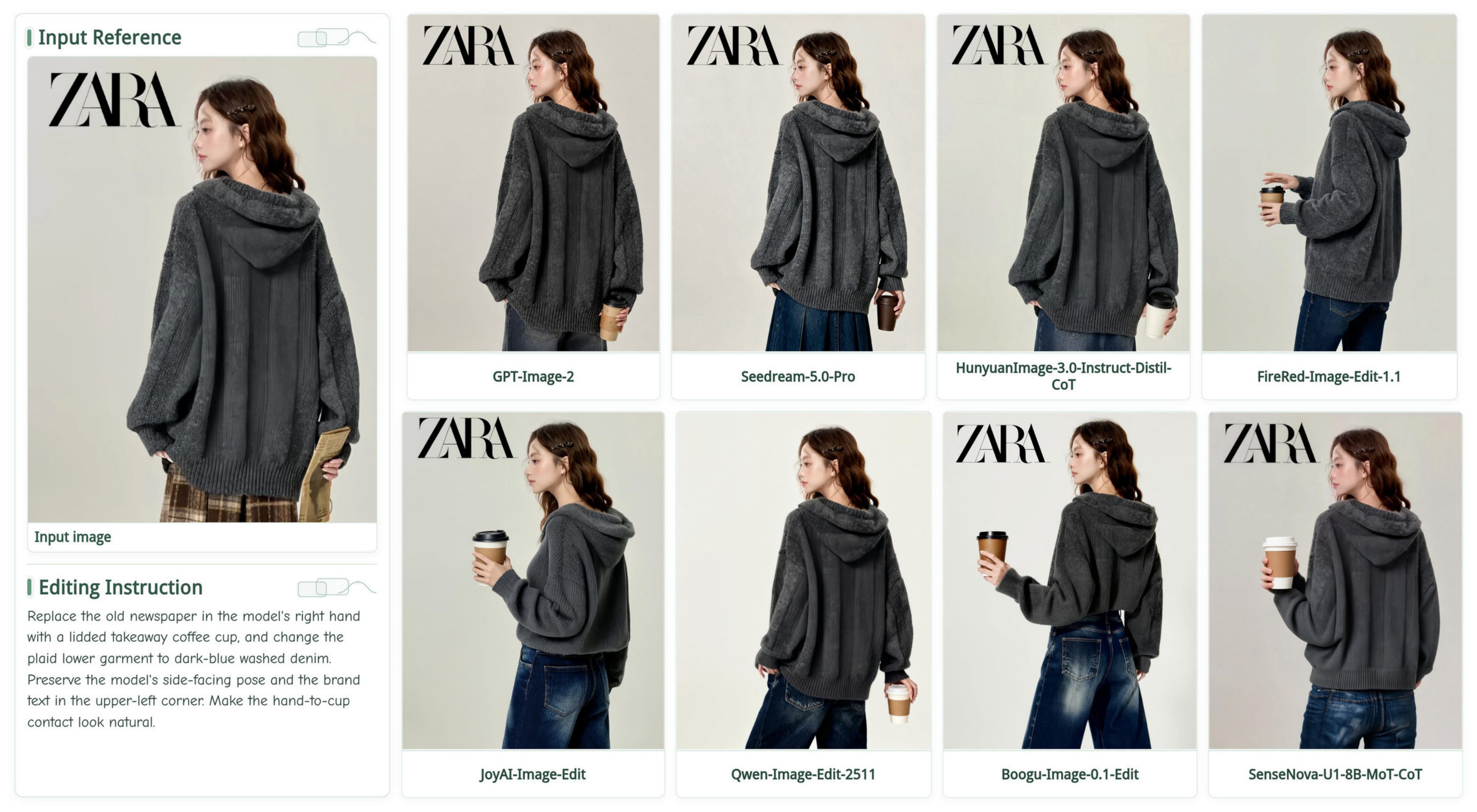}
\caption{\single{}: hand--object interaction (case 661). The panel shows the input image(s), editing instruction, and outputs from the labeled models.}
\label{fig:sup-s14}
\end{figure}

\begin{figure}[!htbp]
\centering
\setlength{\abovecaptionskip}{4pt}
\setlength{\belowcaptionskip}{2pt}
\includegraphics[page=12,trim=12bp 12bp 12bp 12bp,clip,width=\linewidth]{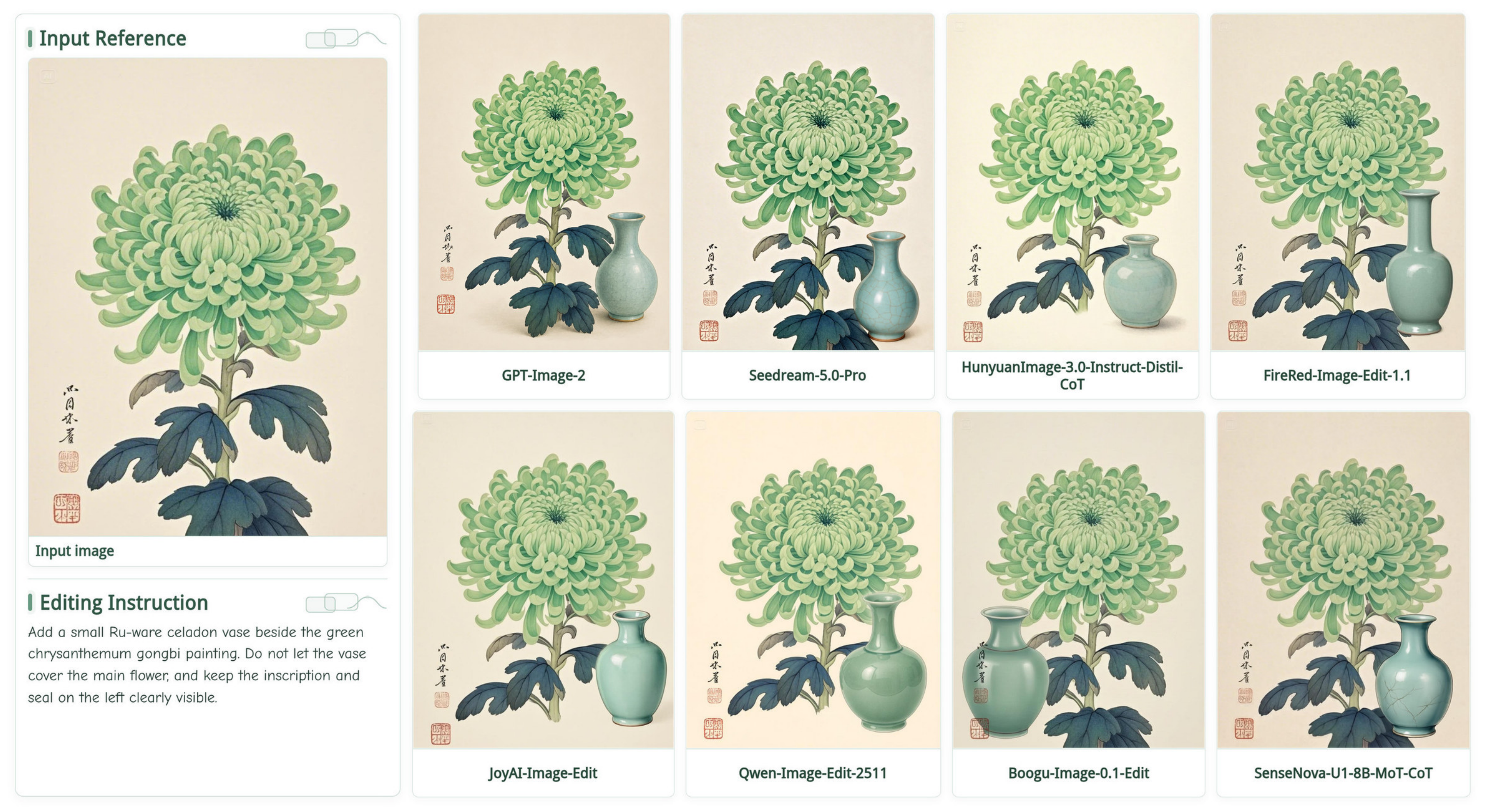}
\caption{\single{}: chinese traditional culture editing (case 550). The panel shows the input image(s), editing instruction, and outputs from the labeled models.}
\label{fig:sup-s15}
\end{figure}

\clearpage
\subsection{\textedit{}: Text-Intensive Editing Cases}
\label{app:cases-text}
Compare literal wording, typography, carrier geometry, and unaffected visual content separately.
\begin{figure}[!htbp]
\centering
\setlength{\abovecaptionskip}{4pt}
\setlength{\belowcaptionskip}{2pt}
\includegraphics[page=1,trim=12bp 12bp 12bp 12bp,clip,width=0.95\linewidth]{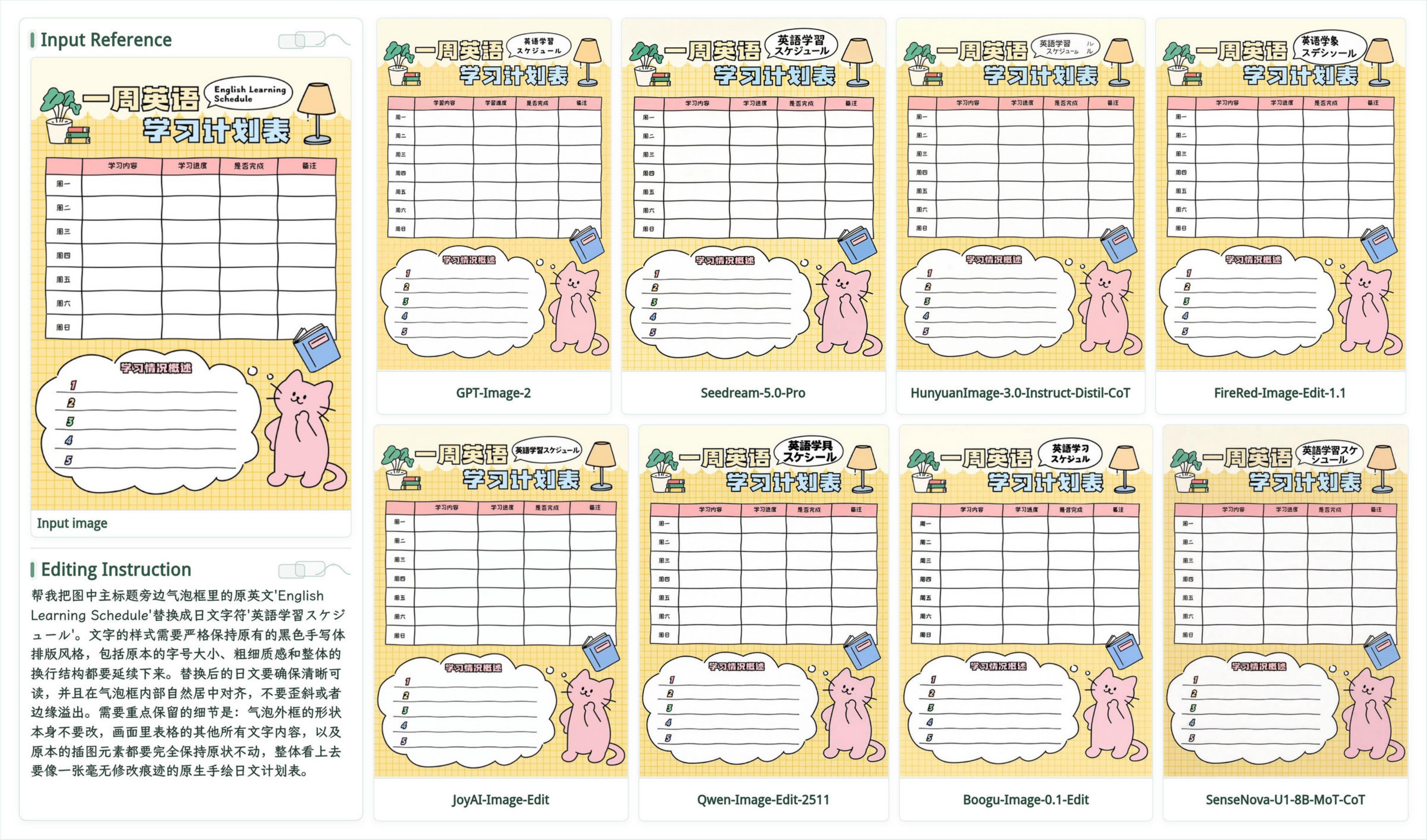}
\caption{\textedit{}: multilingual text replacement (case 418). The panel shows the input image(s), editing instruction, and outputs from the labeled models.}
\label{fig:sup-t01}
\end{figure}

\begin{figure}[!htbp]
\centering
\setlength{\abovecaptionskip}{4pt}
\setlength{\belowcaptionskip}{2pt}
\includegraphics[page=2,trim=12bp 12bp 12bp 12bp,clip,width=0.95\linewidth]{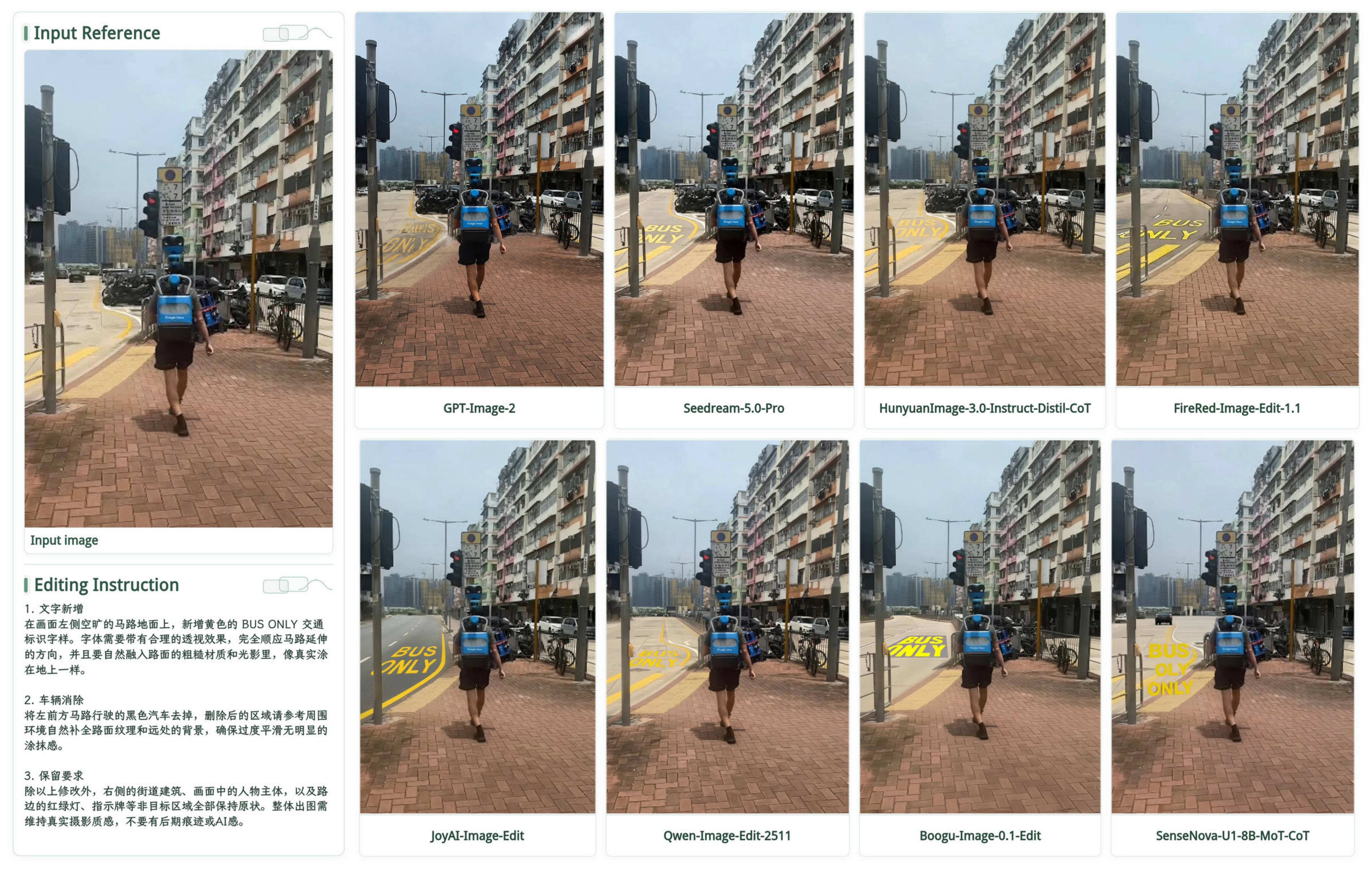}
\caption{\textedit{}: perspective-aware text editing (case 541). The panel shows the input image(s), editing instruction, and outputs from the labeled models.}
\label{fig:sup-t02}
\end{figure}

\begin{figure}[!htbp]
\centering
\setlength{\abovecaptionskip}{4pt}
\setlength{\belowcaptionskip}{2pt}
\includegraphics[page=3,trim=12bp 12bp 12bp 12bp,clip,width=\linewidth]{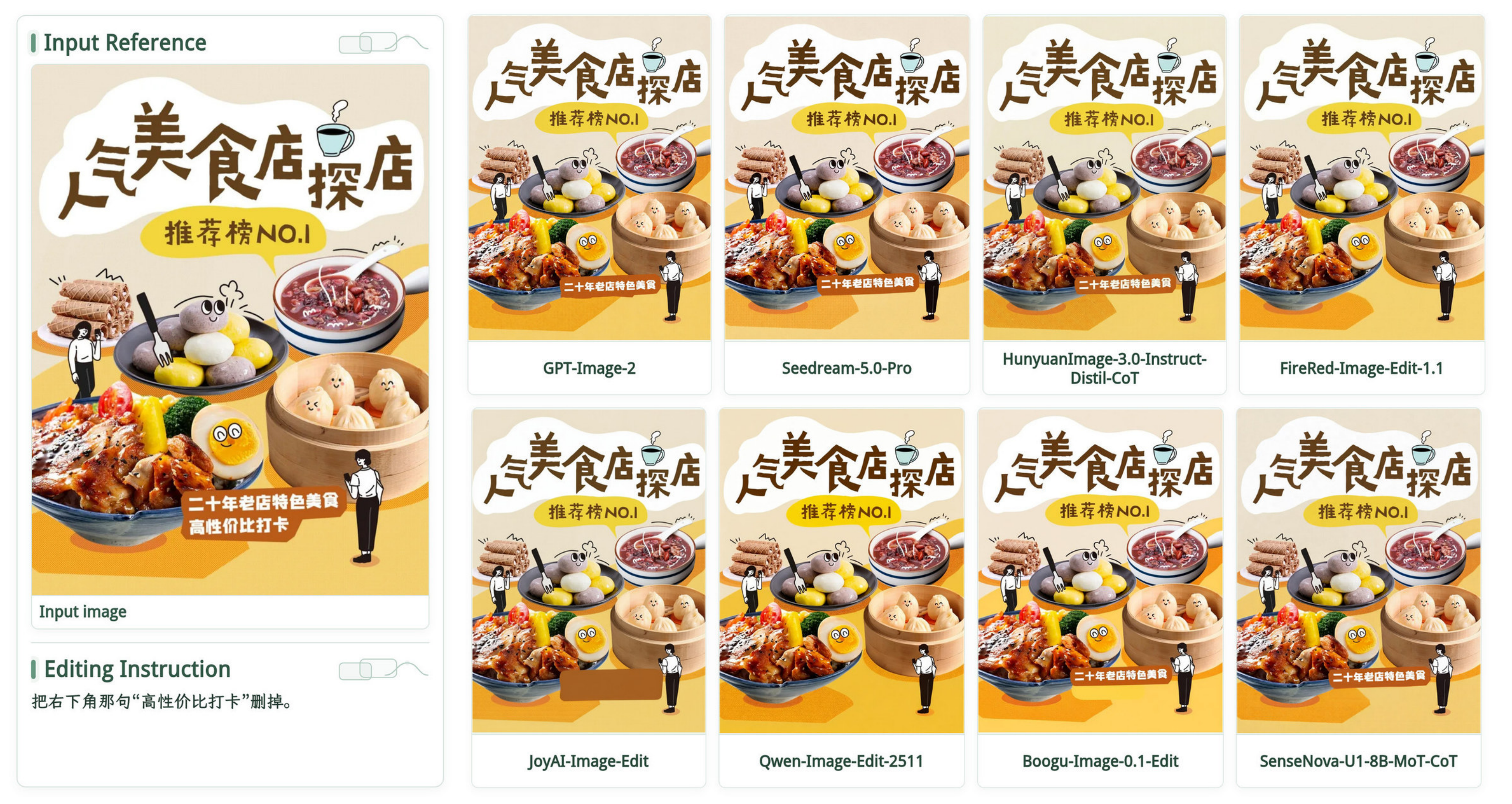}
\caption{\textedit{}: text removal (case 650). The panel shows the input image(s), editing instruction, and outputs from the labeled models.}
\label{fig:sup-t03}
\end{figure}

\begin{figure}[!htbp]
\centering
\setlength{\abovecaptionskip}{4pt}
\setlength{\belowcaptionskip}{2pt}
\includegraphics[page=4,trim=12bp 12bp 12bp 12bp,clip,width=\linewidth]{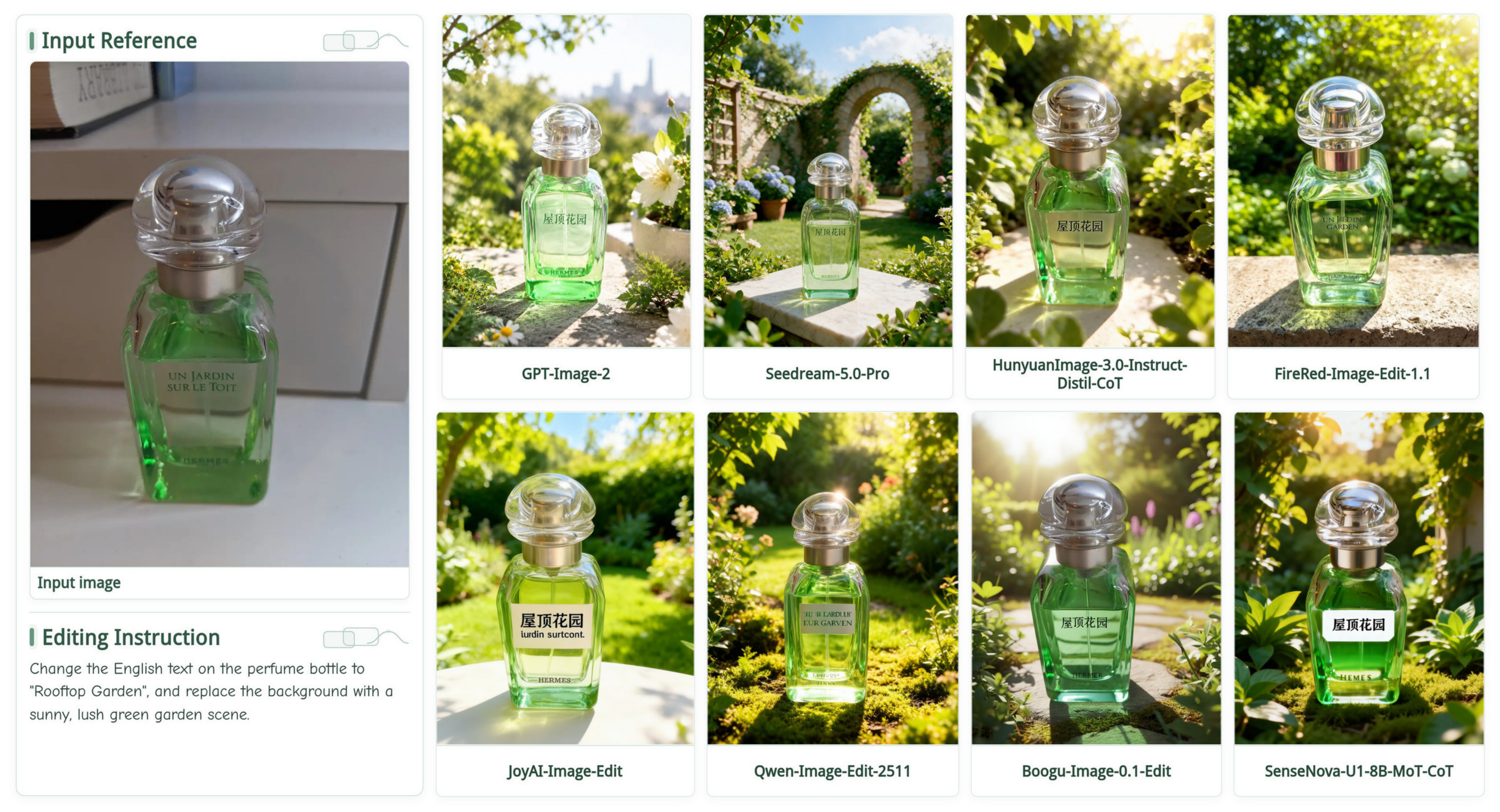}
\caption{\textedit{}: product-packaging text editing (case 399). The panel shows the input image(s), editing instruction, and outputs from the labeled models.}
\label{fig:sup-t05}
\end{figure}

\begin{figure}[!htbp]
\centering
\setlength{\abovecaptionskip}{4pt}
\setlength{\belowcaptionskip}{2pt}
\includegraphics[page=5,trim=12bp 12bp 12bp 12bp,clip,width=\linewidth]{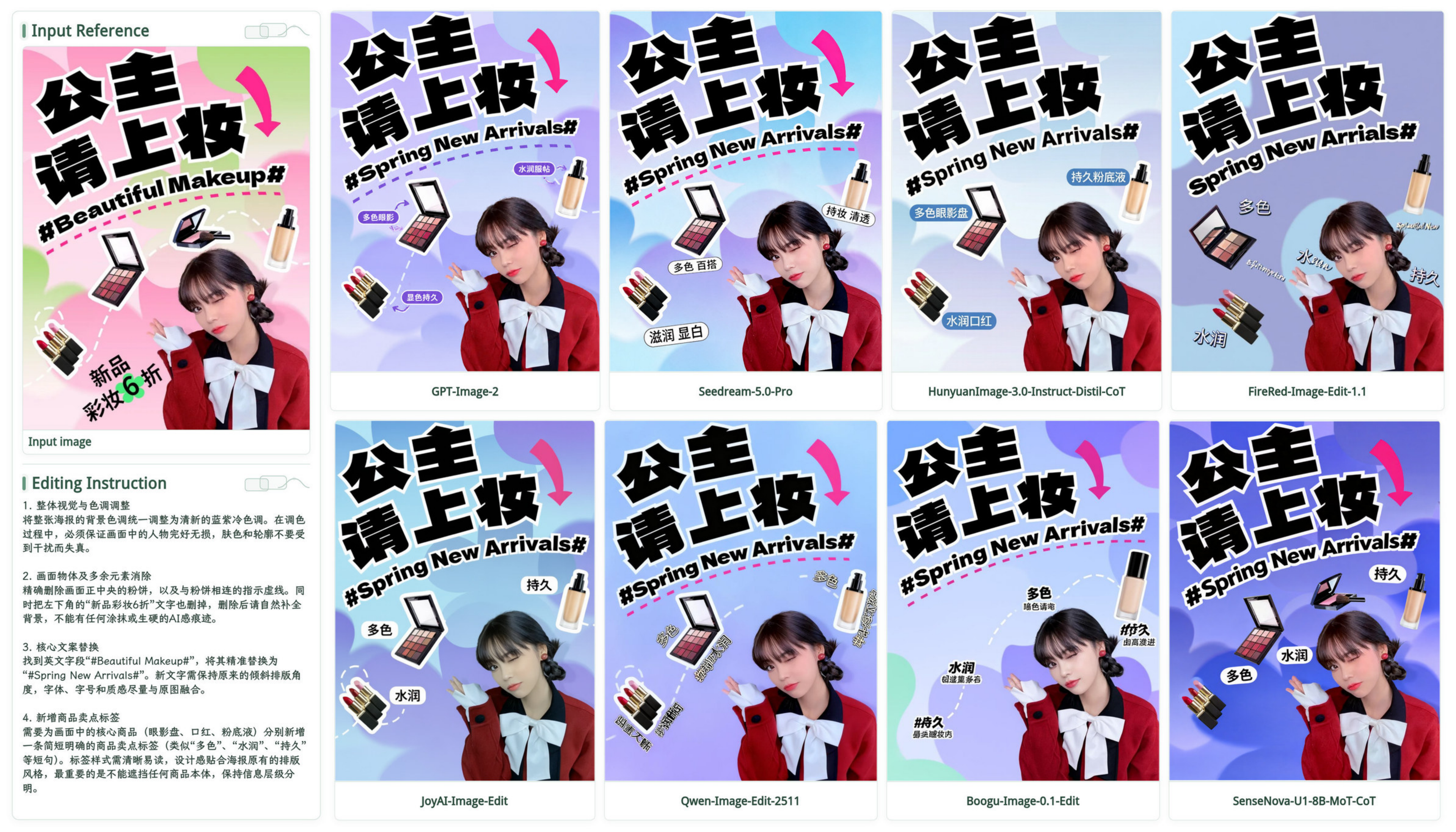}
\caption{\textedit{}: product-feature labels (case 502). The panel shows the input image(s), editing instruction, and outputs from the labeled models.}
\label{fig:sup-t07}
\end{figure}

\begin{figure}[!htbp]
\centering
\setlength{\abovecaptionskip}{4pt}
\setlength{\belowcaptionskip}{2pt}
\includegraphics[page=6,trim=12bp 12bp 12bp 12bp,clip,width=\linewidth]{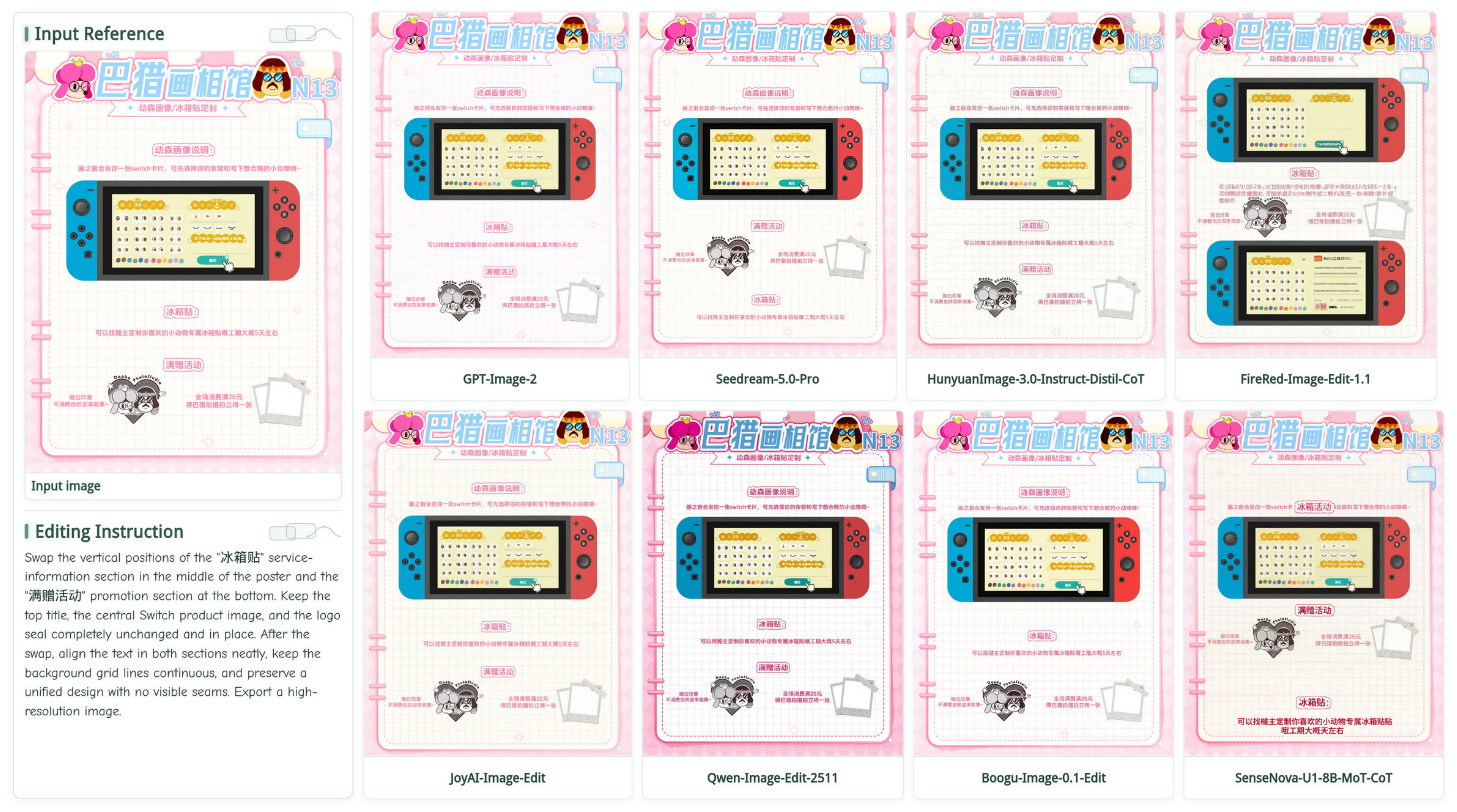}
\caption{\textedit{}: poster layout reorganization (case 111). The panel shows the input image(s), editing instruction, and outputs from the labeled models.}
\label{fig:sup-t08}
\end{figure}

\begin{figure}[!htbp]
\centering
\setlength{\abovecaptionskip}{4pt}
\setlength{\belowcaptionskip}{2pt}
\includegraphics[page=7,trim=12bp 12bp 12bp 12bp,clip,width=\linewidth]{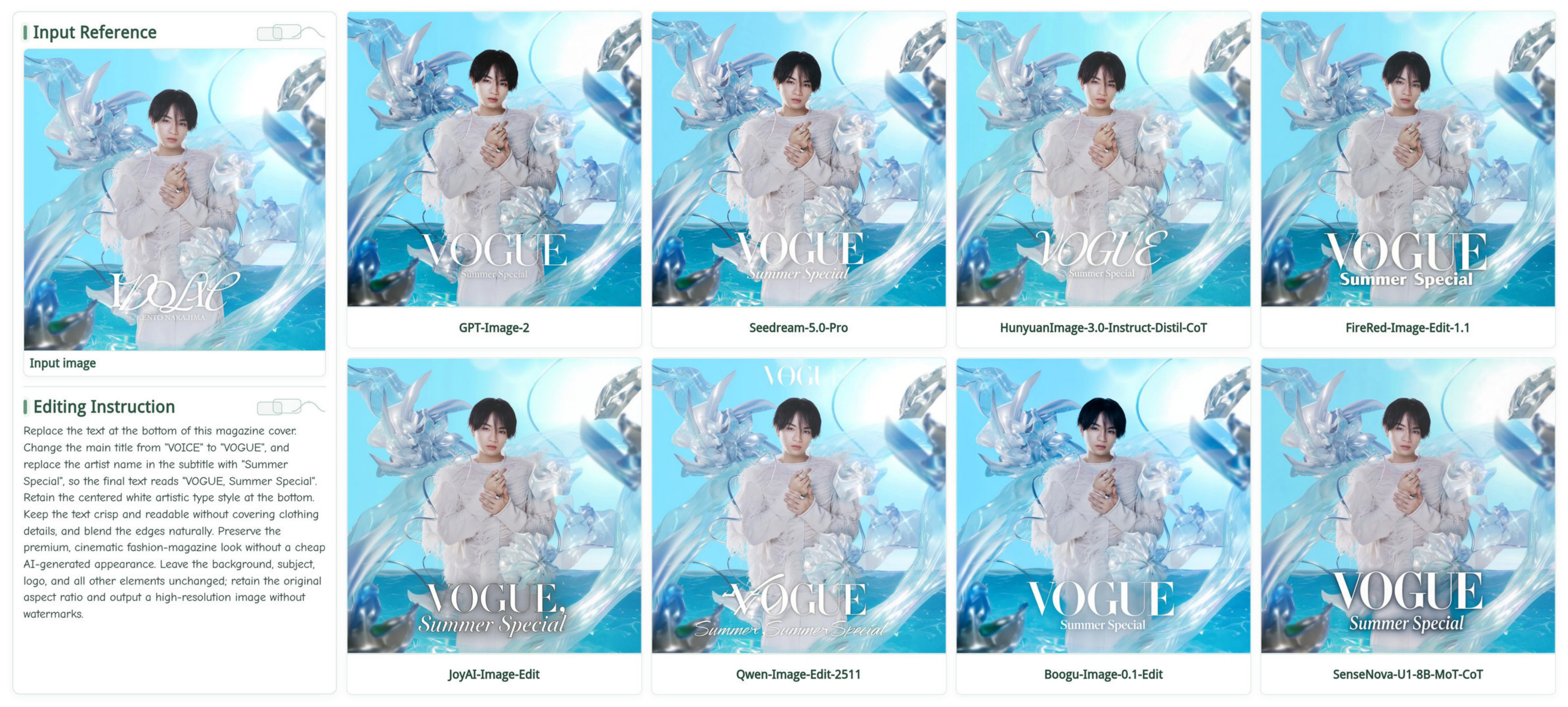}
\caption{\textedit{}: magazine cover (case 287). The panel shows the input image(s), editing instruction, and outputs from the labeled models.}
\label{fig:sup-t09}
\end{figure}

\begin{figure}[!htbp]
\centering
\setlength{\abovecaptionskip}{4pt}
\setlength{\belowcaptionskip}{2pt}
\includegraphics[page=8,trim=12bp 12bp 12bp 12bp,clip,width=\linewidth]{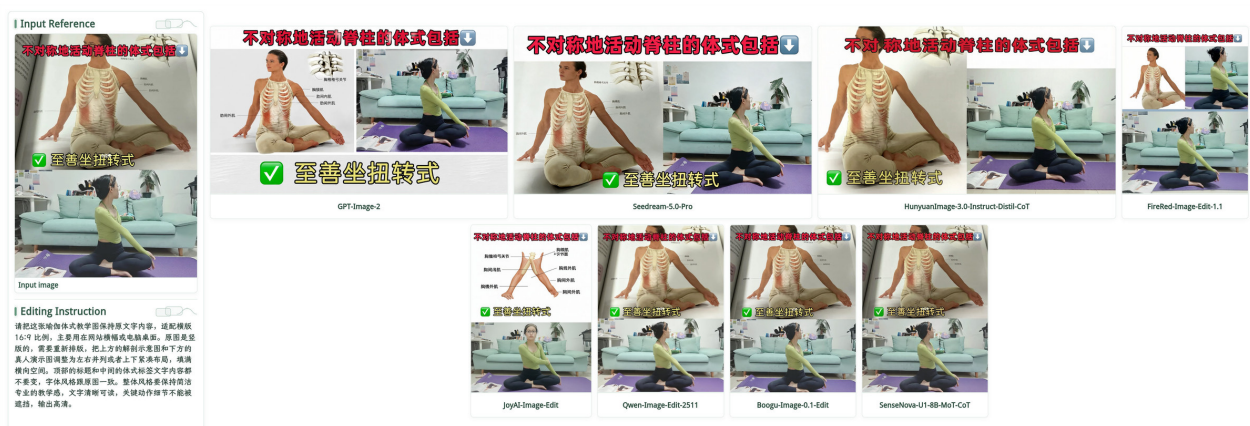}
\caption{\textedit{}: multi-format and multi-platform adaptation (case 980). The panel shows the input image(s), editing instruction, and outputs from the labeled models.}
\label{fig:sup-t10}
\end{figure}

\begin{figure}[!htbp]
\centering
\setlength{\abovecaptionskip}{4pt}
\setlength{\belowcaptionskip}{2pt}
\includegraphics[page=9,trim=12bp 12bp 12bp 12bp,clip,width=\linewidth]{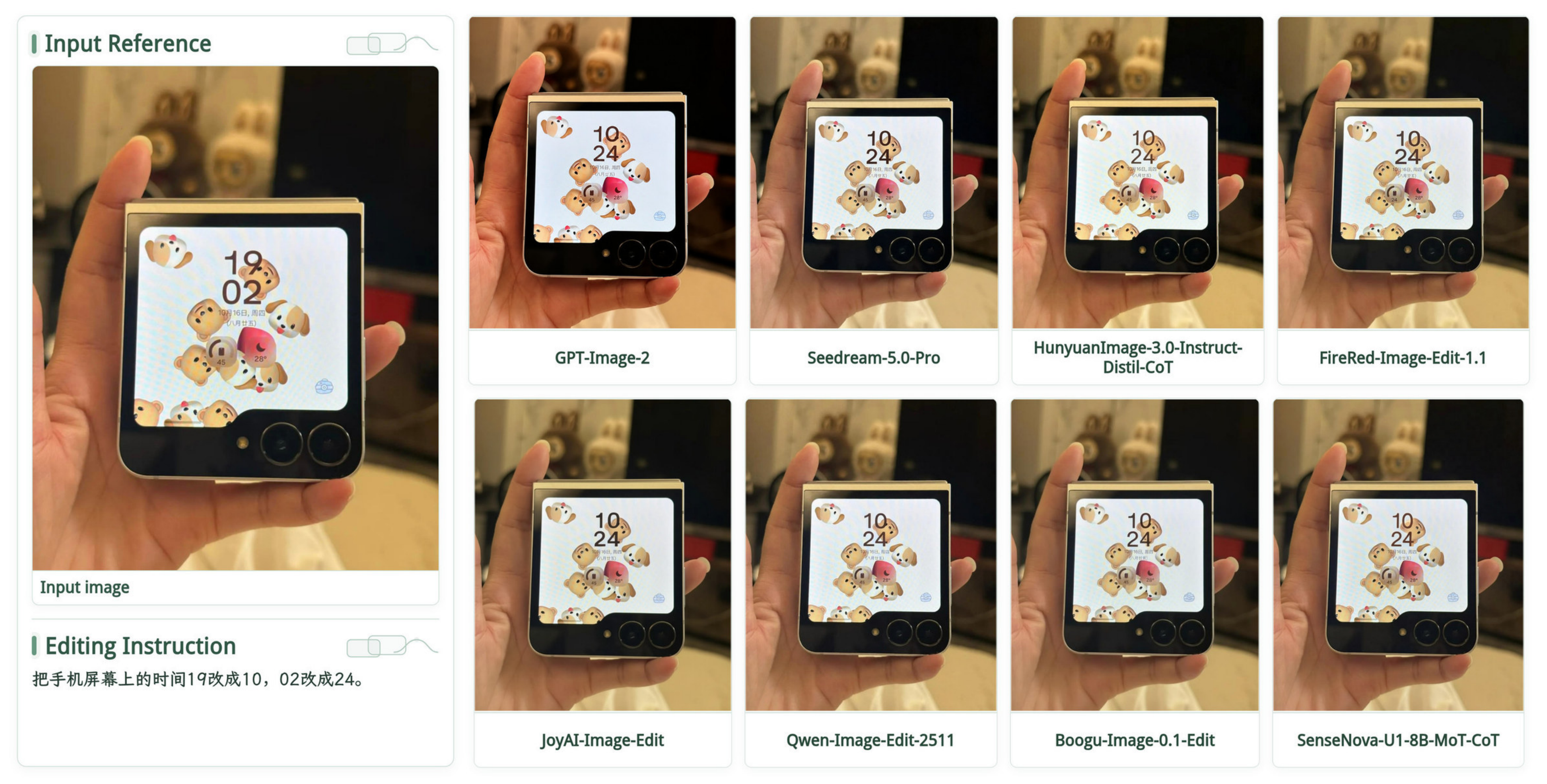}
\caption{\textedit{}: interface and screenshot text editing (case 446). The panel shows the input image(s), editing instruction, and outputs from the labeled models.}
\label{fig:sup-t12}
\end{figure}

\begin{figure}[!htbp]
\centering
\setlength{\abovecaptionskip}{4pt}
\setlength{\belowcaptionskip}{2pt}
\includegraphics[page=10,trim=12bp 12bp 12bp 12bp,clip,width=\linewidth]{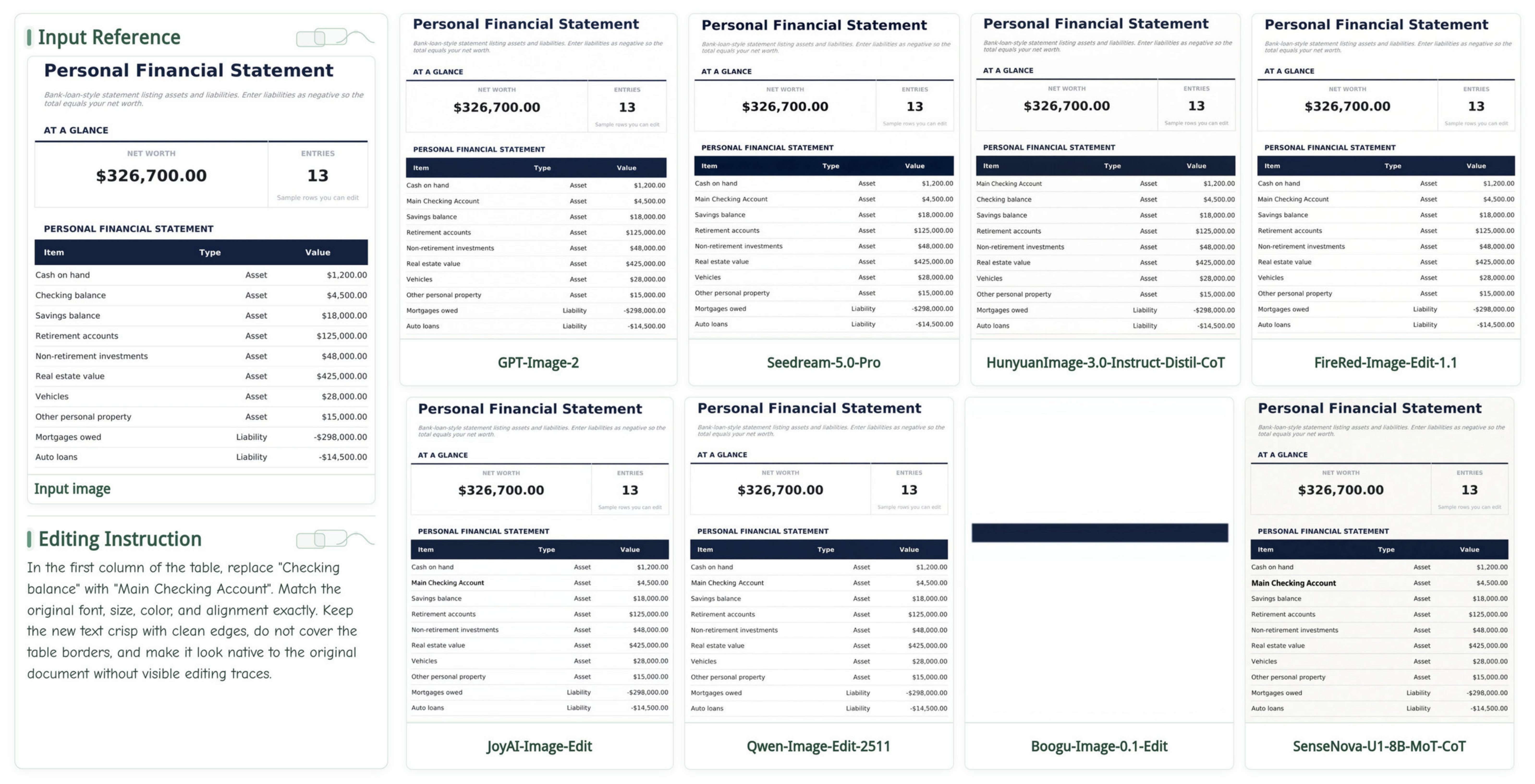}
\caption{\textedit{}: local table and document editing (case 291). The panel shows the input image(s), editing instruction, and outputs from the labeled models.}
\label{fig:sup-t13}
\end{figure}

\begin{figure}[!htbp]
\centering
\setlength{\abovecaptionskip}{4pt}
\setlength{\belowcaptionskip}{2pt}
\includegraphics[page=11,trim=12bp 12bp 12bp 12bp,clip,width=\linewidth]{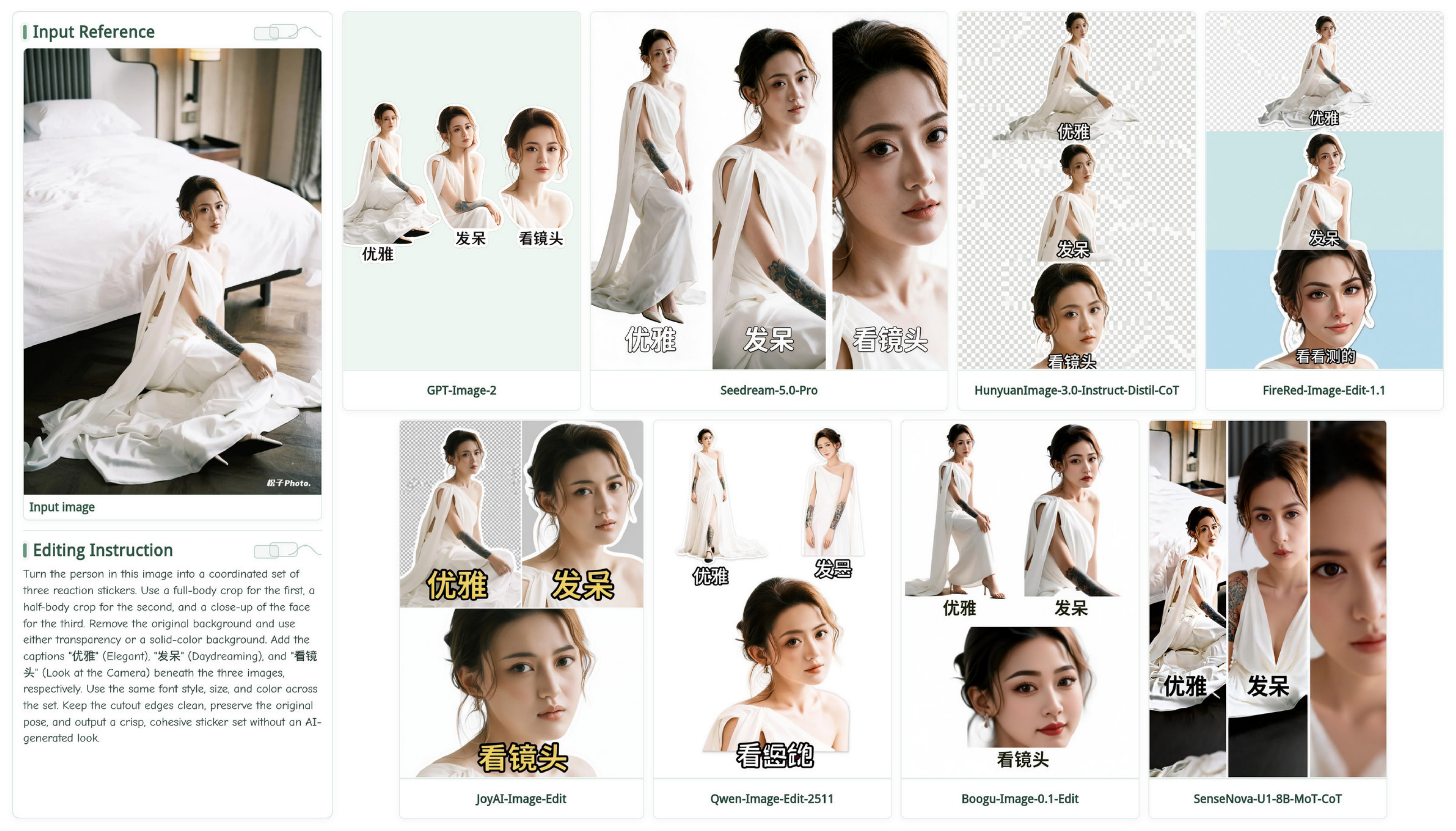}
\caption{\textedit{}: meme and sticker design (case 27). The panel shows the input image(s), editing instruction, and outputs from the labeled models.}
\label{fig:sup-t14}
\end{figure}

\begin{figure}[!htbp]
\centering
\setlength{\abovecaptionskip}{4pt}
\setlength{\belowcaptionskip}{2pt}
\includegraphics[page=12,trim=12bp 12bp 12bp 12bp,clip,width=\linewidth]{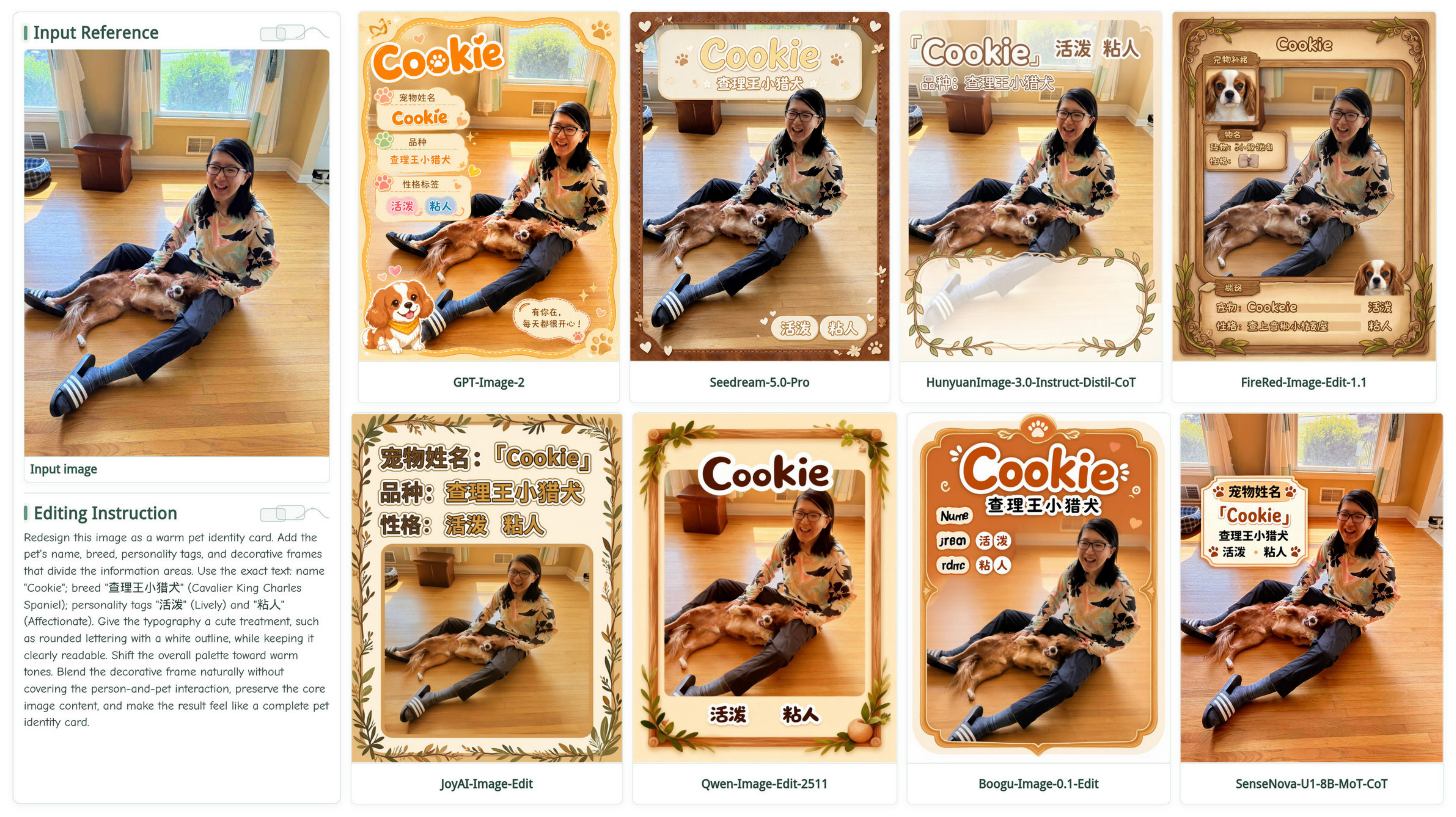}
\caption{\textedit{}: pet identification card (case 794). The panel shows the input image(s), editing instruction, and outputs from the labeled models.}
\label{fig:sup-t16}
\end{figure}

\clearpage
\subsection{\multi{}: Multi-Reference Editing Cases}
\label{app:cases-multi}
Check whether each transferred identity, attribute, pose, or scene element is bound to its assigned reference, alongside the requested composition.
\begin{figure}[!htbp]
\centering
\setlength{\abovecaptionskip}{4pt}
\setlength{\belowcaptionskip}{2pt}
\includegraphics[page=1,trim=12bp 12bp 12bp 12bp,clip,width=\linewidth]{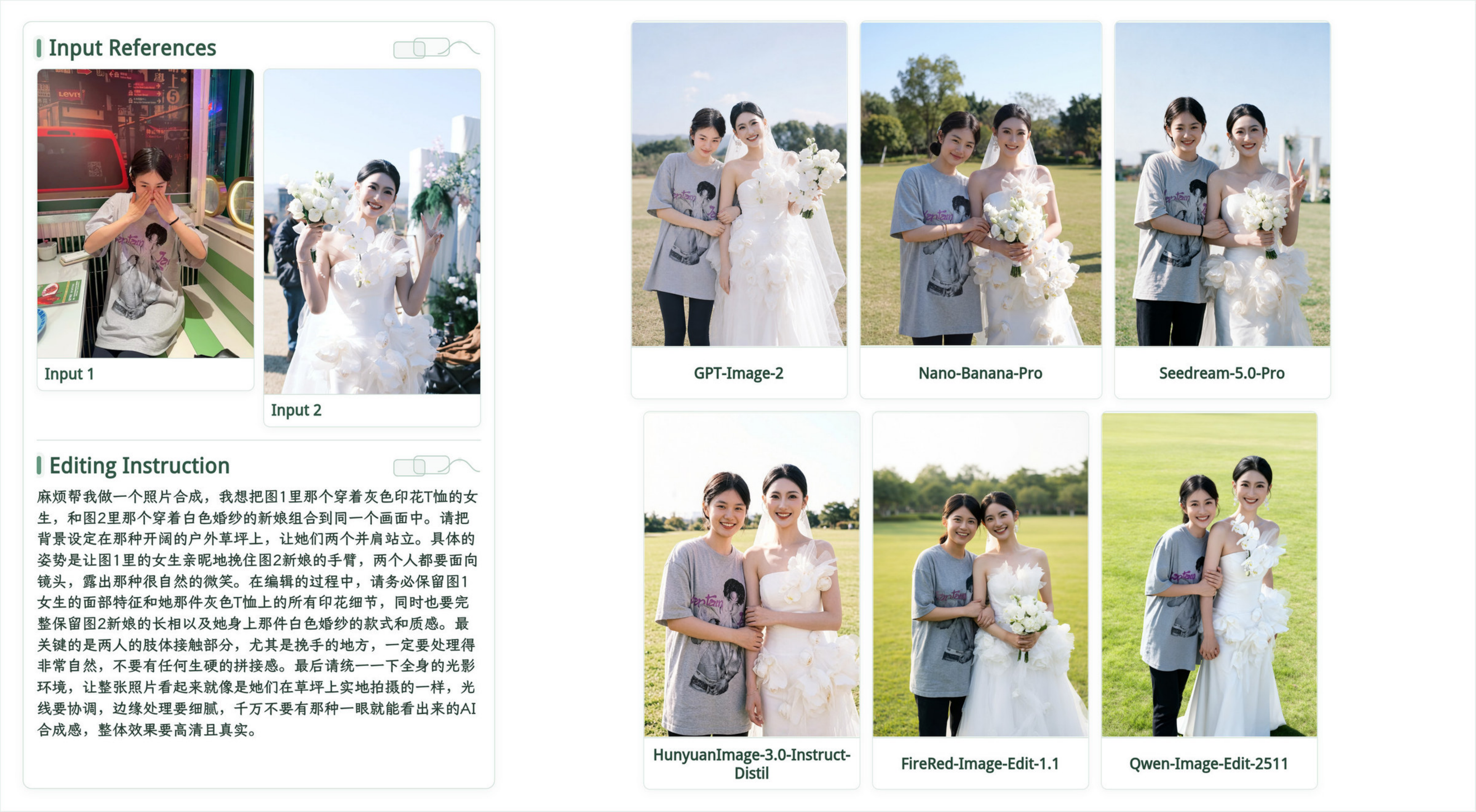}
\caption{\multi{}: interpersonal action transfer (case 164). The panel shows the input image(s), editing instruction, and outputs from the labeled models.}
\label{fig:sup-m01}
\end{figure}

\begin{figure}[!htbp]
\centering
\setlength{\abovecaptionskip}{4pt}
\setlength{\belowcaptionskip}{2pt}
\includegraphics[page=2,trim=12bp 12bp 12bp 12bp,clip,width=\linewidth]{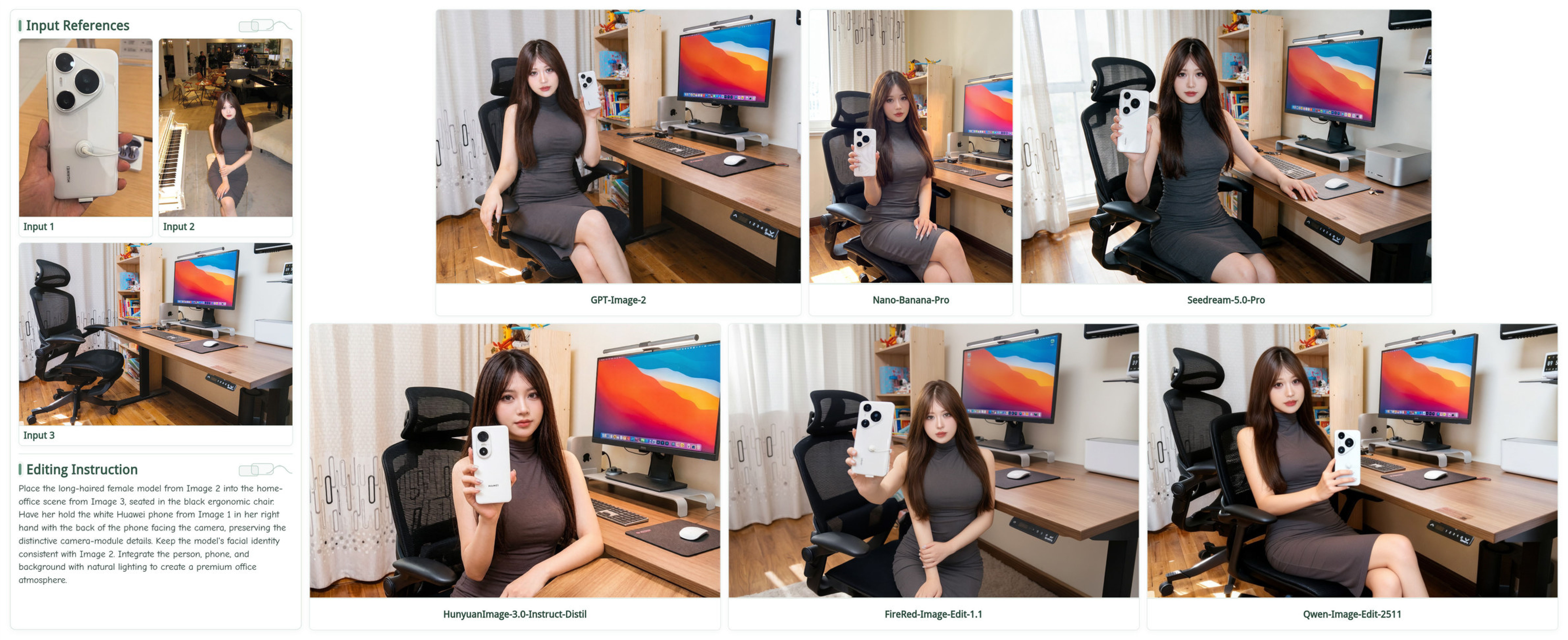}
\caption{\multi{}: model--product--scene composition (case 279). The panel shows the input image(s), editing instruction, and outputs from the labeled models.}
\label{fig:sup-m02}
\end{figure}

\begin{figure}[!htbp]
\centering
\setlength{\abovecaptionskip}{4pt}
\setlength{\belowcaptionskip}{2pt}
\includegraphics[page=3,trim=12bp 12bp 12bp 12bp,clip,width=\linewidth]{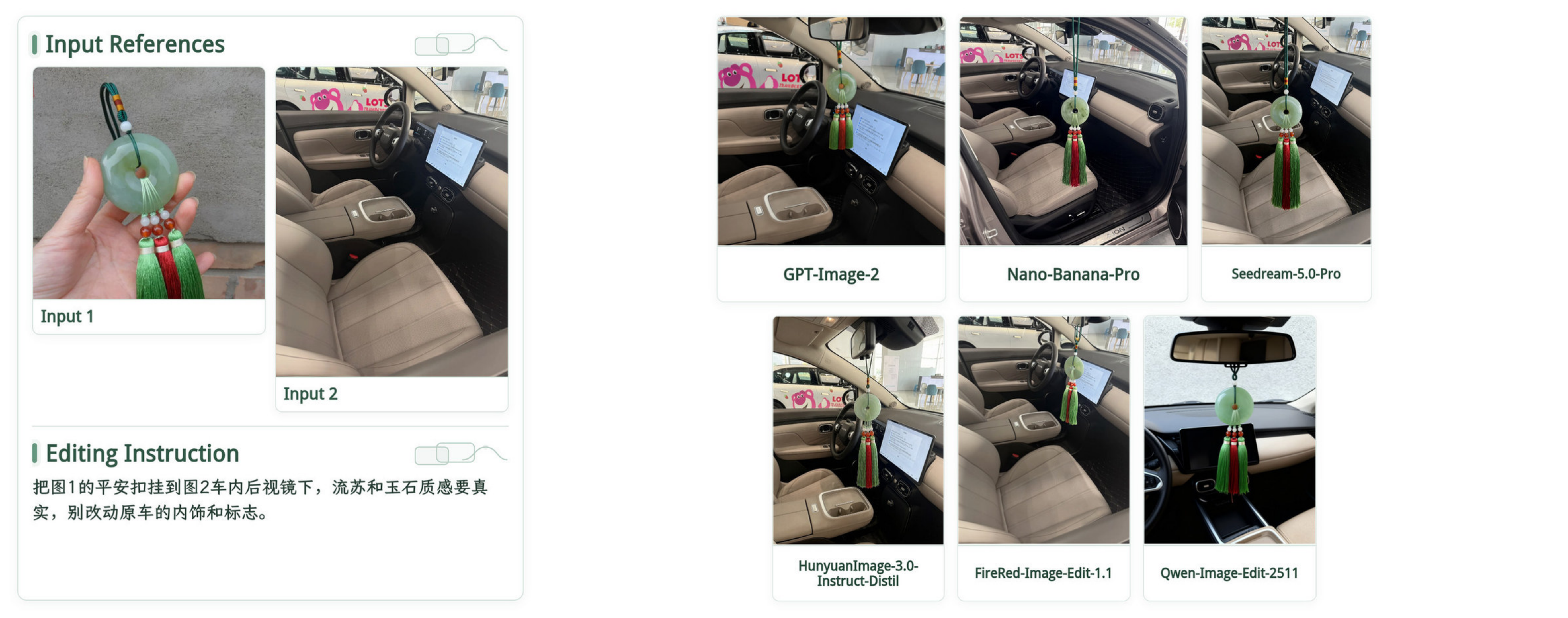}
\caption{\multi{}: multi-product composition (case 573). The panel shows the input image(s), editing instruction, and outputs from the labeled models.}
\label{fig:sup-m03}
\end{figure}

\begin{figure}[!htbp]
\centering
\setlength{\abovecaptionskip}{4pt}
\setlength{\belowcaptionskip}{2pt}
\includegraphics[page=4,trim=12bp 12bp 12bp 12bp,clip,width=\linewidth]{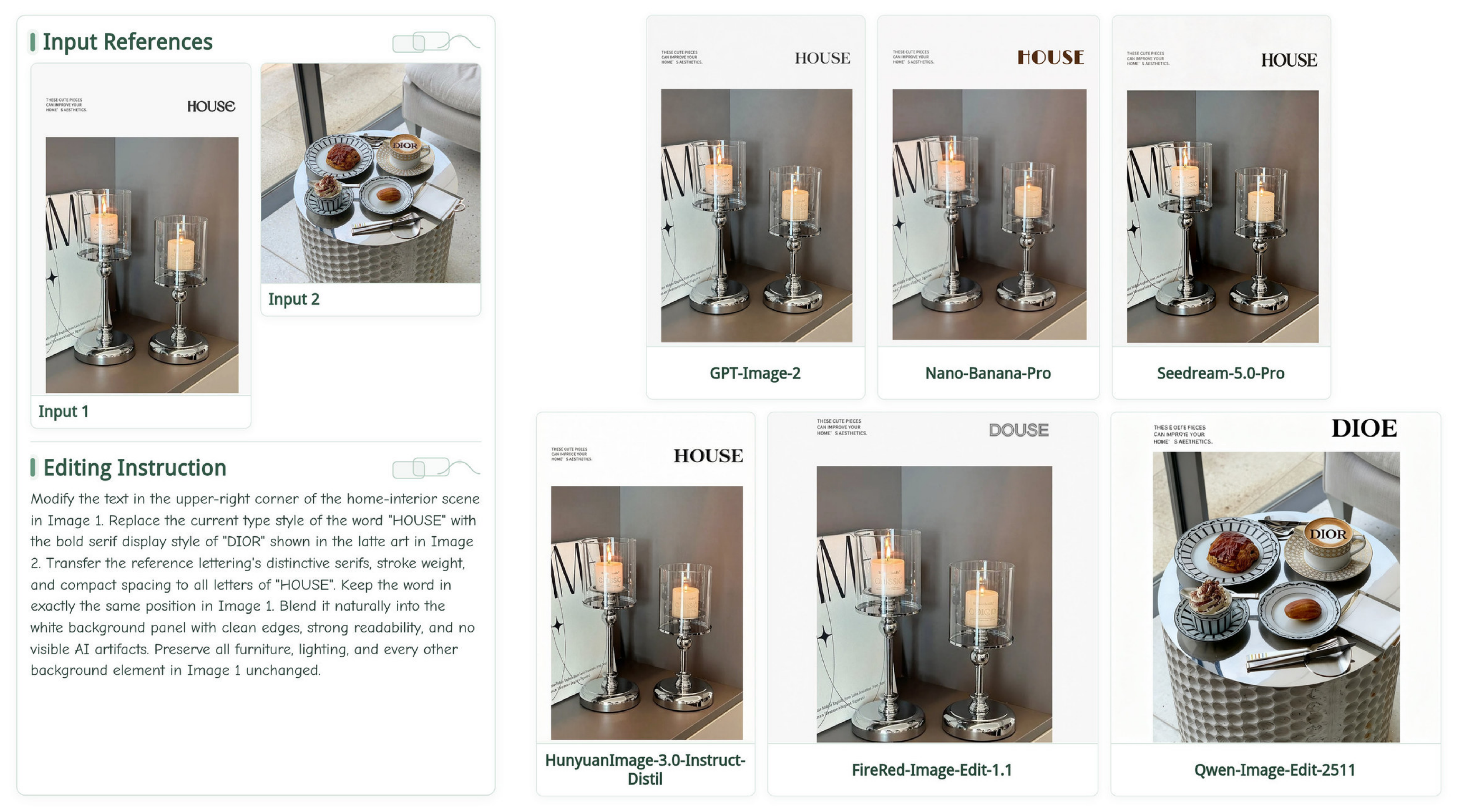}
\caption{\multi{}: font-reference transfer (case 439). The panel shows the input image(s), editing instruction, and outputs from the labeled models.}
\label{fig:sup-m04}
\end{figure}

\begin{figure}[!htbp]
\centering
\setlength{\abovecaptionskip}{4pt}
\setlength{\belowcaptionskip}{2pt}
\includegraphics[page=5,trim=12bp 12bp 12bp 12bp,clip,width=\linewidth]{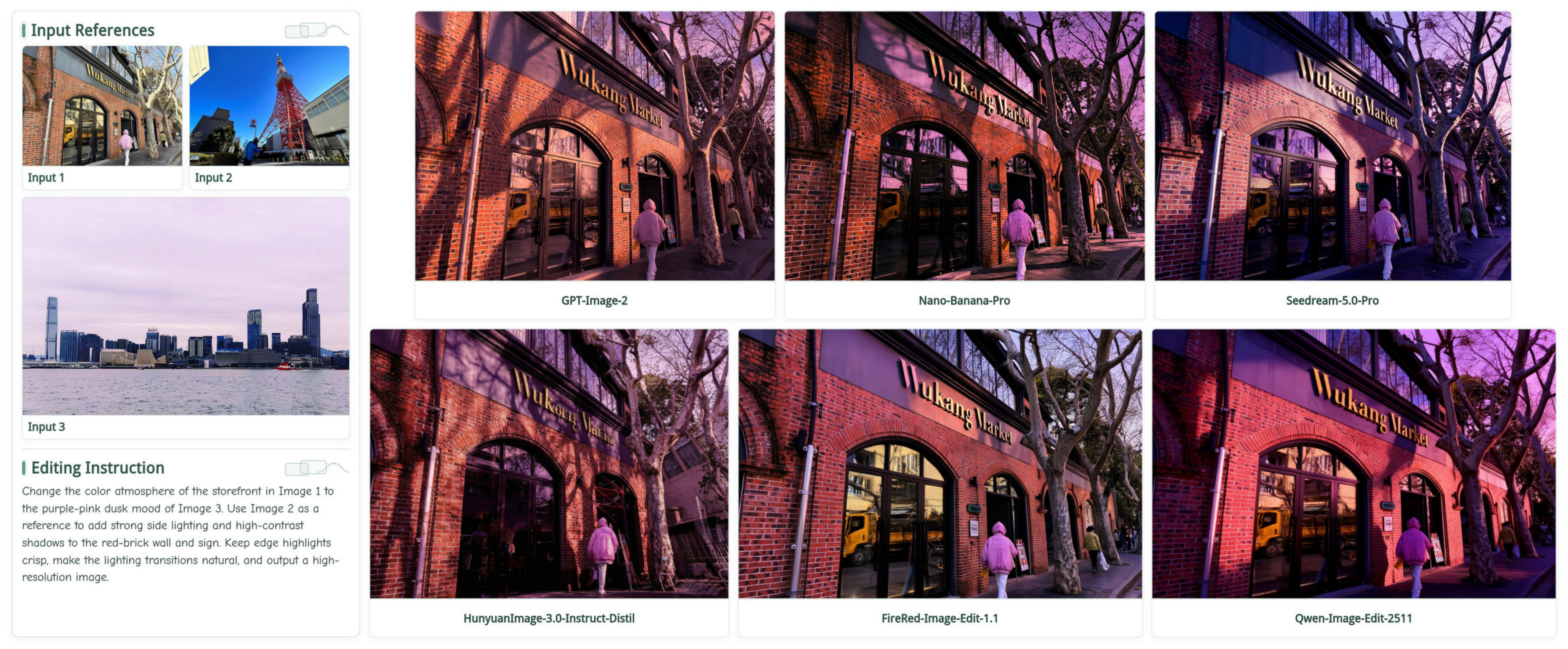}
\caption{\multi{}: scene lighting and time-of-day transfer (case 173). The panel shows the input image(s), editing instruction, and outputs from the labeled models.}
\label{fig:sup-m05}
\end{figure}

\begin{figure}[!htbp]
\centering
\setlength{\abovecaptionskip}{4pt}
\setlength{\belowcaptionskip}{2pt}
\includegraphics[page=6,trim=12bp 12bp 12bp 12bp,clip,width=\linewidth]{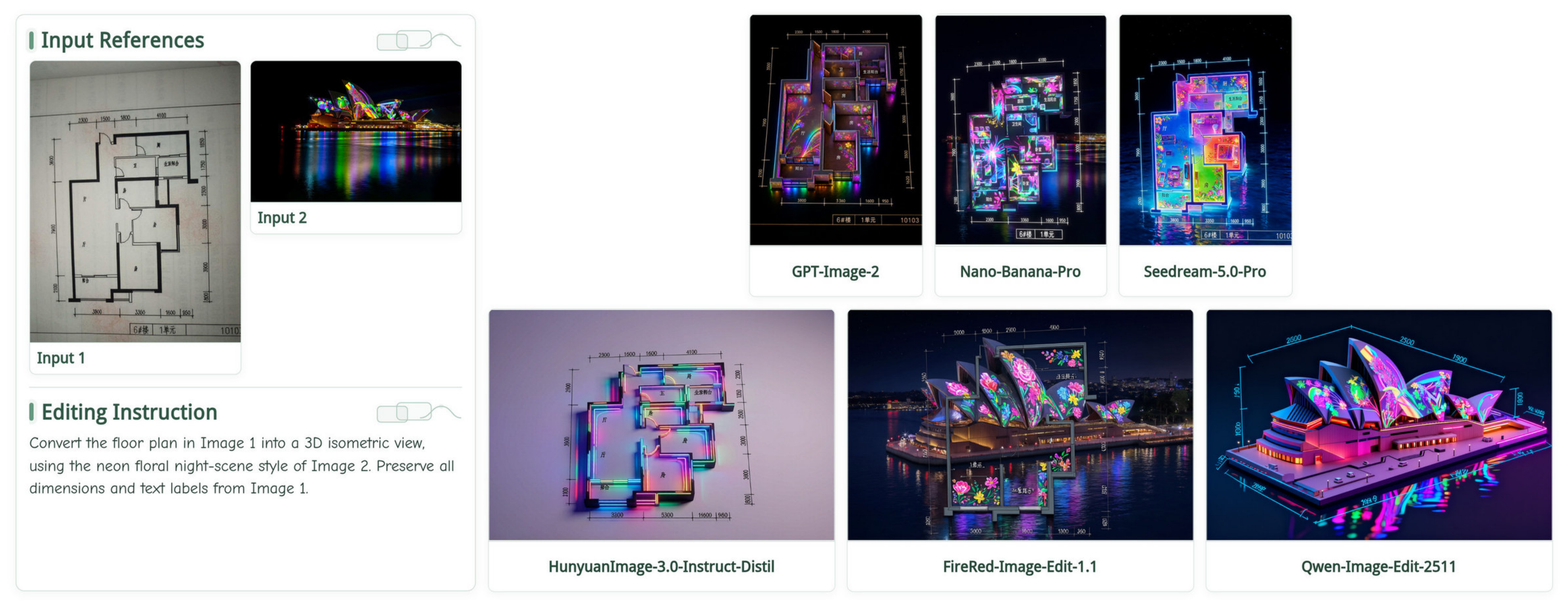}
\caption{\multi{}: map or floor-plan to scene (case 674). The panel shows the input image(s), editing instruction, and outputs from the labeled models.}
\label{fig:sup-m06}
\end{figure}

\begin{figure}[!htbp]
\centering
\setlength{\abovecaptionskip}{4pt}
\setlength{\belowcaptionskip}{2pt}
\includegraphics[page=7,trim=12bp 12bp 12bp 12bp,clip,width=\linewidth]{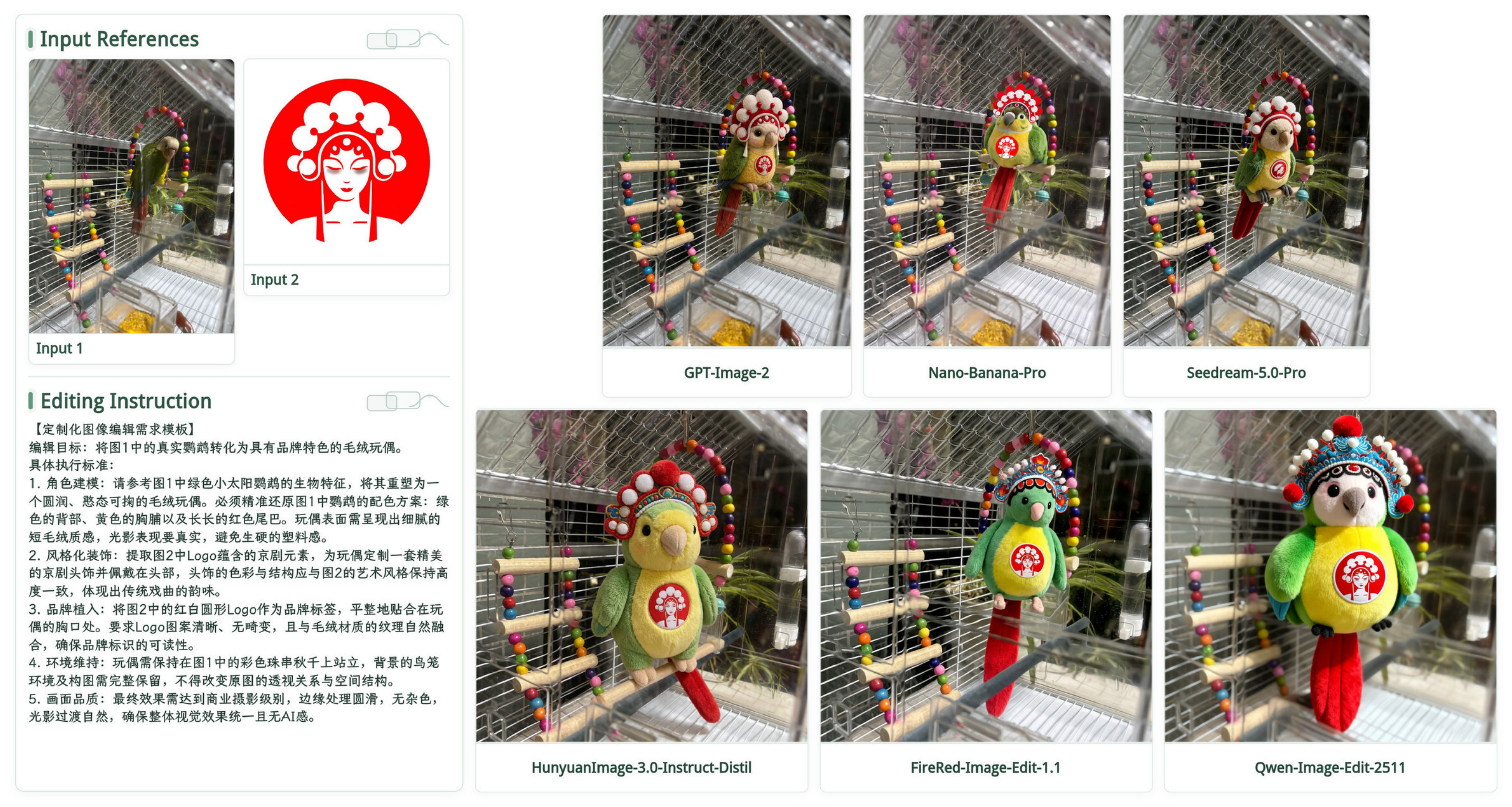}
\caption{\multi{}: pet-to-toy transformation (case 277). The panel shows the input image(s), editing instruction, and outputs from the labeled models.}
\label{fig:sup-m07}
\end{figure}

\begin{figure}[!htbp]
\centering
\setlength{\abovecaptionskip}{4pt}
\setlength{\belowcaptionskip}{2pt}
\includegraphics[page=8,trim=12bp 12bp 12bp 12bp,clip,width=\linewidth]{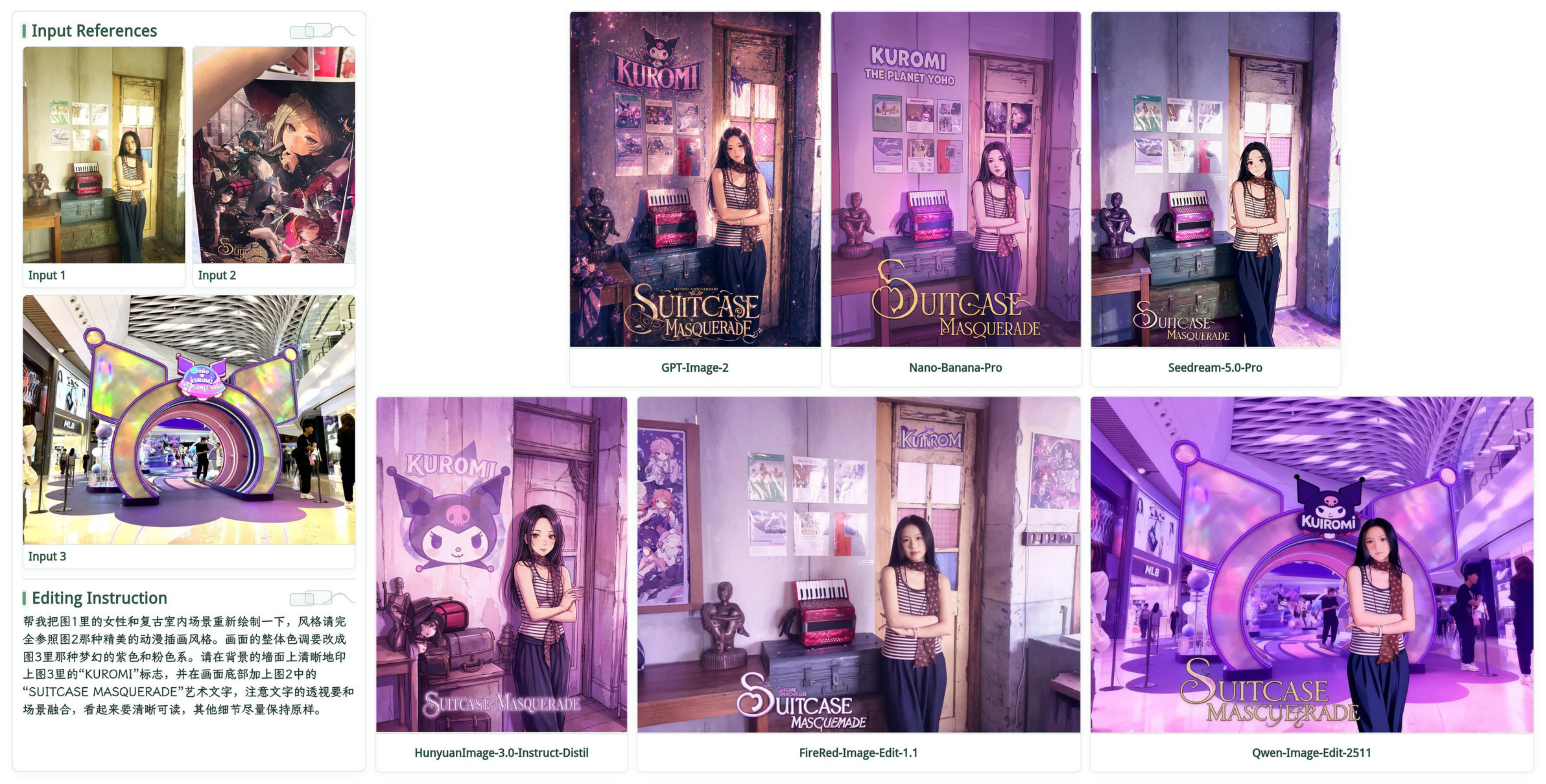}
\caption{\multi{}: style transfer (case 203). The panel shows the input image(s), editing instruction, and outputs from the labeled models.}
\label{fig:sup-m08}
\end{figure}

\clearpage
\subsection{Qualitative Evidence of Model Adaptation}
\label{app:evolution-qualitative-cases}
\begin{figure}[!htp]
\centering
\setlength{\abovecaptionskip}{4pt}
\setlength{\belowcaptionskip}{2pt}
\includegraphics[width=\linewidth]{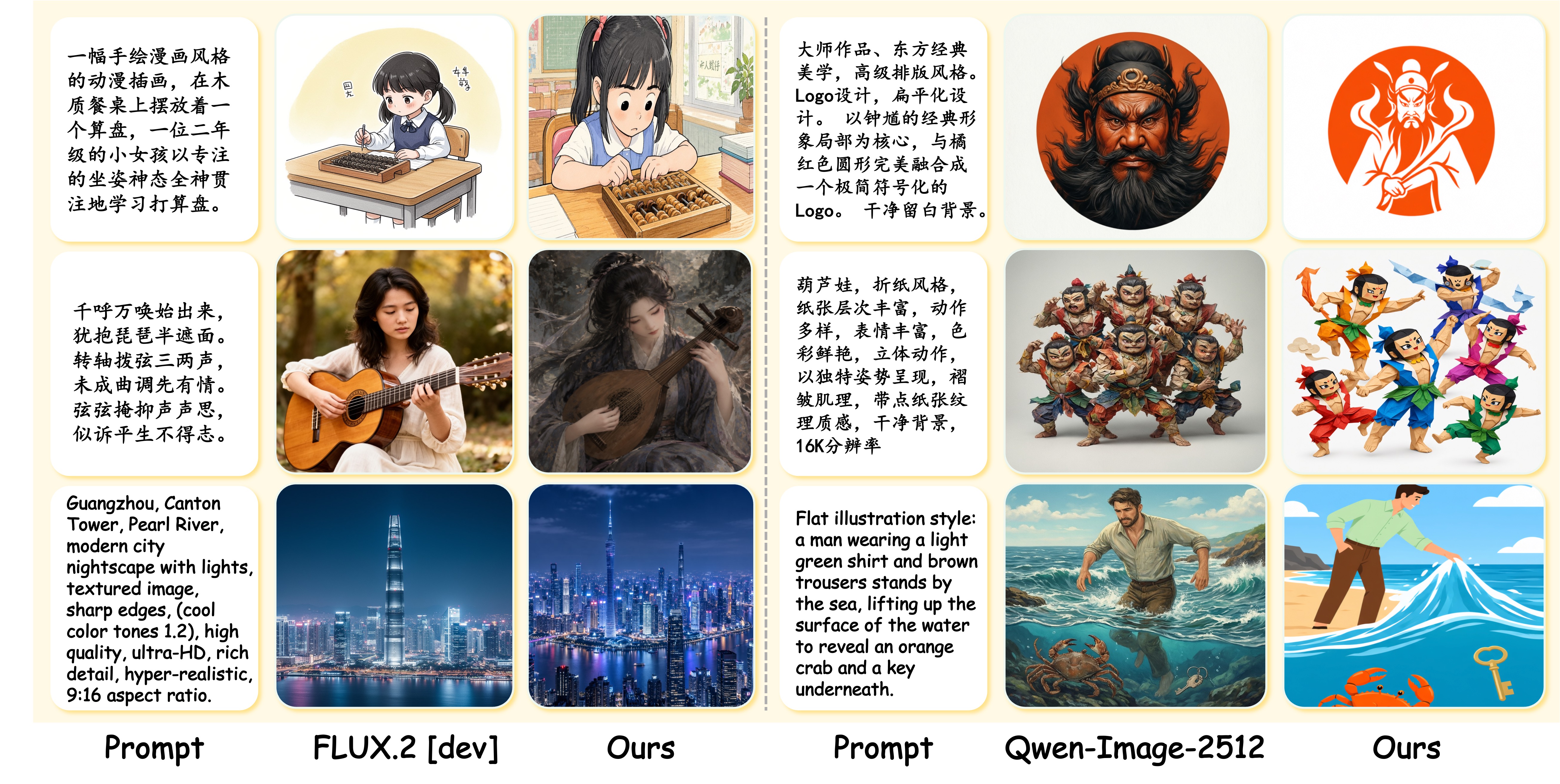}
\caption{Qualitative text-to-image comparisons before and after \studio{} adaptation. Each triplet shows the prompt, baseline, and adapted output, with FLUX.2 [dev] on the left and Qwen-Image-2512 on the right.}
\label{fig:additional-generation}
\vspace{10pt}
\includegraphics[width=\linewidth]{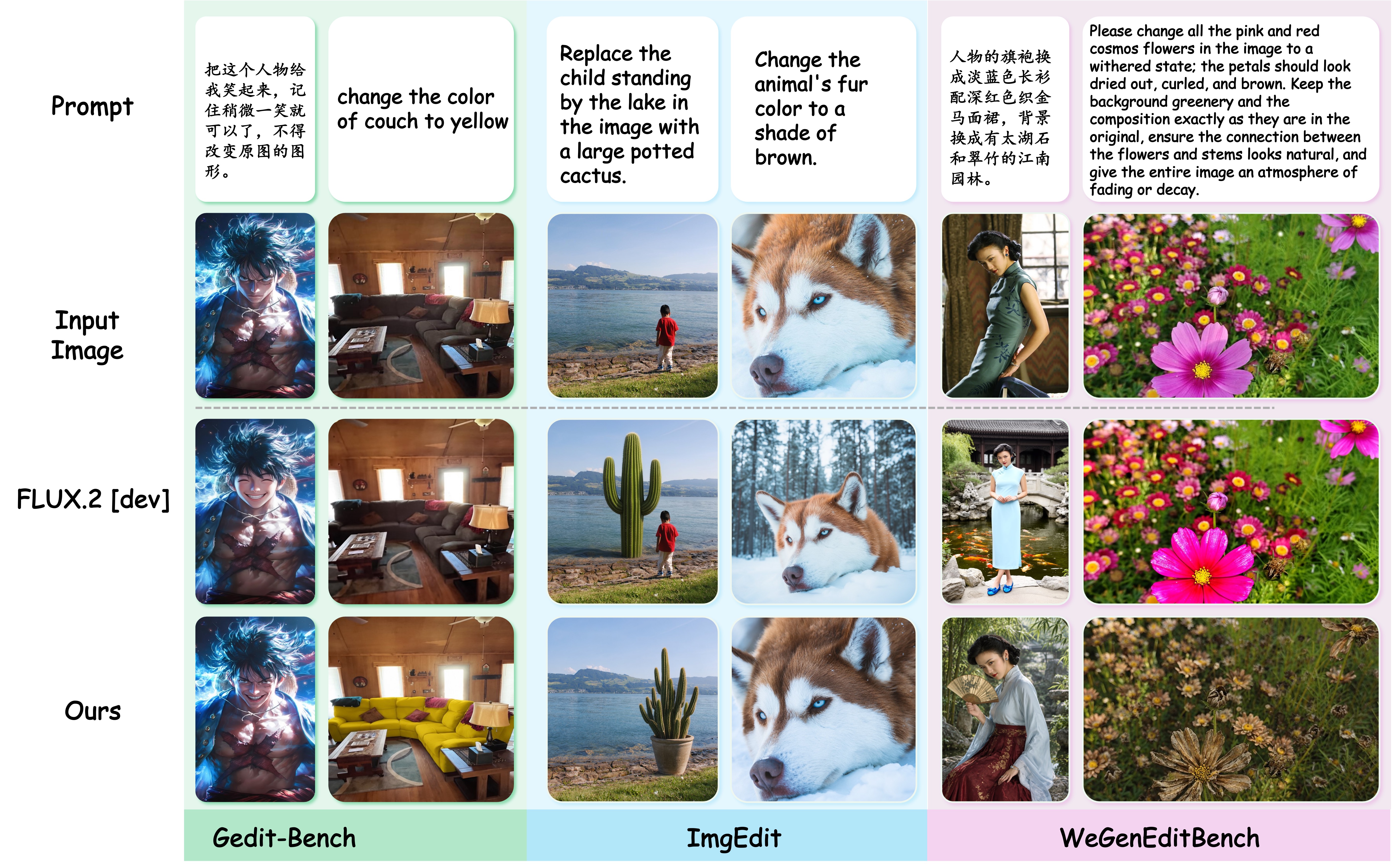}
\caption{Additional editing comparisons for FLUX.2 [dev], following Figure~\ref{fig:evolution-qualitative}. Examples cover expression and color changes, object replacement, clothing and scene composition, and flower withering.}
\label{fig:additional-flux-editing}
\end{figure}

\end{document}